\documentclass{article}

\usepackage{microtype}
\usepackage{graphicx}
\usepackage{subfigure}
\usepackage{booktabs} 

\usepackage{hyperref}

\usepackage[accepted]{icml2021}

\usepackage{amsfonts}       
\usepackage{amsmath}        
\usepackage{amssymb}        
\usepackage{graphicx}       
\usepackage{bm}             
\usepackage{url}            
\usepackage{nicefrac}       
\usepackage[symbol]{footmisc}
\usepackage{float}
\usepackage{wrapfig}

\newcommand{\bc}{\bm{c}}
\newcommand{\bx}{\bm{x}}
\newcommand{\bz}{\bm{z}}

\newcommand{\cD}{\ensuremath{\mathcal{D}}}

\newcommand{\mR}{\mathbb R}

\newcommand{\NN}[1]{\phi_\theta(#1)}

\def\Vhrulefill{\leavevmode\leaders\hrule height 0.7ex depth \dimexpr0.4pt-0.7ex\hfill\kern0pt}

\icmltitlerunning{Rethinking Assumptions in Deep Anomaly Detection}

\begin{document}

\twocolumn[
\icmltitle{Rethinking Assumptions in Deep Anomaly Detection}



\icmlsetsymbol{equal}{*}

\begin{icmlauthorlist}

\icmlauthor{Lukas Ruff}{aix,equal}
\icmlauthor{Robert A.~Vandermeulen}{tub,equal}
\icmlauthor{Billy Joe Franks}{tukl}
\icmlauthor{Klaus-Robert M{\"u}ller}{tub,kor,mpi}
\icmlauthor{Marius Kloft}{tukl}
\end{icmlauthorlist}

\icmlaffiliation{aix}{Aignostics, Germany (majority of work done while with TU Berlin)}
\icmlaffiliation{tub}{TU Berlin, Germany}
\icmlaffiliation{tukl}{TU Kaiserslautern, Germany}
\icmlaffiliation{kor}{Korea University, Seoul, South Korea}
\icmlaffiliation{mpi}{MPII, Saarbr{\"u}cken, Germany}

\icmlcorrespondingauthor{Lukas Ruff}{contact@lukasruff.com}

\icmlkeywords{Anomaly Detection, Deep Learning, ICML, Novelty Detection, One-Class Classification, Outlier Detection}

\vskip 0.3in
]



\printAffiliationsAndNotice{\icmlEqualContribution} 

\begin{abstract}
Though anomaly detection (AD) can be viewed as a classification problem (nominal vs.~anomalous) it is usually treated in an unsupervised manner since one typically does not have access to, or it is infeasible to utilize, a dataset that sufficiently characterizes what it means to be ``anomalous.'' 
In this paper we present results demonstrating that this intuition surprisingly seems not to extend to deep AD on images. 
For a recent AD benchmark on ImageNet, classifiers trained to discern between normal samples and just a few (64) random natural images are able to outperform the current state of the art in deep AD.
Experimentally we discover that the multiscale structure of image data makes example anomalies exceptionally informative.
\end{abstract}

\section{Introduction}
\label{sec:intro}

Anomaly detection (AD) \citep{chandola2009anomaly} is the task of determining if a sample is anomalous compared to a corpus of data. 
Recently there has been a great interest in developing novel deep methods for AD \cite{ruff2021,pang2021}. 
Some of the best performing new AD methods for images were proposed by \citet{golan2018deep} and \citet{hendrycks2019using}. 
These methods, like most previous works on AD, are performed in an \emph{unsupervised} way: they only utilize an unlabeled corpus of mostly nominal data. 
While AD can be interpreted as a classification problem of ``nominal vs.~anomalous,'' it is typically treated as an unsupervised problem due to the rather tricky issue of finding or constructing a dataset that somehow captures \emph{everything different} from a nominal dataset.

\begin{figure}[ht] \label{fig:toy_example}
  \centering
  %
  \includegraphics[width=1.01\columnwidth]{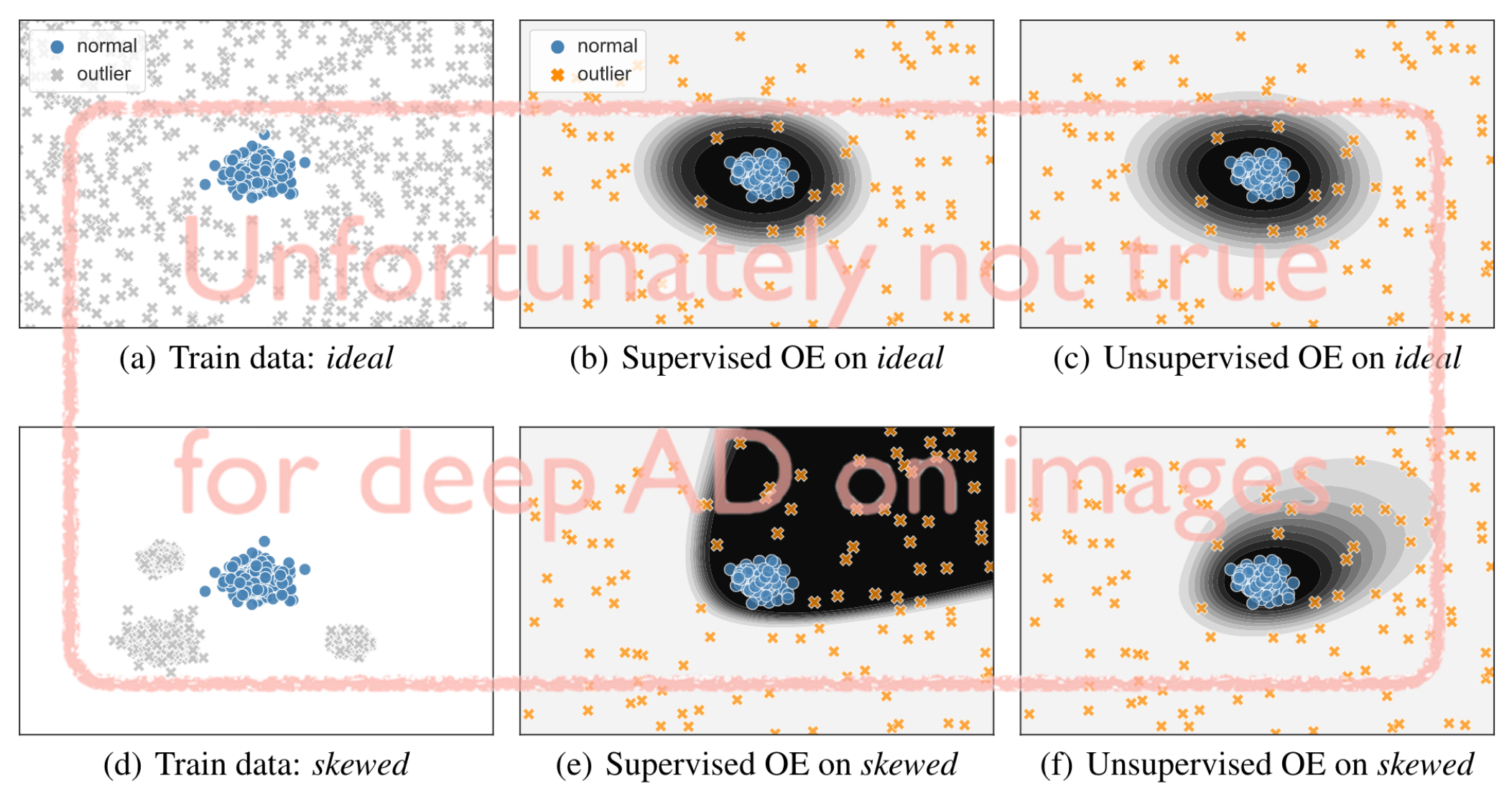}
  %
  \vspace{-1.2em}
  \caption{The decision boundaries of a supervised OE method (neural net with binary cross entropy) and an unsupervised OE method (neural net with hypersphere loss) on two toy data settings: \emph{ideal} ((a)--(c)) and \emph{skewed} ((d)--(f)).
  The unsupervised OE method ((c) + (f)) learns compact decision regions of the nominal class. A supervised OE approach ((b) + (e)) learns decision regions that do not generalize well on this toy AD task. Our results suggest that this intuition does not hold for a deep approach to image AD, where supervised OE performs remarkably well.}
\end{figure}

One often has, in addition to a corpus of nominal data, access to some data which is known to be anomalous. 
There exist deep methods for incorporating anomalous data to augment unsupervised AD \cite{hendrycks2019deep,ruff2020}. 
This setting has also been called ``semi-supervised'' AD \cite{gornitz2013,ruff2020}. 
In \citet{hendrycks2019deep} it was noted that, for an image AD problem, one has access to a virtually limitless amount of random natural images from the internet that are likely not nominal, and that such data should be utilized to improve unsupervised methods. 
They term the utilization of such data \emph{outlier exposure} (OE). 
The state-of-the-art method presented in \citet{hendrycks2019using} utilizes tens of thousands of OE samples combined with a modified version of the self-supervised method from \citet{golan2018deep} and is one of the best performing AD method to date on standard image AD benchmarks. 
For clarity, we here delineate the following three basic approaches to anomaly detection:\\[0.2em]
\underline{Unsupervised:} Methods trained on (mostly) nominal data. This is the classic and most common approach to AD.\\[0.2em]
\underline{Unsupervised OE:} Adaptations of unsupervised methods that incorporate auxiliary data that is not nominal. Elsewhere this is also called ``semi-supervised'' AD \cite{gornitz2013,ruff2020}.\\[0.2em]
\underline{Supervised OE:} The approach of simply applying a standard classification method to discern between nominal data and an auxiliary dataset that is not nominal.\\[0.2em]
Using unsupervised OE rather than supervised OE to discern between the nominal data and known anomalies seems intuitive since the presented anomalies likely do not completely characterize ``anomalousness.''
This is illustrated in Figure \ref{fig:toy_example}.  
This intuition and the benefits of the unsupervised OE approach when incorporating known anomalies has also been observed previously \cite{tax2001,gornitz2013,ruff2020}.

In this paper, we present experimental results that challenge the assumption that deep AD on images needs an unsupervised approach (with or without OE). 
We find that, using the same experimental OE setup as \citet{hendrycks2019using}, a standard classifier is able to outperform current state-of-the-art AD methods on the one vs.~rest AD benchmarks on MNIST and CIFAR-10.  
The one vs.~rest benchmark has been recommended as a general approach to experimentally validate AD methods \cite{emmott13}. This benchmark applied to the aforementioned datasets is used as a litmus test in virtually all deep AD papers published at top-tier venues; see for example \cite{ruff2018deep,deecke2018image,golan2018deep,hendrycks2019using,samet19,abati19,perera19,wang19,ruff2020,bergman20,kim20}. 
Additionally, we find that remarkably few OE examples are necessary to characterize ``anomalousness.'' 
With 128 OE samples a classifier is competitive with state-of-the-art unsupervised methods on the CIFAR-10 one vs.~rest benchmark. 
With only 64 OE samples a classifier outperforms unsupervised methods (with or without OE) on the ImageNet one vs.~rest benchmark \cite{hendrycks2019using}. This test was recently proposed as a more challenging successor to the CIFAR-10 benchmark. 

Our results seem to contradict the following pieces of common wisdom in deep learning and AD:
\vspace{-0.5em}
\begin{itemize}\itemsep0em
  \item Many (thousands) samples are needed for a deep method to understand a class \cite{goodfellow16}.
  \item Anomalies are unconcentrated and thus inherently difficult to characterize with data \cite{steinwart2005classification, chapelle2009}.
\end{itemize}
\vspace{-0.5em}
These points should imply that classification with few OE samples should be ineffective at deep AD. 
Instead, we find that relatively few random OE samples are necessary to yield state-of-the-art detection performance. 
In all of our experiments, the nominal and OE data available during training are exactly those used in \cite{hendrycks2019using} and do not contain any representatives from the ground-truth anomaly classes. 
The OE data is not tailored to be representative of the anomalies used at test time.
Based on the presence of information at multiple spatial scales in images \cite{olshausen96}, which is one key difference between classic AD and deep image AD, we hypothesize that each OE image contains multiple features at different scales that present informative examples of anomalousness. 


\section{Deep One-Class Classification}
\label{sec:hypsphcla}

Deep one-class classification \citep{ruff2018deep}, which learns (or transfers) data representations such that normal data is concentrated in feature space, has been introduced as a deep learning extension of the one-class classification approach to anomaly detection \citep{scholkopf2001,tax2001,ruff2021}.
Specifically, the Deep SVDD method \citep{ruff2018deep} is trained to map nominal samples close to a center $\bc$ in feature space. 
For a neural network $\phi_\theta$ with parameters $\theta$, the Deep SVDD objective is given by
\begin{equation}\label{eqn:DSVDD}
  \min_{\theta} \quad \frac{1}{n} \sum_{i=1}^n \left\lVert\NN{\bx_i}-\bc\right\rVert^2.
\end{equation}
In \citet{ruff2020}, an extension of Deep SVDD that incorporates known anomalies is proposed, called \emph{Deep Semi-supervised Anomaly Detection} (Deep SAD). 
Deep SAD trains a network to concentrate nominal data near a center $\bc$ and maps anomalous samples away from that center. 
This is therefore an unsupervised OE approach to AD. 
We here present a principled modification of Deep SAD based on cross-entropy classification that concentrates nominal samples.
We call this method \emph{hypersphere classification} (HSC). 
We found this modification to significantly improve upon the performance of Deep SAD and use it in our experiments as a representative of the unsupervised OE approach to AD.

Let $\cD = \{(\bx_1,y_1), (\bx_2,y_2), \ldots, (\bx_n, y_n)\}$ be a dataset with $\bx_i \in \mR^d$ and $y \in \{0, 1\}$ with $y=1$ denoting nominal and $y=0$ anomalous data points. Let $\phi_\theta : \mR^d \to \mR^r$ be a neural network and $l:\mR^r \to [0,1]$ be a function which maps the output to a probabilistic score. Then, we can formulate the cross-entropy loss as
\begin{equation}\label{eqn:cross}
  - \frac{1}{n} \sum_{i=1}^n y_i \log{l(\NN{\bx_i})} + (1{-}y_i) \log{(1{-}l(\NN{\bx_i}))}.
\end{equation}
For standard binary deep classifiers, $l$ is a linear layer followed by the sigmoid activation and the decision region for the mapped samples $\NN{\bx_1},\ldots,\NN{\bx_n}$ is a half-space $S$. 
In this case the preimage of $S$, $\phi_\theta^{-1}(S)$, is not guaranteed to be compact. 
In order to enforce the preimage of our nominal decision region to be compact, thereby encouraging the mapped nominal data to be concentrated in a way similar to Deep SAD, we propose $l$ to be a radial basis function. 
To construct a spherical decision boundary we let $l(\bz):=\exp{(-\left\|\bz\right\|^2)}$.
In this case, (\ref{eqn:cross}) becomes
\begin{equation*}\label{eqn:radial}
  \frac{1}{n} \sum_{i=1}^n y_i\left\lVert\NN{\bx_i}\right\rVert^2 - (1{-}y_i)\log{(1{-}\exp{({-}\left\lVert\NN{\bx_i}\right\rVert^2)})}.
\end{equation*}
If there are no anomalies, the HSC loss simplifies to $\frac{1}{n}\sum_{i=1}^n\left\lVert\NN{\bx_i}\right\rVert^2.$
For $\bc = 0$, we thus recover Deep SVDD (\ref{eqn:DSVDD}) as a special case.
Similar to Deep SVDD/SAD, we define our anomaly score as $s(\bx) := \left\lVert\NN{\bx}\right\rVert^2$.

Motivated by robust statistics \citep{hampel2005,huber2009} we also considered replacing $l$ with other radial functions where the squared-norm is replaced with a robust alternative. 
We found that using a pseudo-Huber loss \cite{charbonnier1997deterministic} $l(\bz) = \exp\left(-h(\bz) \right)$ that interpolates between squared and absolute value penalization yielded the best results: $h(\bz) = \sqrt{\left\lVert\bz\right\rVert^2+1}-1.$ We include a sensitivity analysis comparing various choices of norms for the hypersphere classifier in Appendix \ref{appx:losses}.

\section{Experiments}
\label{sec:exper}

\paragraph{One vs.~Rest Benchmark} The one vs.~rest evaluation procedure is a ubiquitous benchmark in the deep AD literature as mentioned in the introduction (Section \ref{sec:intro}).
This benchmark constructs AD settings from classification datasets (e.g., MNIST) by considering the ``one'' class (e.g., digit 0) as being nominal and the ``rest'' classes (e.g., digits  1--9) as being anomalous at test time.
In each experiment, we train a model using only the training set of the nominal class as well as random samples from an OE set (e.g., EMNIST-Letters) which is disjoint from the ground-truth anomaly classes of the benchmark. 
We use the same OE auxiliary datasets as suggested in previous works \cite{hendrycks2019deep,hendrycks2019using}.
To evaluate detection performance, we use the common Area Under the ROC curve (AUC) on the one vs.~rest test sets. 
This is repeated over classes and multiple random seeds. 

\textbf{Datasets}\\
\underline{MNIST:} The ten MNIST classes are used as our one vs.~rest classes.
For OE we use the EMNIST-Letters dataset \cite{emnist} which shares no common classes with MNIST.\\[0.2em]
\underline{CIFAR-10:} The ten CIFAR-10 classes are used as our one vs.~rest classes. 
For OE we use 80 Million Tiny Images (80MTI) \cite{torralba200880} with CIFAR-10 and CIFAR-100 images removed. This follows the experimental setup in \citet{hendrycks2019using}.
In one ablation experiment on OE diversity, we alternatively use CIFAR-100 for OE.\\[0.2em]
\underline{ImageNet:} 30 classes from the ImageNet-1K \cite{imagenet} dataset are used as the one vs.~rest classes. These are the classes as proposed in \citet{hendrycks2019using}. For OE we use the ImageNet-22K dataset with the ImageNet-1K removed also following \citet{hendrycks2019using}.

\vspace{-1em}
\paragraph{Methods} We present results from methods that achieve state-of-the-art performance on the one vs.~rest benchmarks.\\[0.2em]
\underline{Unsupervised:} 
We use Deep SVDD \cite{ruff2018deep}, GT \cite{golan2018deep}, GT+ \cite{hendrycks2019using}, and IT \cite{huang2019inverse} as shorthands for the state-of-the-art unsupervised methods.
We also report results of the shallow SVDD \cite{tax2001,scholkopf2001} baseline.\\[0.2em]
\underline{Unsupervised OE:} We implement the HSC from Section \ref{sec:hypsphcla} and Deep SAD \cite{ruff2020} as unsupervised OE methods.
We also report the results from the state-of-the-art unsupervised OE GT+ variant \cite{hendrycks2019using}.\\[0.2em]
\underline{Supervised OE:} We consider a standard binary cross-entropy classifier (BCE).
Moreover, we implement the Focal loss classifier \cite{lin17}, a BCE variant that specifically addresses class imbalance, which was also presented in \citet{hendrycks2019using}.
We indicate the results from \cite{hendrycks2019using} with an asterisk as Focal*. 
We set $\gamma=2$ as recommended in \cite{lin17}.\\[0.2em]
We provide more background on the above methods as well as network architecture and optimization details in Appendices \ref{appx:methods_background} and \ref{appx:architectures} respectively.
Due to space constraints, we report the mean AUC performance over all classes and seeds of the competitive methods in the main paper and report individual results per class and method in Appendix \ref{appx:full_results}.

\vspace{-1em}
\paragraph{Varying the OE Size} For HSC and BCE, we also present extensive experiments showing the performance as the OE training set size is varied on a log scale starting from just $2^0=1$ sample to using the maximal amount of OE data such that no OE sample is seen twice during training. 
If the number of OE samples is less than the OE batch size of 128, we sample with replacement. 
Note that applying data augmentation introduces some variety to the OE set, even in the extreme case of having only $2^0=1$ sample.

\subsection{Results on the CIFAR-10 Benchmark}
\label{subsec:exp_cifar10}

\begin{figure*}[htb]
  \centering \small
  \subfigure[CIFAR-10 (over 10 classes $\times$ 10 seeds)]{\label{fig:cifarvstiny}\includegraphics[width=0.495\textwidth]{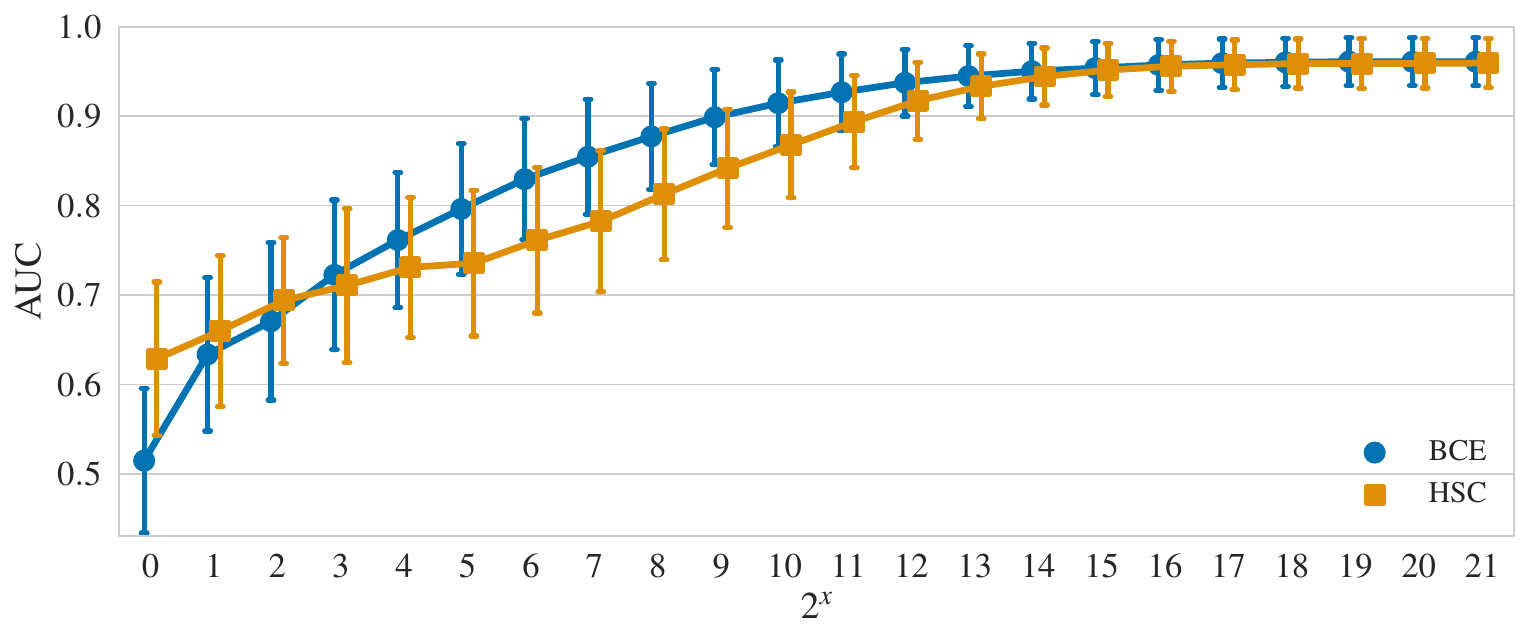}}
  \subfigure[ImageNet-1K (over 30 classes $\times$ 5 seeds)]{\label{fig:imagenet1kvs22k}\includegraphics[width=0.495\textwidth]{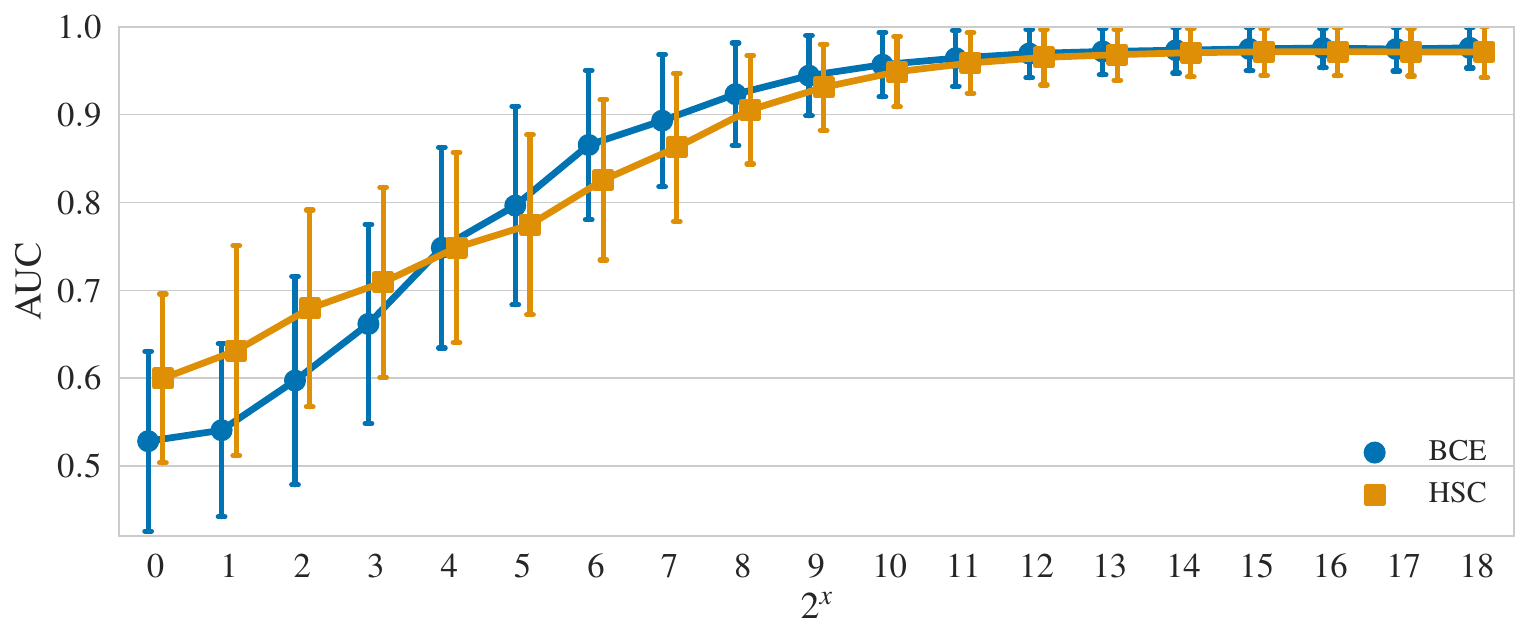}}
  \vspace{-1em}
  \caption{Mean AUC detection performance in \% on the CIFAR-10 and ImageNet-1K one vs.~rest benchmarks when varying the number of 80MTI and ImageNet-22K OE samples respectively.}
  \vspace{-1em}
\end{figure*}

The results on CIFAR-10 are shown in Table \ref{tab:cifar10}.
We observe that the unsupervised OE methods GT+ and HSC yield a comparable detection performance.
Interestingly, the supervised Focal and BCE methods also show state-of-the-art performance, with
BCE attaining the best mean AUC overall. 
To understand the informativeness of OE for unsupervised and supervised approaches, we compare the performance of HSC and BCE while varying the OE set size. 
The results in Figure \ref{fig:cifarvstiny} demonstrate that surprisingly few OE samples already yield a very competitive detection performance.

\begin{table}[th]
  \vspace{-1.2em}
  \caption{Mean AUC detection performance in \% (over 10 classes and 10 seeds) on the CIFAR-10 one vs.~rest benchmark using 80MTI as OE (* indicates results from the literature).}
  \label{tab:cifar10}
  \centering 
  \resizebox{\columnwidth}{!}{\begin{tabular}{ccccccccccc}
    \toprule
    & \multicolumn{2}{c|}{Unsupervised} & \multicolumn{3}{c|}{Unsupervised OE} & \multicolumn{3}{c}{Supervised OE}\\
    & DSVDD*    & \multicolumn{1}{c|}{GT+*}     & GT+*      & DSAD  & \multicolumn{1}{c|}{HSC}   & Focal*    & Focal & BCE\\
    \midrule
    & 64.8      & 90.1                          & 95.6      & 94.5      & 95.9  & 87.3      & 95.8  & \bf 96.1\\
    \bottomrule
\end{tabular}
}
\end{table}

\subsection{Results on the ImageNet Benchmark}
\label{subsec:exp_imagenet}

The results on ImageNet are given in Table \ref{tab:imagenet}. 
GT+* and Focal* are the results from \citet{hendrycks2019using}, where this benchmark was introduced. 
Deep SAD, HSC, Focal, and BCE all outperform the current state of the art (GT+*) by a surprisingly wide margin.
We are unsure as to why the Focal* results from \citet{hendrycks2019using} are so poor since their experimental code is not public. 
We found performance to be insensitive to the choice of $\gamma$ in the Focal loss (see Appendix \ref{appx:focal}). 
Not balancing the number of nominal and OE samples in each batch may be one explanation. 
We also compare the performance of HSC and BCE while varying the OE set size. 
The results are in Figure \ref{fig:imagenet1kvs22k}. 
Again we see that there is a transition from HSC to BCE performing best, which can be understood as a transition from unsupervised OE to supervised OE. Remarkably, classification beats previous methods with only 64 OE samples.

\begin{table}[ht]
  \vspace{-1.2em}
  \caption{Mean AUC detection performance in \% (over 30 classes and 10 seeds) on the ImageNet-1K one vs.~rest benchmark using ImageNet-22K (with the 1K classes removed) as OE (* indicates results from the literature).}
  \label{tab:imagenet}
  \centering
  \footnotesize
  \begin{tabular}{lcccccc}
    \toprule
    & \multicolumn{3}{c|}{Unsupervised OE} & \multicolumn{3}{c}{Supervised OE}\\
    & GT+* & DSAD  & \multicolumn{1}{c|}{HSC}   & Focal*    & Focal & BCE\\
    \midrule
    & 85.7  & 96.7      & 97.3                          & 56.1      & 97.5  & \bf 97.7\\
    \bottomrule
\end{tabular}
\end{table}

In addition to studying the effect of OE set size, we also evaluate the effect of OE data diversity on detection performance. 
For this, we vary the number of anomaly classes that comprise the OE set. 
As expected, we find that performance overall increases with OE data diversity, but interestingly drawing OE samples from just one class (which is not present as anomaly at test time!) yields surprisingly good performance. 
We provide the full results in Appendix \ref{appx:oe_diversity}.

\subsection{Removing Multiscale Information}
\label{ssec:blurring}
To investigate the hypothesis that the exceptional informativeness of OE samples is due to the multiscale structure of natural images, we perform an experiment in which we removes small scale features from the OE data.
For this we compare the performance of HSC and BCE on the ImageNet one vs.~rest task while increasingly blurring the OE samples with a Gaussian filter. 
The blurring gradually removes the small scale (high frequency) features from the OE data.

\begin{figure}[htb]
  \centering\small
  \subfigure[ImageNet-1K]{\includegraphics[width=0.495\columnwidth]{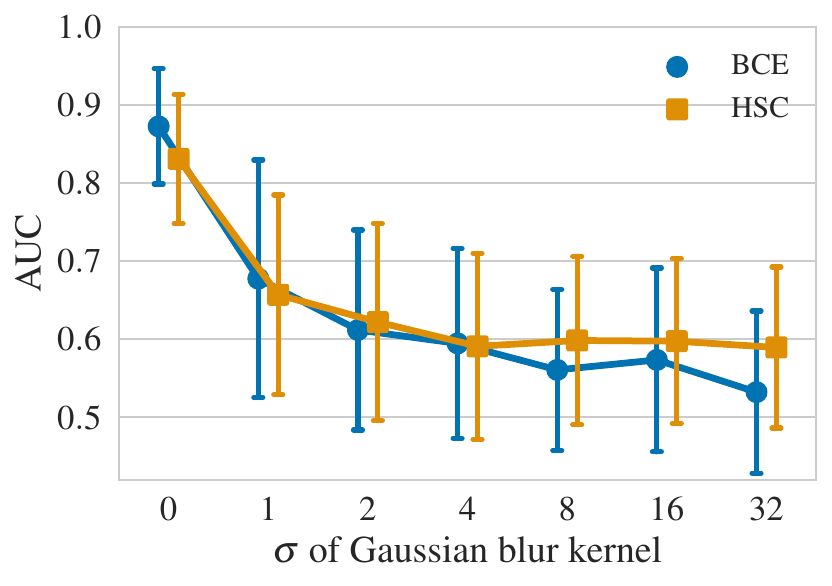}}
  \subfigure[Degrees of blurring]{\includegraphics[width=0.495\columnwidth]{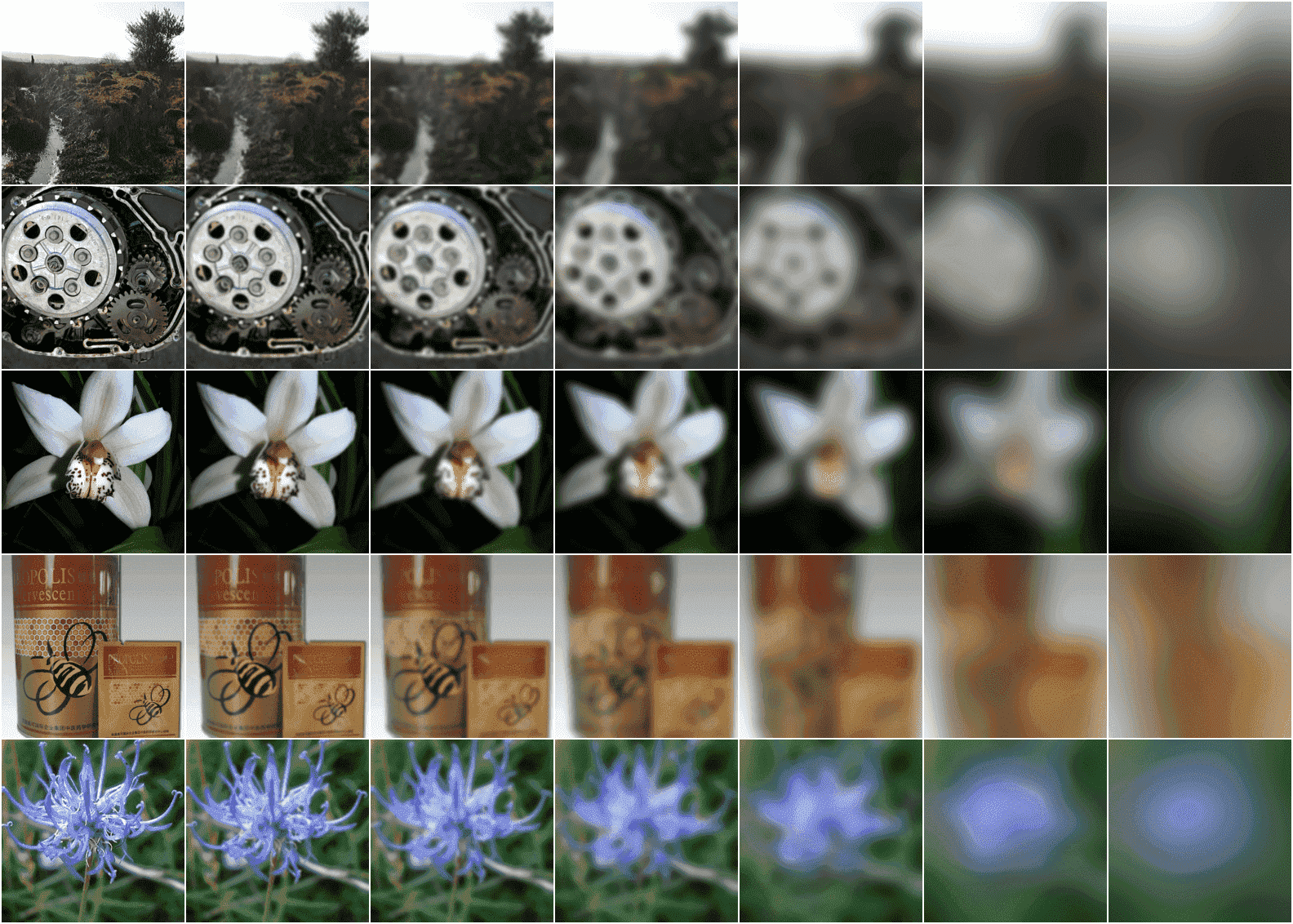}}
  \vspace{-1em}
  \caption{Mean AUC detection performance in \% (over 30 classes and 5 seeds) on ImageNet-1K for $2^6 = 64$ OE samples when increasingly blurring the OE images with a Gaussian kernel (see (a)). A visual example of the Gaussian blurring is shown in (b). The sharp decline in AUC suggests the exceptional informativeness of OE for images is due to their multiscale nature.}
  \label{fig:blur_imagenet}
  \vspace{-2em}
\end{figure}

In Figure \ref{fig:blur_imagenet}, we see that detection performance quickly drops with even a small amount of blurring. 
With sufficient blurring, the unsupervised OE method (HSC) performs better. 
We provide results for MNIST and CIFAR-10 in Appendix \ref{appx:multiscale}, where we observe a similar decrease in performance for CIFAR-10. 
For MNIST, however, HSC outperforms BCE at any degree of blurring and retains a good performance.
This suggests that their are two regimes in deep image AD: (i) on high-dimensional multiscale images, OE samples are highly informative and standard as well as hypersphere classification perform similarly well, and (ii) on low-dimensional images (e.g., MNIST), OE samples are also informative, but hypersphere performs better than standard classification, which reflects the classic intuition that anomaly detection requires a compact model.

\section{Conclusion}
\label{sec:disc}
We have shown that deep AD on images displays a phenomenon which is quite different from what is expected in classic AD. 
Compared to classic AD, a few example outliers are exceptionally informative on common image AD benchmarks. 
Furthermore, we have shown that this phenomenon is tied to the multiscale nature of natural images.
Finally, please note that we do not claim that a supervised OE approach is \emph{the} solution to AD in general, or that there is no utility for unsupervised OE. However, our results suggest that it may be time for the community to move to more challenging benchmarks (e.g., MVTec-AD \cite{Bergmann_2019_CVPR}) to gauge the significance of deep AD works. 

\bibliography{references}
\bibliographystyle{icml2021}

%
%
%

\clearpage
\appendix

\textbf{\LARGE Appendix}

\section{Methods Background}
\label{appx:methods_background}

While there exist many shallow methods for AD, it has been observed that these methods perform poorly for high-dimensional data \cite{huang2006large,kriegel2008angle,erfani2015r1svm,erfani2016high}. 
Deep approaches have been proposed to fill this gap.
The most common approaches to deep AD employ autoencoders trained on nominal data, where samples not reconstructed well are deemed anomalous \cite{hawkins2002outlier,sakurada2014anomaly,chen2017outlier,zhou2017,huang2019inverse,nguyen2019,kim20}. 
Deep generative models have also been used to detect anomalies via a variety of methods \cite{schlegl2017unsupervised,deecke2018image,zenati2018efficient,schlegl2019}, yet their effectiveness has been called into question \cite{nalisnick2019}.

Another recent avenue of research on deep AD uses self-supervision on images \cite{gidaris2018unsupervised,golan2018deep,mariusnips19,hendrycks2019using}. 
In \citet{golan2018deep}. the authors use a composition of image transformations---including identity, rotations, flips, and translations---to create a self-supervised classification task. 
Every training sample is transformed using each of these transformations and a label is assigned to every transformed sample corresponding to the applied transformation. 
This creates a multi-class classification task for predicting image transformations. 
A network is then trained on this data to predict the applied transformation. 
For a test sample these transformations and network outputs are utilized to determine an anomaly score.

To our knowledge, one of the best performing AD method on image data is the self-supervised approach from \citet{hendrycks2019using} which extends \citet{golan2018deep}'s method by using three classification heads to predict a combination of three types of transformations. 
They train their network on transformed nominal data as was done in \citet{golan2018deep}. 
On a test sample the network's certainty (how close to 1) on predicting correct transformations is used as an anomaly score, with certainty being a signifier that a sample is not anomalous. 
Essentially this assumes the network predictions to be less concentrated on the correct output for unfamiliar looking data. 
In that paper, the authors validate their method on the CIFAR-10 and ImageNet one vs.~rest one-class learning benchmarks.

\subsection{Auxiliary Data and the State of the Art for Deep AD on Images}
Many deep learning methods have been proposed to incorporate the large amount of unorganized data that is now easily accessible on the web. 
A common way to use this data is via unsupervised or self-supervised learning. In the realm of NLP, word2vec \cite{mikolov2013efficient} and more recent language models such as ELMo \cite{peters-etal-2018-deep} or BERT \cite{devlin-etal-2019-bert} are now standard and responsible for significant improvements on various NLP tasks. 
For image tasks, using an auxiliary dataset for pretraining has been found to be effective \cite{zeiler2014visualizing}. 
Moreover many deep semi-supervised methods have been introduced to enhance classification performance via incorporating unlabeled data into training \cite{kingma2014b,rasmus2015,odena2016,oliver2018}.

The use of a large unstructured corpus of image data to improve deep AD was first proposed in \citet{hendrycks2019deep}, where they call the general use of such data \emph{outlier exposure} (OE).
In \citet{hendrycks2019using} the authors use OE to further improve existing self-supervised classification methods. 
They do this by training the aforementioned self-supervised methods to predict the uniform distribution for all transforms on OE samples, while leaving training on nominal samples unchanged. 
To our knowledge the AD method with OE presented in \citet{hendrycks2019using} is the current state of the art on the CIFAR-10 and ImageNet image anomaly detection benchmarks, outperforming previous unsupervised AD methods with or without OE.

\subsection{Anomaly Detection as Binary Classification}
\label{sec:classad}
Traditionally AD is understood as the problem of estimating the support (or level sets of the support) of the nominal data-generating distribution. This is also known as \emph{density level set estimation} \cite{polonik1995measuring,tsybakov1997,ruff2021}.
The motivation for density level set estimation is the common assumption that nominal data is concentrated whereas anomalies are not concentrated \cite{scholkopf2002}.
\citet{steinwart2005classification} remark that the problem of density level set estimation can be interpreted as binary classification between the nominal and an anomalous distribution. 
Many of the classic AD methods (e.g., KDE or OC-SVM) implicitly assume the anomalies to follow a uniform, i.e.~they make an uninformed prior assumption on the anomalous distribution \cite{steinwart2005classification}. 
These methods, as well as a binary classifier trained to discriminate between nominal samples and uniform noise, are asymptotically consistent density level set estimators \cite{steinwart2005classification,vert2006}. 
Obviously it is better to directly estimate the level set rather than introducing the auxiliary task of classifying against uniform noise. 
Such a classification approach is particularly ineffective and inefficient in high dimensions since it would require massive amounts of noise samples to properly fill the sample space.
As demonstrated through various experiments, however, we find that this intuition does not seem to extend to a deep approach to image anomaly detection when the anomalous examples are natural images.

\section{Diversity of the Outlier Exposure Data}
\label{appx:oe_diversity}

Here we evaluate how data diversity influences detection performance for unsupervised and supervised OE, again comparing HSC to BCE.
For this purpose, instead of 80MTI, we now use CIFAR-100 as OE varying the number of anomaly classes available for the CIFAR-10 benchmark. We further evaluate the methods on the MNIST one vs.~rest benchmark where EMNIST-Letters is used as the OE dataset. For both experiments, the OE data is varied by choosing $k$ classes at random for each random seed and using the union of these classes as the OE dataset.

\begin{figure}[htb]
  \centering \small
  \subfigure[MNIST]{\label{fig:diversity_mnist}\includegraphics[width=0.495\columnwidth]{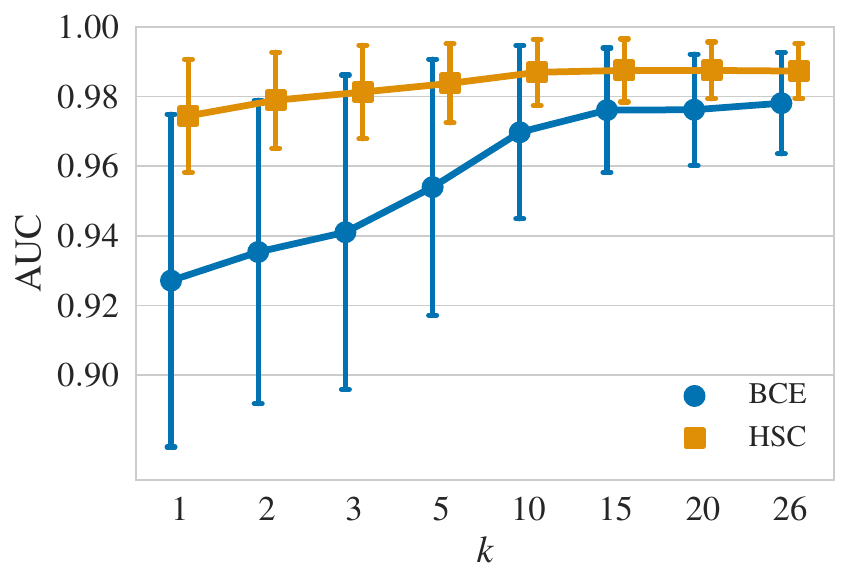}}
  \subfigure[CIFAR-10]{\label{fig:diversity_cifar10}\includegraphics[width=0.495\columnwidth]{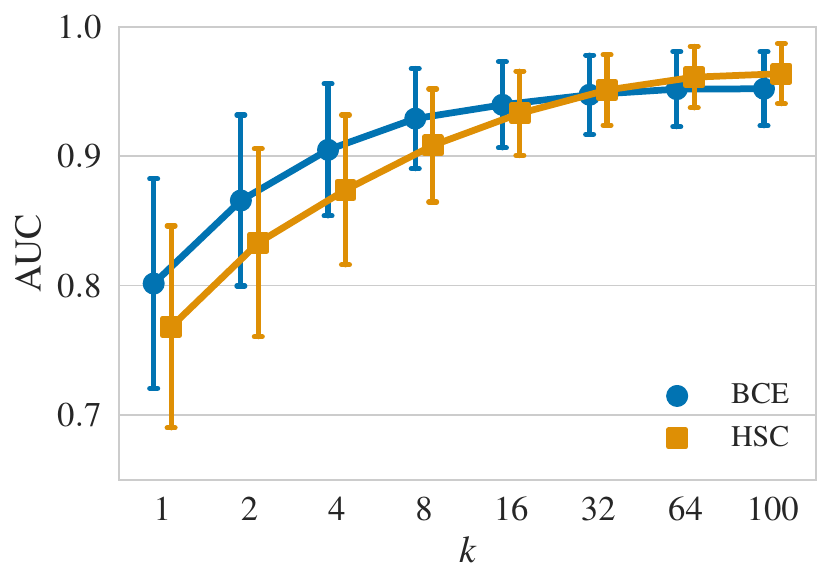}}
  \caption{Mean AUC detection performance in \% (over 10 classes with 10 seeds per class) on the MNIST with EMNIST-Letters OE (a) and CIFAR-10 with CIFAR-100 OE (b) one vs.~rest benchmarks when varying the number of $k$ classes that comprise the OE dataset.}
  \label{fig:diversity}
\end{figure}

The results are presented in Figure \ref{fig:diversity}. 
As expected, the performance increases with the diversity of the OE dataset.
Interestingly, drawing OE samples from just $k=1$ class, i.e.~binary classification between the nominal class and a single OE class (which is not present as an anomaly class at test time!) already yields good detection performance on the CIFAR-10 benchmark. For example training a standard classification network to discern between automobiles and beavers performs competitively as an automobile anomaly detector, even when no beavers are present as anomalies during test time.

For the MNIST experiment we see that HSC outperforms BCE for any number of classes. 
We hypothesize that this is due to the lack of multiscale spatial structure in the MNIST and EMNIST datasets. This intuition is consistent with the classic understanding of AD mentioned in the introduction of the main paper (Section \ref{sec:intro}).

\section{Removing Multiscale Information on MNIST and CIFAR-10}
\label{appx:multiscale}

\begin{figure}[htb]
  \centering\small
  \subfigure[MNIST]{\includegraphics[width=0.495\columnwidth]{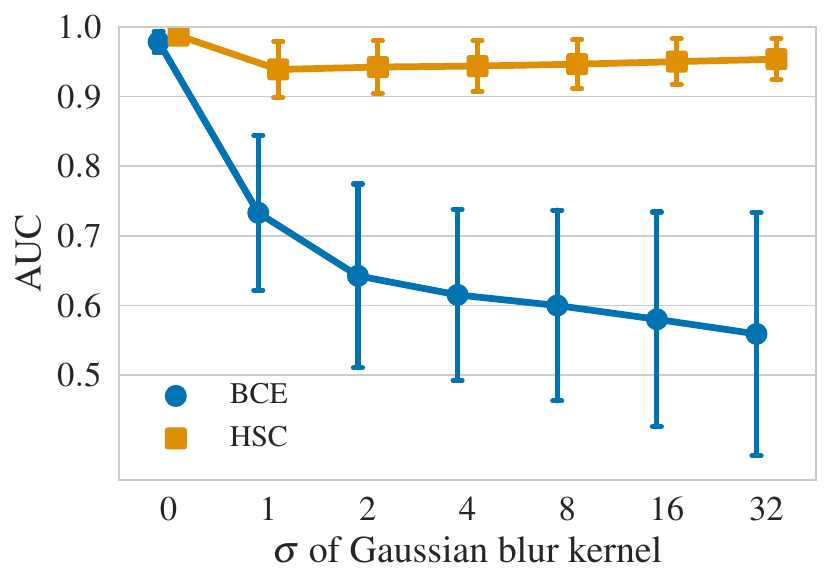}}
  \subfigure[Degrees of blurring]{\includegraphics[width=0.495\columnwidth]{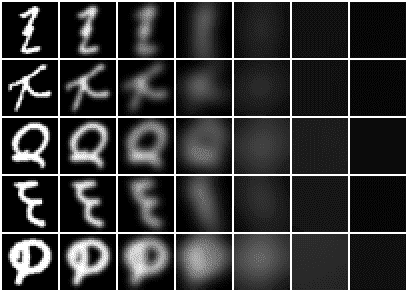}}
  \caption{Mean AUC detection performance in \% (over 10 classes and 10 seeds) on the MNIST one vs.~rest benchmark when increasingly blurring the EMNIST OE images with a Gaussian kernel (see (a)). 
  A visual example of the Gaussian blurring for some EMNIST OE samples is shown in (b).
  We see that HSC clearly outperforms BCE on MNIST, a dataset which has essentially no multiscale information.}
  \label{fig:blur_mnist}
\end{figure}

\begin{figure}[htb]
  \centering\small
  \subfigure[CIFAR-10]{\includegraphics[width=0.495\columnwidth]{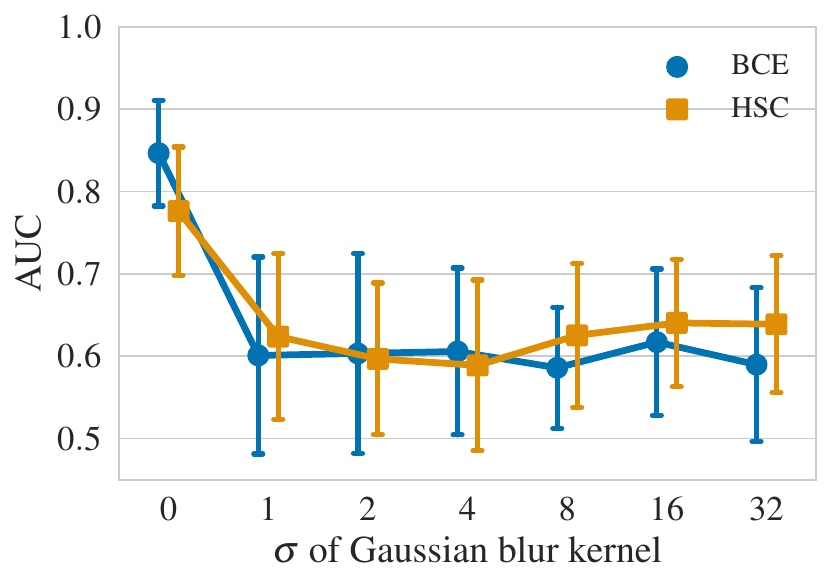}}
  \subfigure[Degrees of blurring]{\includegraphics[width=0.495\columnwidth]{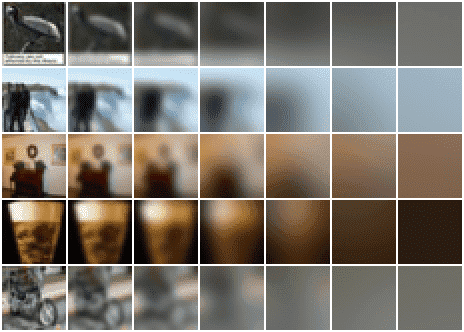}}
  \caption{Mean AUC detection performance in \% (over 10 classes and 10 seeds) on the CIFAR-10 one vs.~rest benchmark for $2^7 = 128$ OE samples when increasingly blurring the 80MTI OE images with a Gaussian kernel (see (a)). A visual example of the Gaussian blurring for some 80MTI OE samples is shown in (b). The rapid decrease in AUC on CIFAR-10 again suggests that the informativeness of OE on images is due to the multiscale structure of images.}
  \label{fig:blur_cifar10}
\end{figure}

Here we present the experimental results on removing multiscale information (see Section \ref{ssec:blurring}) for MNIST and CIFAR-10. 
The results are given in Figures \ref{fig:blur_mnist} and \ref{fig:blur_cifar10} for MNIST and CIFAR-10 respectively. 
For CIFAR-10, we can observe a similar decrease in performance as observed for ImageNet (see Figure \ref{fig:blur_imagenet}). 
For MNIST, however, HSC outperforms BCE at any degree of blurring and retains a good performance.

\section{Hypersphere Classifier Sensitivity Analysis}
\label{appx:losses}

Here we show results for the Hypersphere Classifier (HSC) we introduced in Section 3 when varying the radial function $l(\bz) = \exp\left(-h(\bz) \right)$.
For this, we run the CIFAR-10 one vs.~rest benchmark with 80MTI OE experiment as presented in Table \ref{tab:cifar10} in the main paper for different functions $h:\mR^r \to [0, \infty), \bz \mapsto h(\bz)$.
We also alter training to be with or without data augmentation in these experiments.
The results are presented in Table \ref{tab:loss_sensitivity}.
We see that data augmentation leads to an improvement in performance even in this case where we have the full 80MTI dataset as OE.
HSC shows the overall best performance with data augmentation and using the robust Pseudo-Huber loss $h(\bz) = \sqrt{\left\lVert\bz\right\rVert^2+1}-1$.

\begin{table}[ht]
  \caption{Mean AUC detection performance in \% (over 10 seeds) on the CIFAR-10 one vs.~rest benchmark using 80MTI as OE for different choices of $h(\bz)$ in the radial function $l$ of the HSC.}
  \label{tab:loss_sensitivity}
  \centering\small
  \begin{tabular}{lcccccccc}
    \toprule
    Data augment.\       & $\left\|\bz\right\|_1$ & $\left\|\bz\right\|_2$ & $\left\|\bz\right\|_2^2$ & $\sqrt{\left\lVert\bz\right\rVert^2+1}-1$ \\
    \midrule
    w/o                     & 90.6 & 92.3 & 89.1 & 91.8 \\
    w/                      & 92.5 & 94.1 & 94.5 & 96.1 \\
    \bottomrule
\end{tabular}
\end{table}

\section{Focal Loss With Varying \texorpdfstring{$\gamma$}{TEXT}} 
\label{appx:focal}

\begin{figure}[htb]
  \centering \small
  \subfigure[CIFAR-10]{\includegraphics[width=0.495\columnwidth]{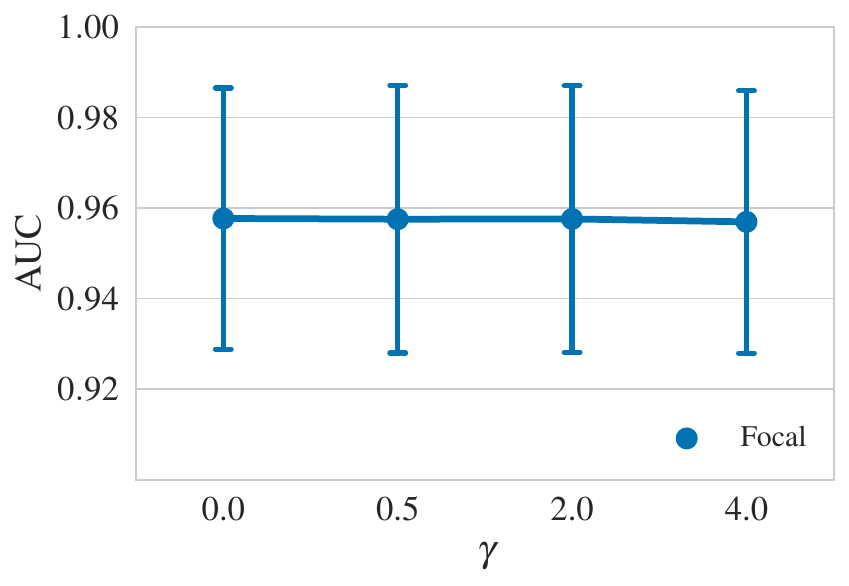}}
  \subfigure[ImageNet-1K]{\includegraphics[width=0.495\columnwidth]{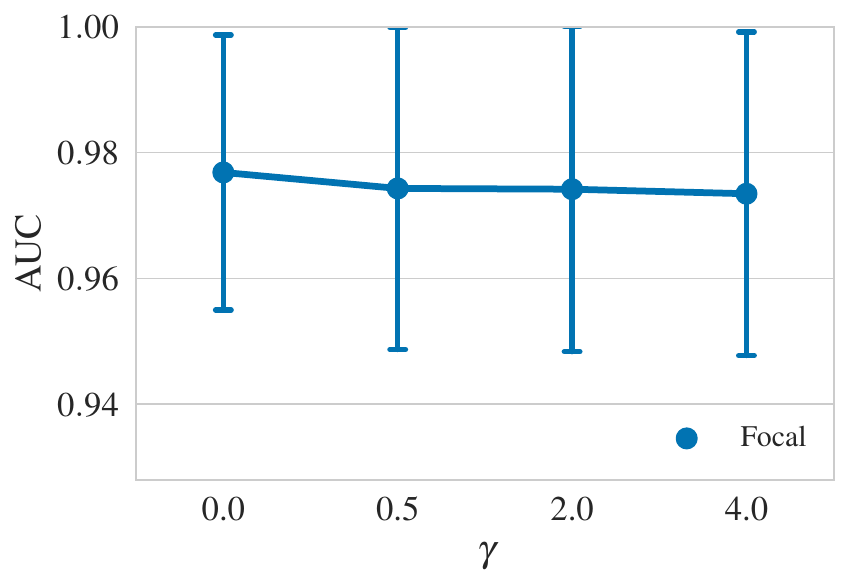}}
  \caption{Focal loss detection performance in mean AUC in \% when varying $\gamma$ on the CIFAR-10 with 80MTI OE (a) and ImageNet-1K with ImageNet-22K OE (b) one vs.~rest benchmarks.}
  \label{fig:focal_sensitivity}
\end{figure}

Here we include results showing how mean AUC detection performance changes with $\gamma$ on the Focal loss.
Since we balance every batch to contain 128 nominal and 128 OE samples during training, we set the weighting factor $\alpha$ to be $\alpha=0.5$ \cite{lin17}.
Again note that $\gamma = 0$ corresponds to standard binary cross entropy.
Figure \ref{fig:focal_sensitivity} shows that mean AUC performance is insensitive to the choice of $\gamma$ on the CIFAR-10 and ImageNet-1K one vs.~rest benchmarks.

\section{Network Architectures and Optimization}
\label{appx:architectures}

We always use the same underlying network $\phi_\theta$ in each experimental setting for our HSC, Deep SAD, Focal, and BCE implementations to control architectural effects. 
For Focal and BCE, the output of the network $\phi_\theta$ is followed by a linear layer with sigmoid activation. 
For the experiments on MNIST and CIFAR-10 we use standard LeNet-style networks having two and three convolutional layers followed by two fully connected layers respectively. 
We use batch normalization \cite{ioffe2015} and (leaky) ReLU activations in these networks. 
For our experiments on ImageNet we use the same WideResNet \cite{zagoruyko2016wide} as \citet{hendrycks2019using}, which has ResNet-18 as its architectural backbone. 
We use Adam \citep{kingma2014} for optimization and balance every batch to contain 128 nominal and 128 OE samples during training. 
For data augmentation, we use standard color jitter, random cropping, horizontal flipping, and Gaussian pixel noise. 
We provide further dataset-specific details below.\footnotetext{Complete code also available at: \url{https://github.com/lukasruff/Classification-AD}}

\subsection{MNIST and CIFAR-10}
On MNIST and CIFAR-10, we use LeNet-style networks having two and three convolutional layers and two fully connected layers respectively. Each convolutional layer is followed by batch normalization, a leaky ReLU activation, and max-pooling. The first fully connected layer is followed by batch normalization, and a leaky ReLU activation, while the last layer is just a linear transformation. The number of kernels in the convolutional layers are, from first to last: 16-32 (MNIST), and 32-64-128 (CIFAR-10). The fully connected layers have 64-32 (MNIST), and 512-256 (CIFAR-10) units respectively.
We use Adam \citep{kingma2014} for optimization and balance every batch to contain
128 nominal and 128 OE samples during training.
We train for 150 (MNIST) and 200 (CIFAR-10) epochs starting with a learning rate of $\eta=0.001$ and have learning rate milestones at 50, 100 (MNIST), and 100, 150 (CIFAR-10) epochs. The learning rate is reduced by a factor of 10 at every milestone.

\subsection{ImageNet}
On ImageNet, we use exactly the same WideResNet \cite{zagoruyko2016wide} as was used in \citet{hendrycks2019using}, which has a ResNet-18 as architectural backbone.
We use Adam \citep{kingma2014} for optimization and balance every batch to contain
128 nominal and 128 OE samples during training.
We train for 150 epochs starting with a learning rate of $\eta=0.001$ and milestones at epochs 100 and 125. The learning rate is reduced by a factor of 10 at every milestone.

\section{Results on Individual Classes}
\label{appx:full_results}

\begin{figure*}[th]
\centering
\caption{Mean AUC detection performance in \% (over 10 seeds) for all classes of the CIFAR-10 one vs.~rest benchmark from Section \ref{subsec:exp_cifar10} when varying the number of 80MTI OE samples. These plots correspond to Figure \ref{fig:cifarvstiny}, but here we report the results for all individual classes.}
\subfigure[Class: airplane]{\label{fig:cifarvstiny0}\includegraphics[width=0.49\linewidth]{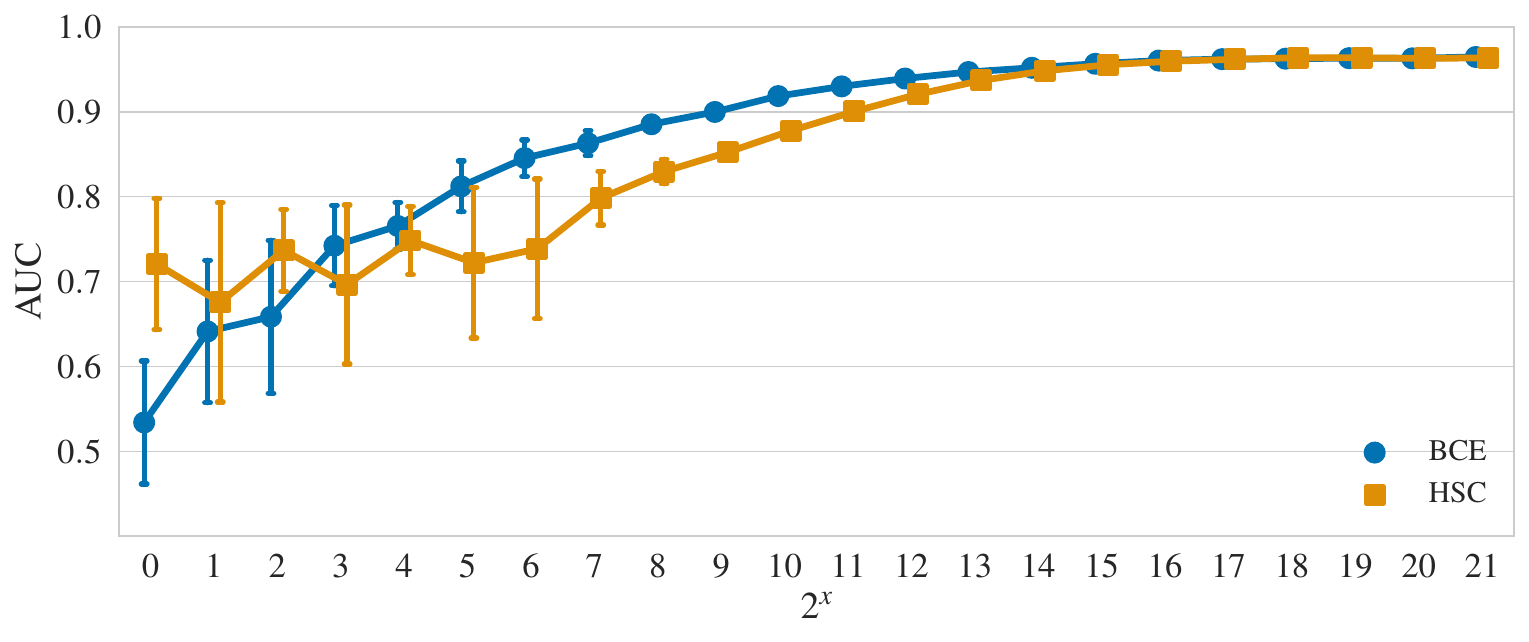}}
\subfigure[Class: automobile]{\label{fig:cifarvstiny1}\includegraphics[width=0.49\linewidth]{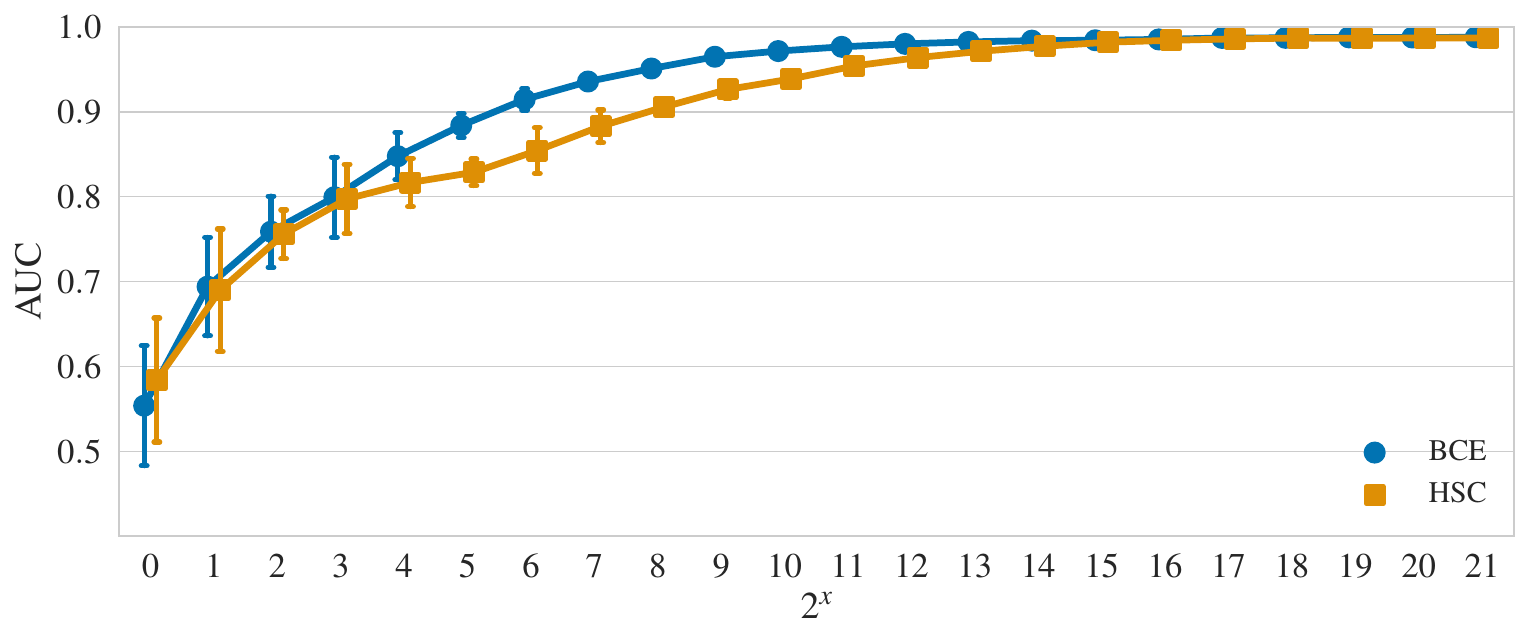}}
\subfigure[Class: bird]{\label{fig:cifarvstiny2}\includegraphics[width=0.49\linewidth]{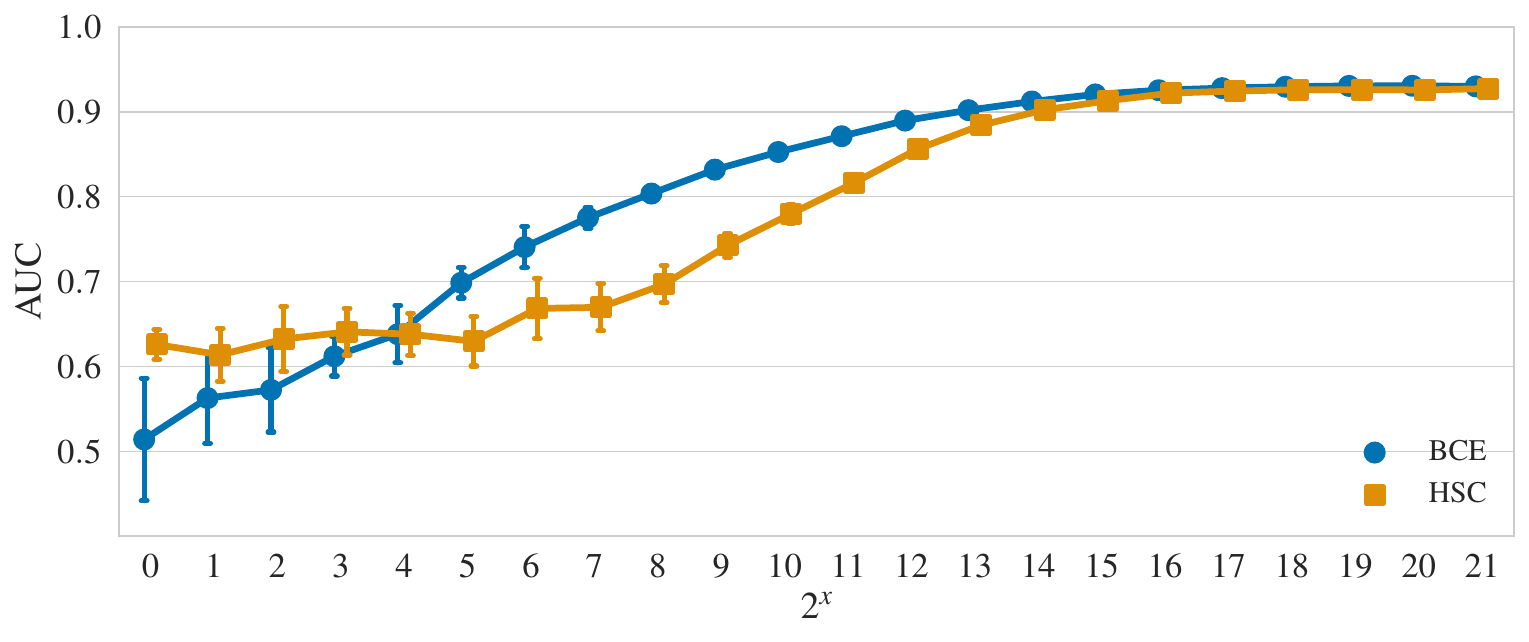}}
\subfigure[Class: cat]{\label{fig:cifarvstiny3}\includegraphics[width=0.49\linewidth]{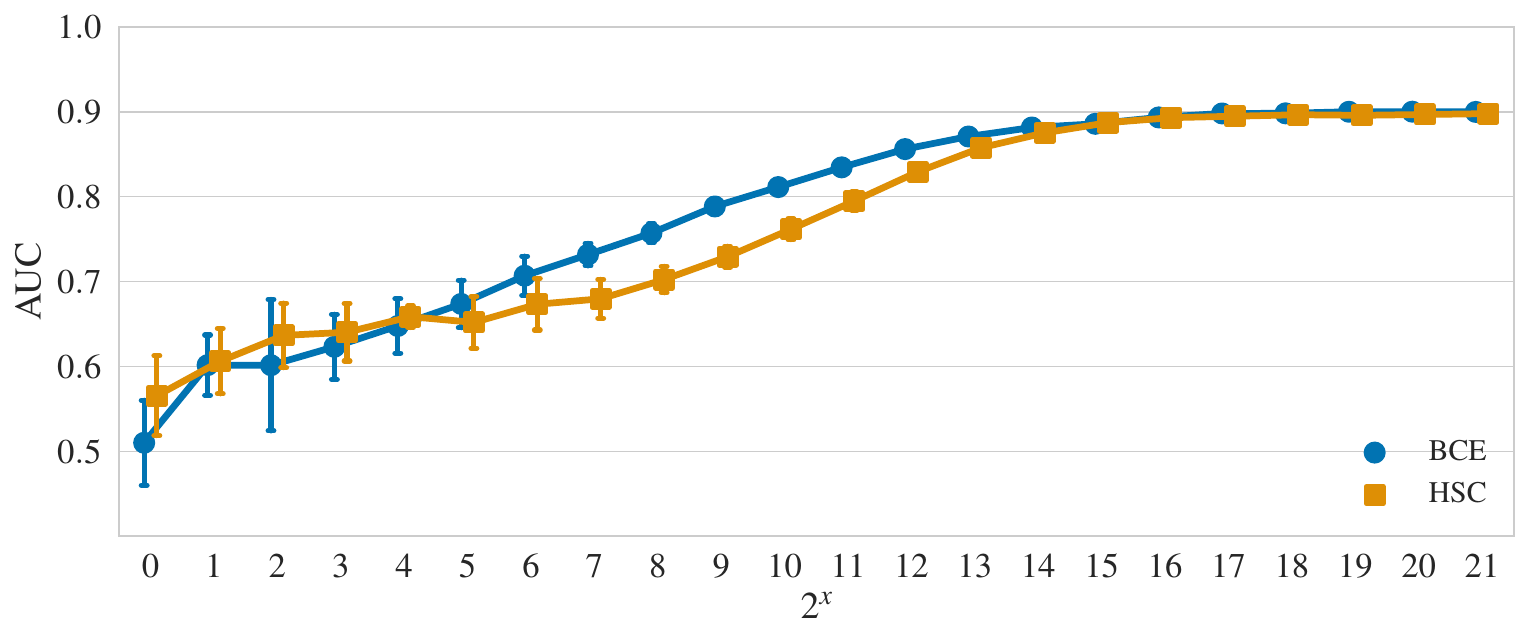}}
\subfigure[Class: deer]{\label{fig:cifarvstiny4}\includegraphics[width=0.49\linewidth]{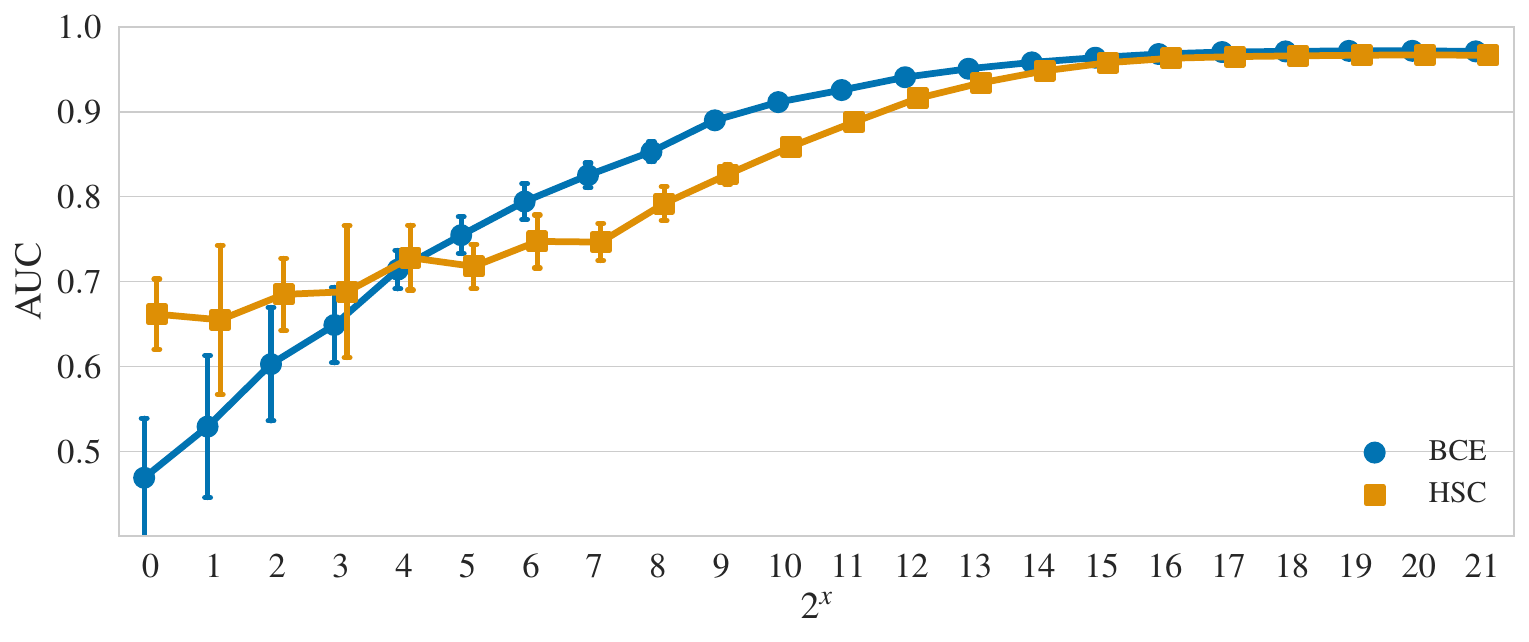}}
\subfigure[Class: dog]{\label{fig:cifarvstiny5}\includegraphics[width=0.49\linewidth]{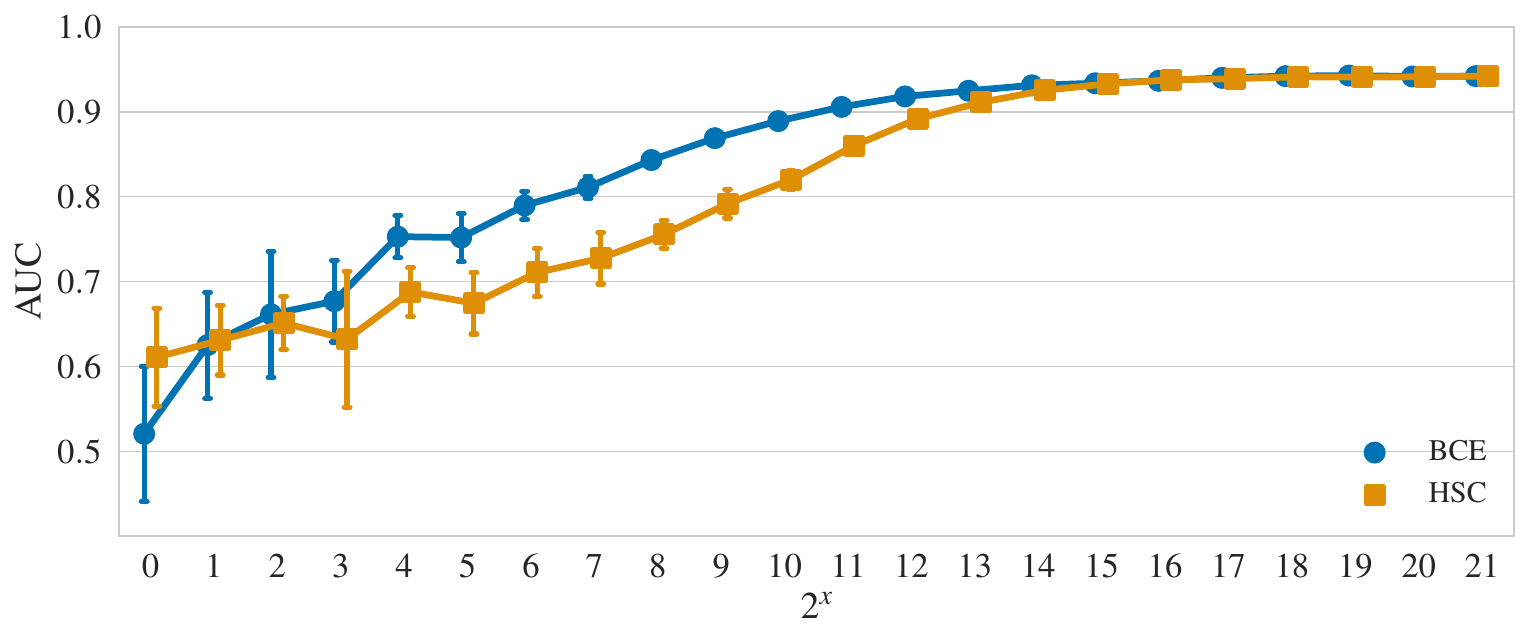}}
\subfigure[Class: frog]{\label{fig:cifarvstiny6}\includegraphics[width=0.49\linewidth]{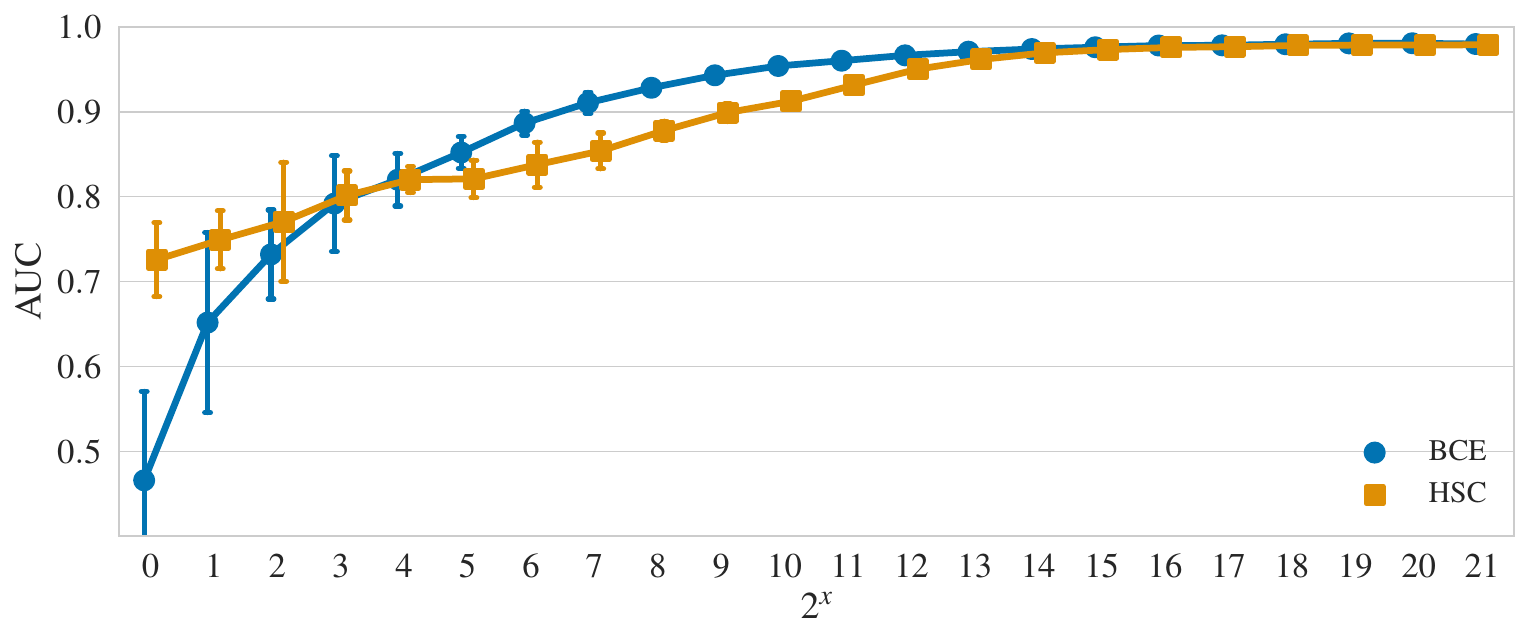}}
\subfigure[Class: horse]{\label{fig:cifarvstiny7}\includegraphics[width=0.49\linewidth]{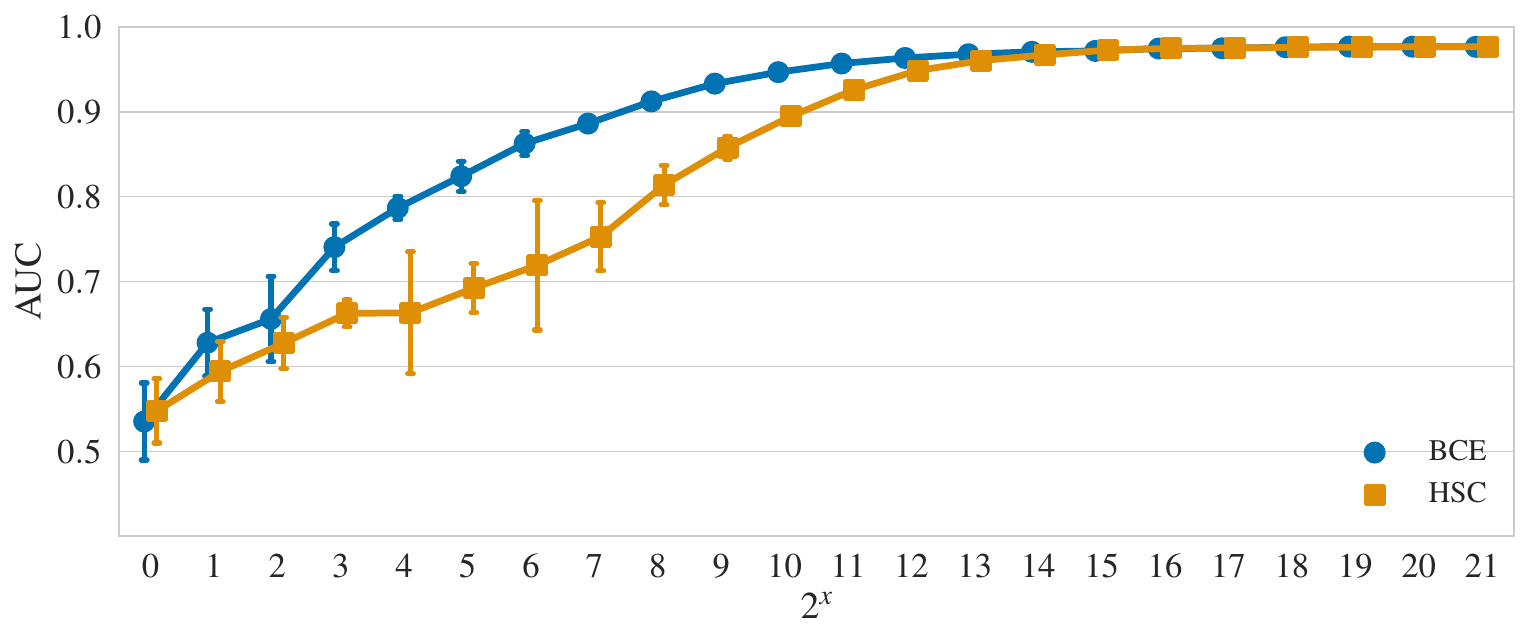}}
\subfigure[Class: ship]{\label{fig:cifarvstiny8}\includegraphics[width=0.49\linewidth]{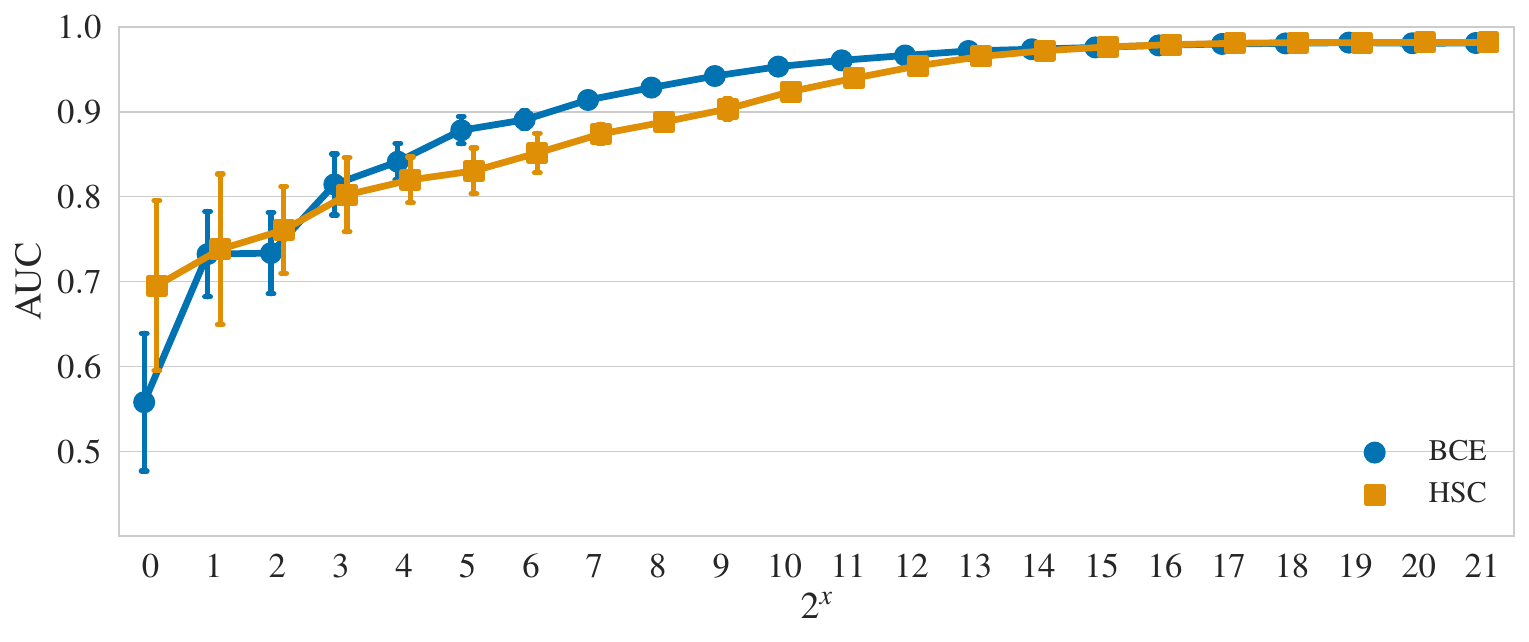}}
\subfigure[Class: truck]{\label{fig:cifarvstiny9}\includegraphics[width=0.49\linewidth]{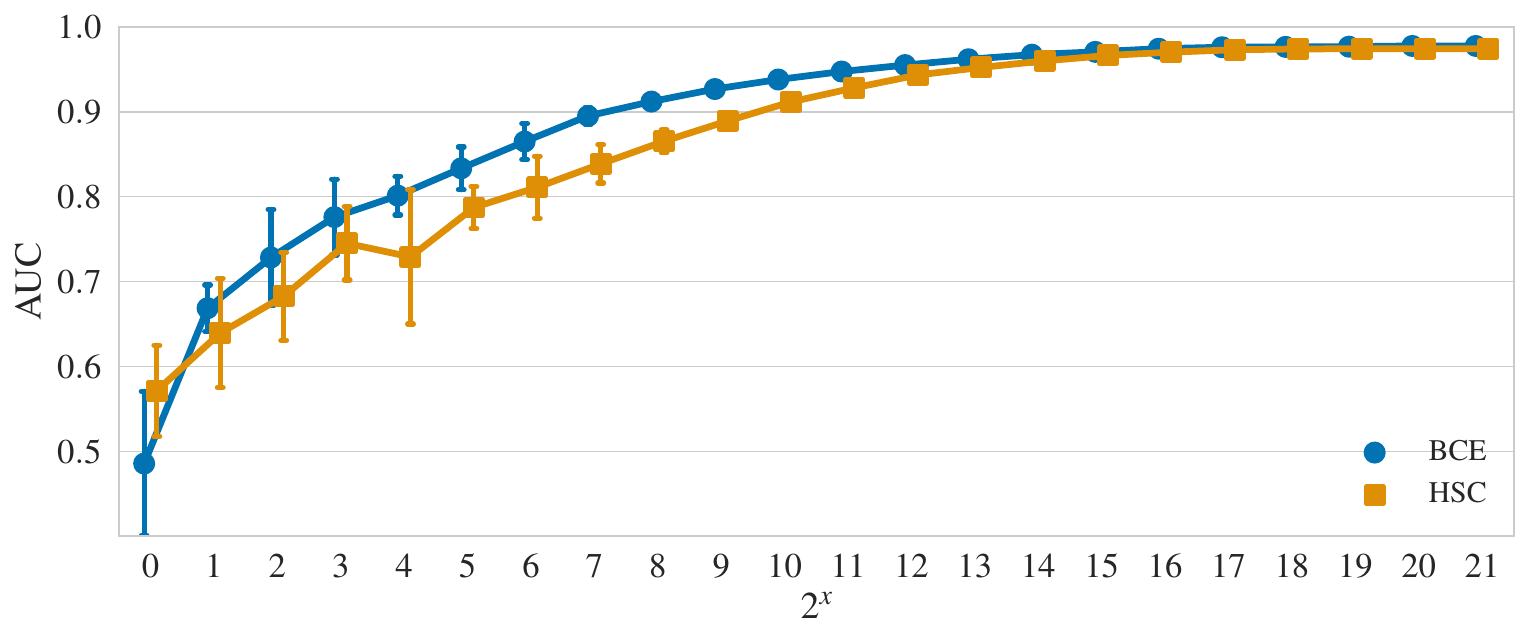}}
\label{fig:cifarvstiny_classes}
\end{figure*}

\begin{figure*}[th]
\centering
\caption{Mean AUC detection performance in \% (over 5 seeds) for all classes of the ImageNet-1K one vs.~rest benchmark from Section \ref{subsec:exp_imagenet} when varying the number of ImageNet-22K OE samples. These plots correspond to Figure \ref{fig:imagenet1kvs22k}, but here we report the results for all individual classes (from class 1 (acorn) to class 15 (hourglass)).}
\subfigure[Class: acorn]{\includegraphics[width=0.329\linewidth]{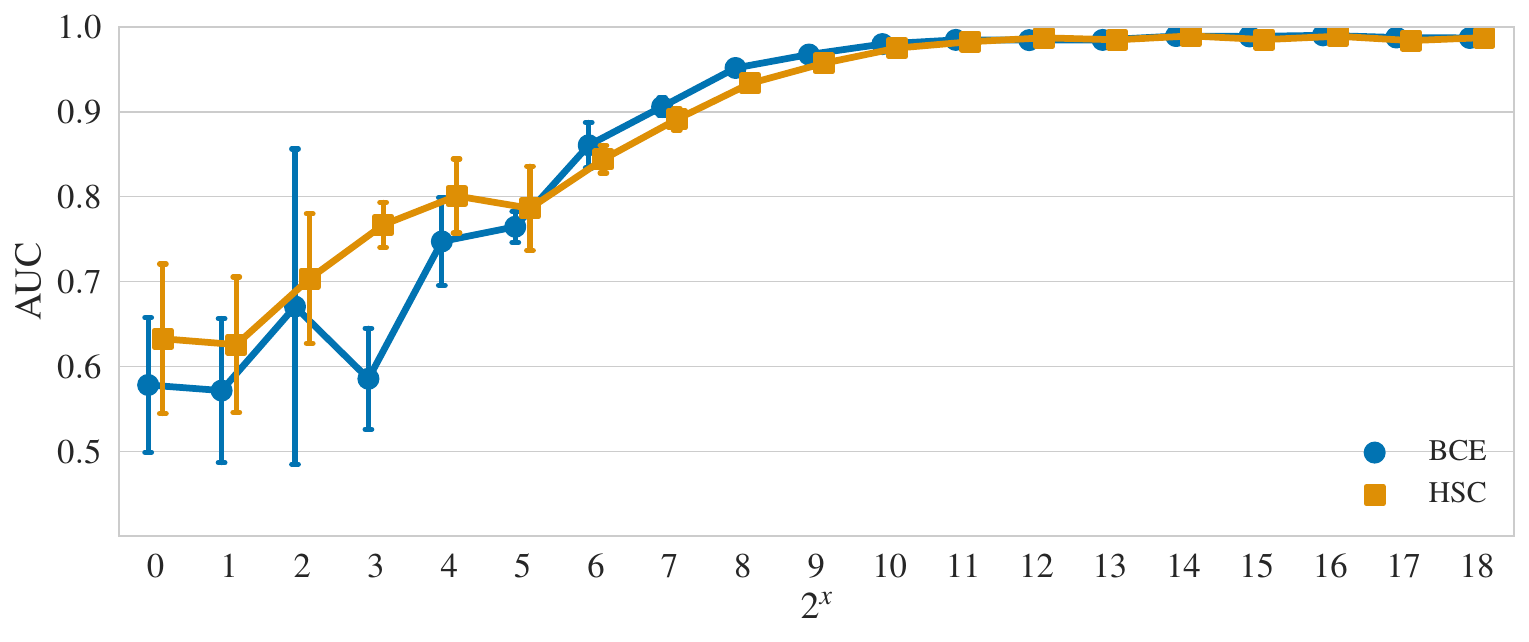}}
\subfigure[Class: airliner]{\includegraphics[width=0.329\linewidth]{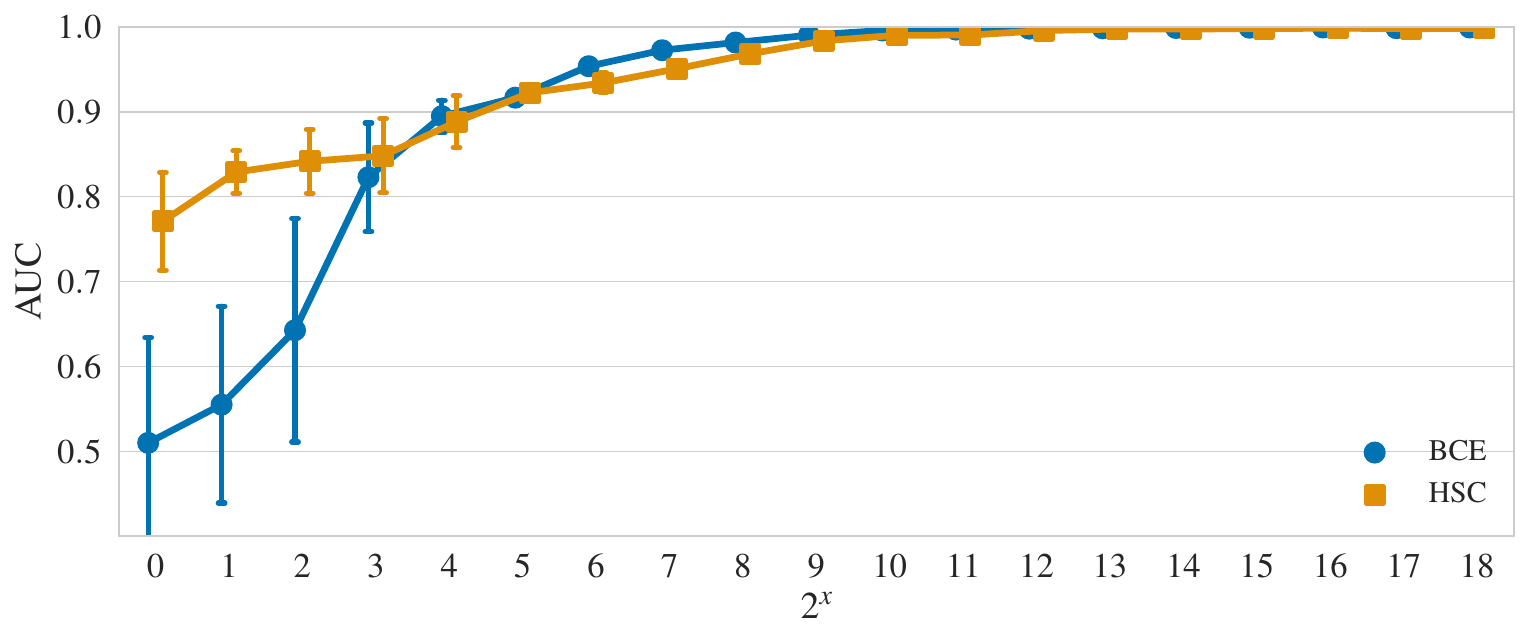}}
\subfigure[Class: ambulance]{\includegraphics[width=0.329\linewidth]{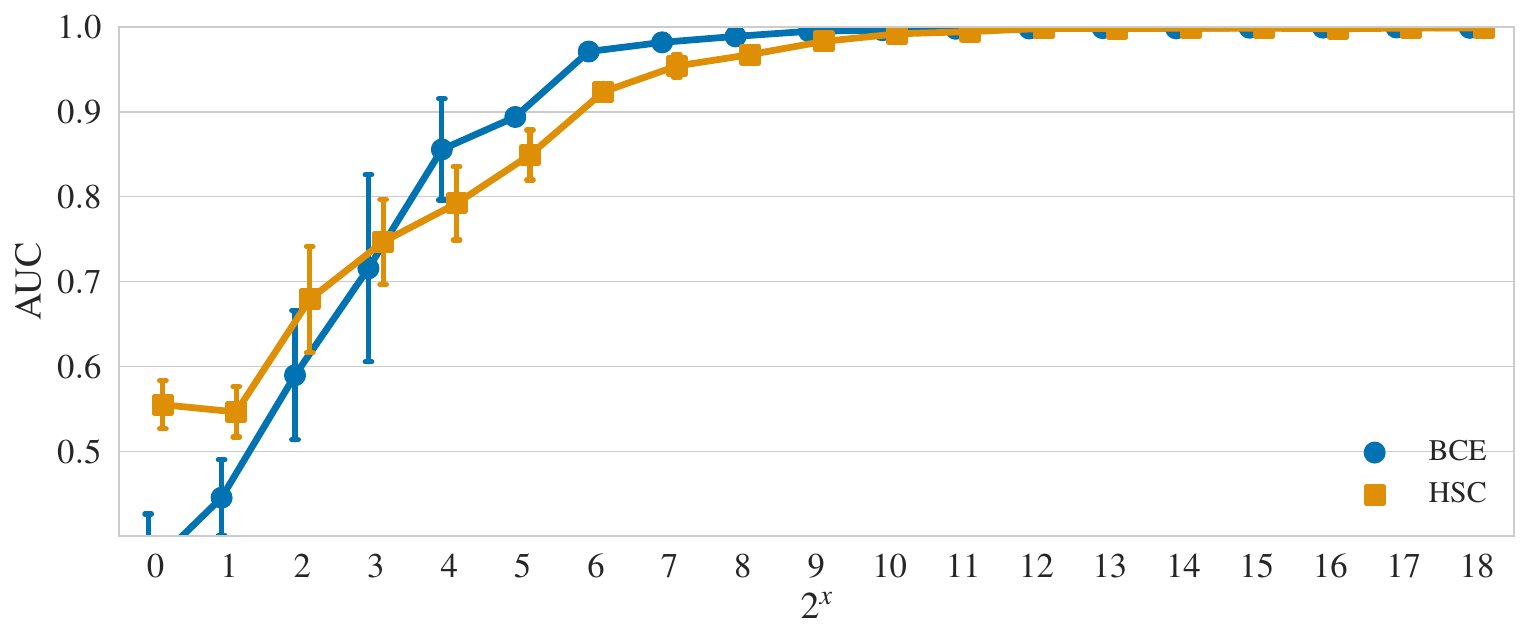}}
\subfigure[Class: american alligator]{\includegraphics[width=0.329\linewidth]{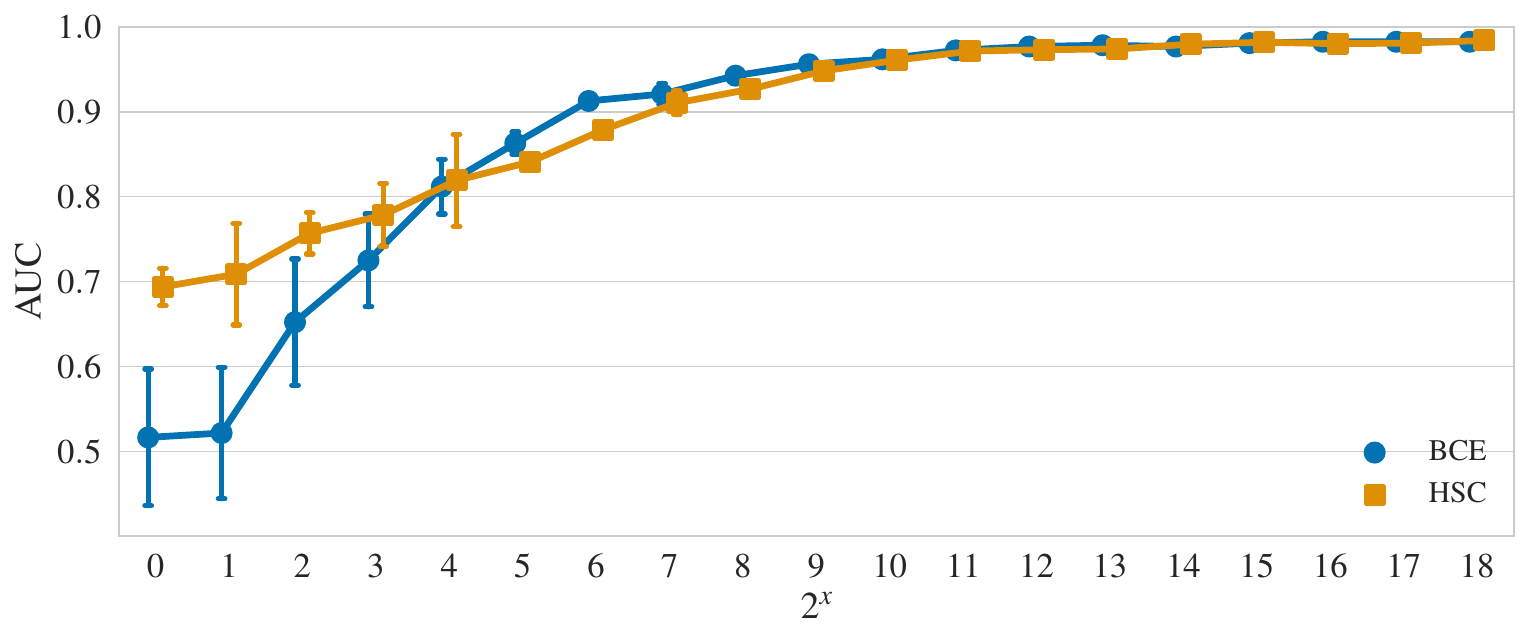}}
\subfigure[Class: banjo]{\includegraphics[width=0.329\linewidth]{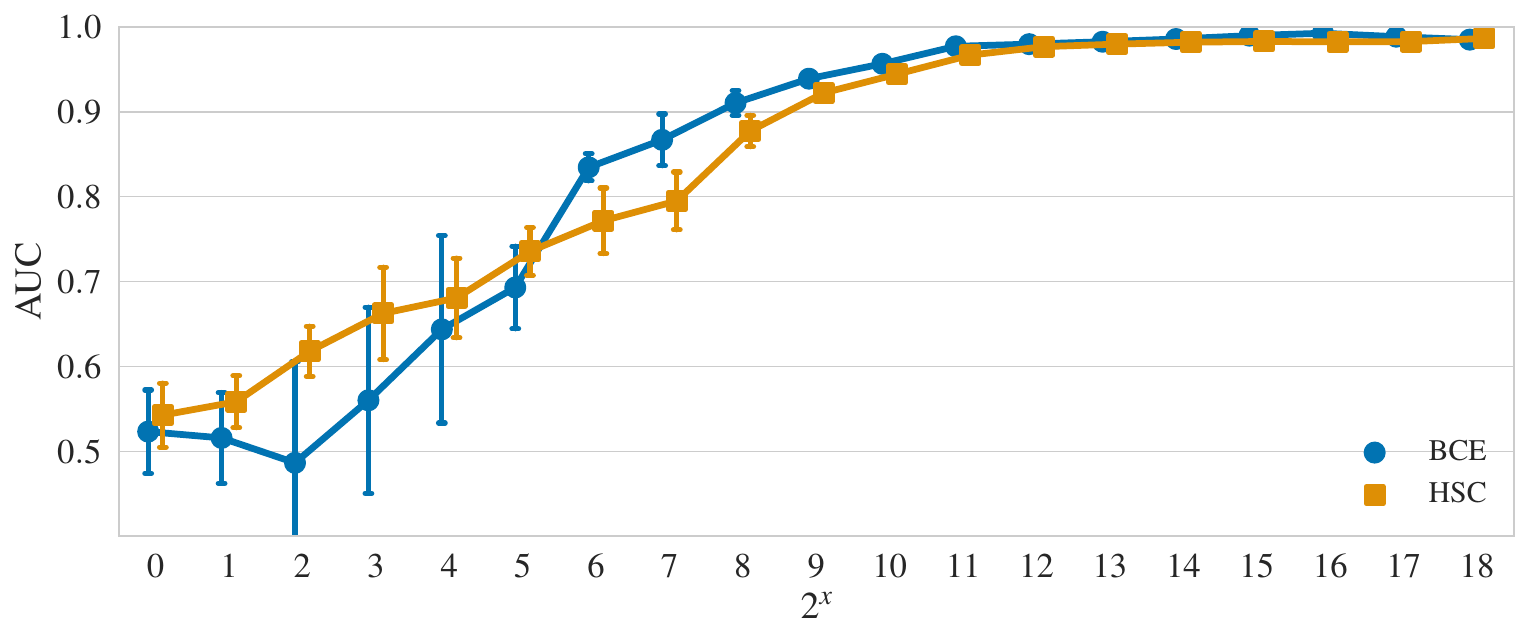}}
\subfigure[Class: barn]{\includegraphics[width=0.329\linewidth]{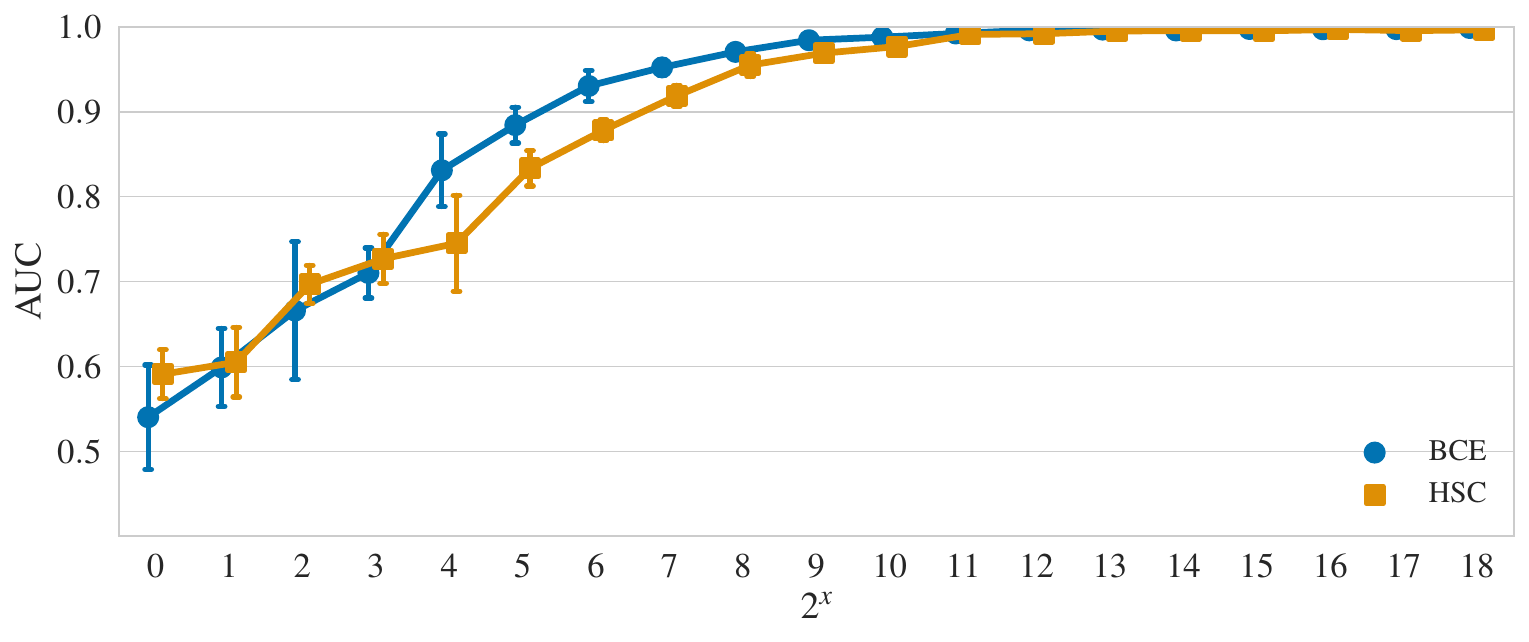}}
\subfigure[Class: bikini]{\includegraphics[width=0.329\linewidth]{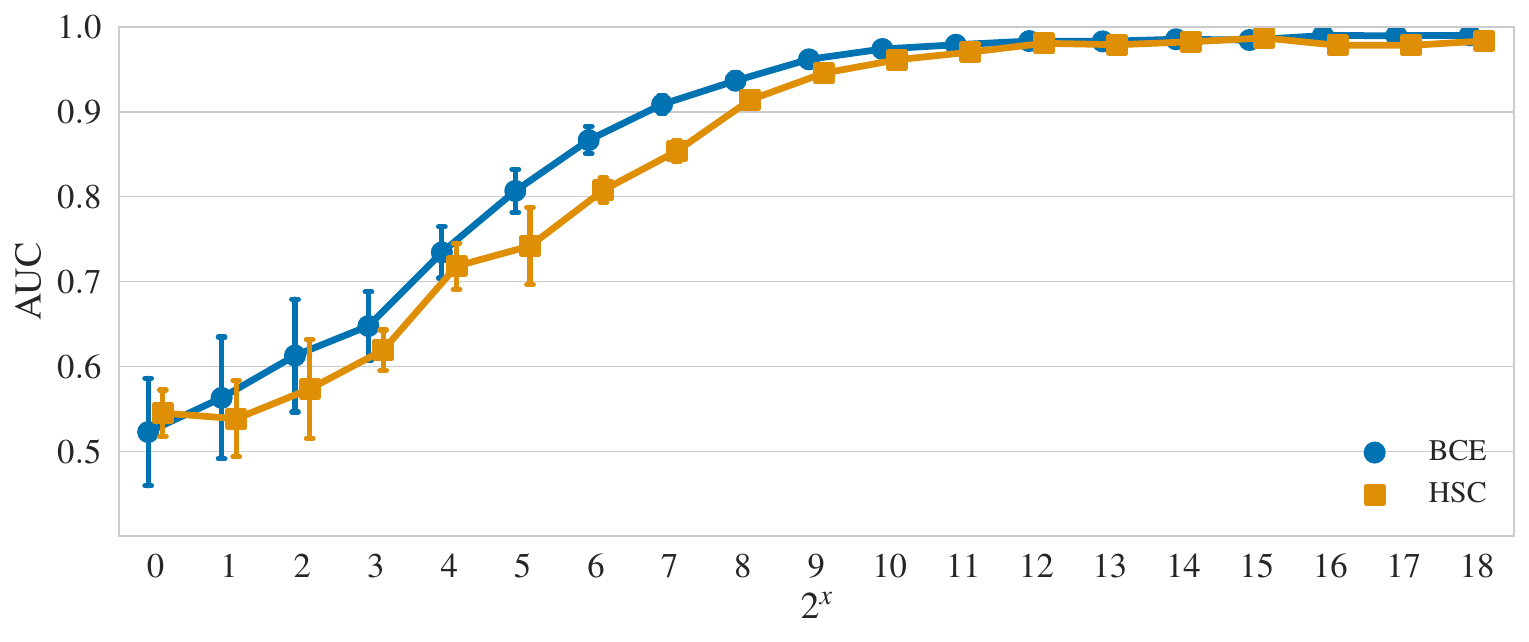}}
\subfigure[Class: digital clock]{\includegraphics[width=0.329\linewidth]{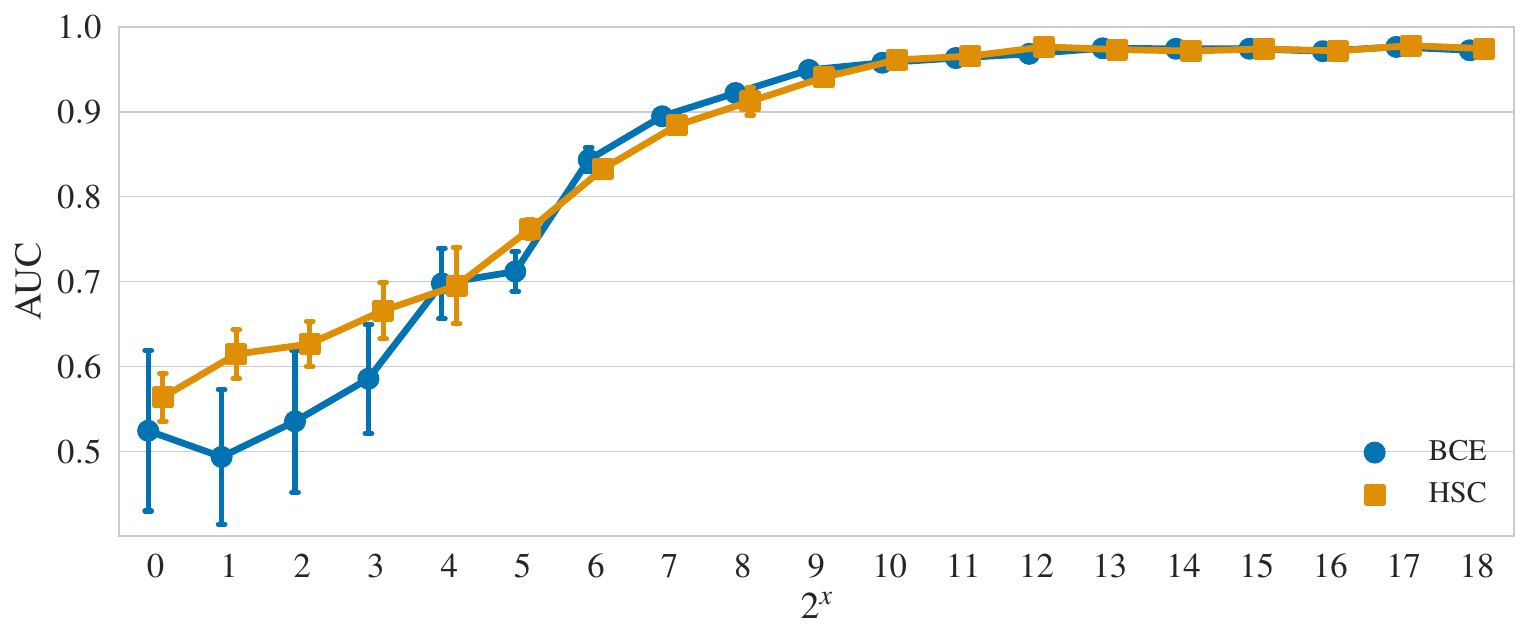}}
\subfigure[Class: dragonfly]{\includegraphics[width=0.329\linewidth]{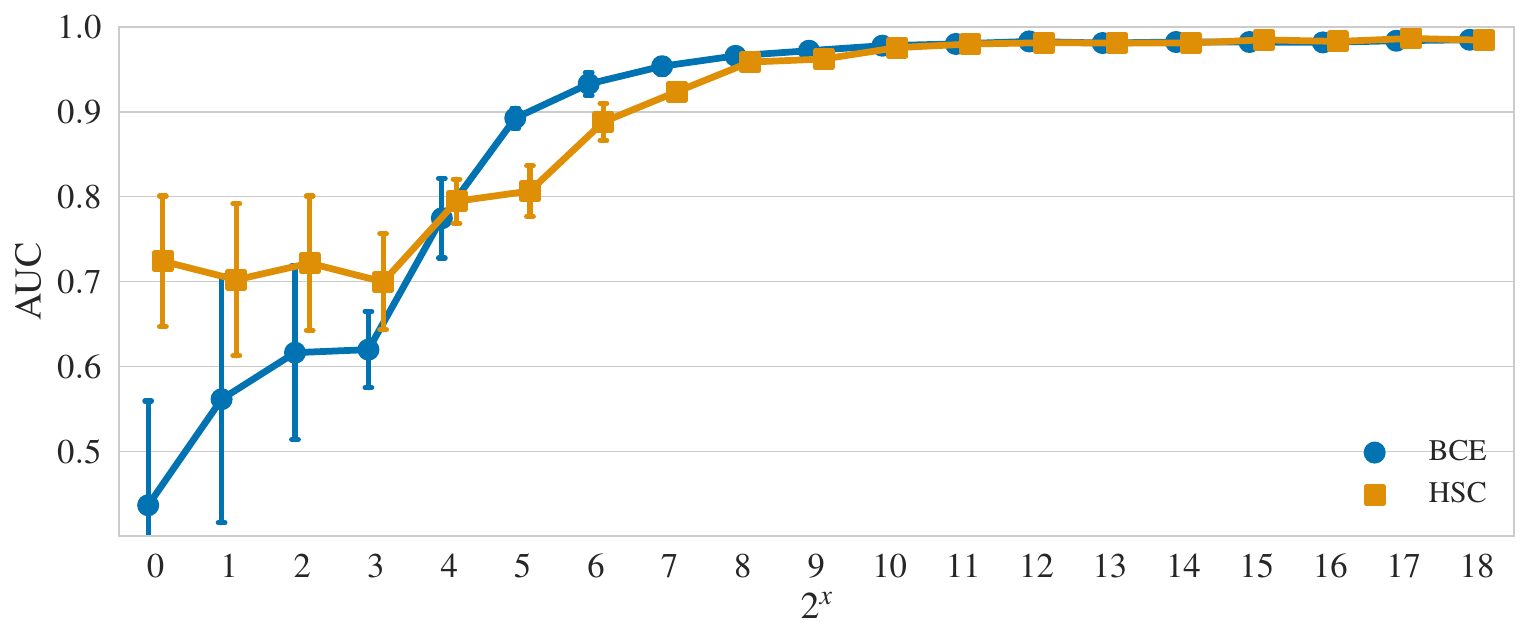}}
\subfigure[Class: dumbbell]{\includegraphics[width=0.329\linewidth]{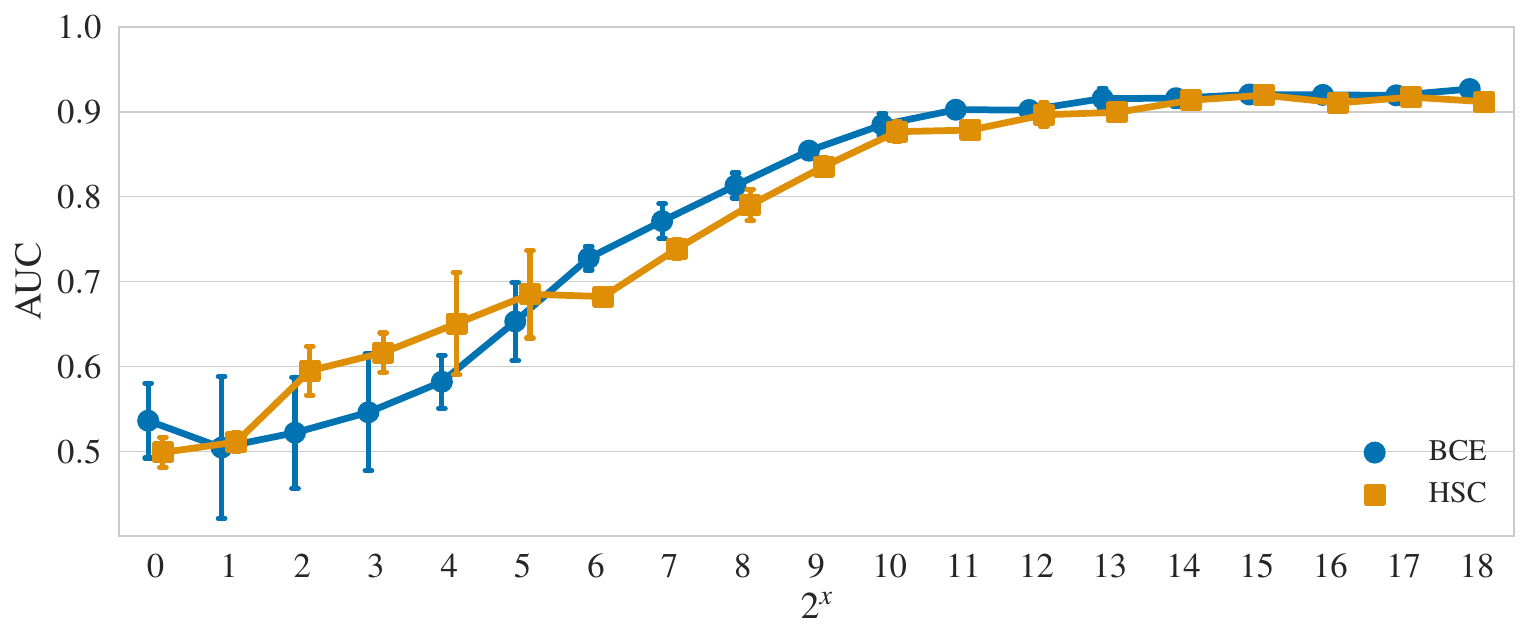}}
\subfigure[Class: forklift]{\includegraphics[width=0.329\linewidth]{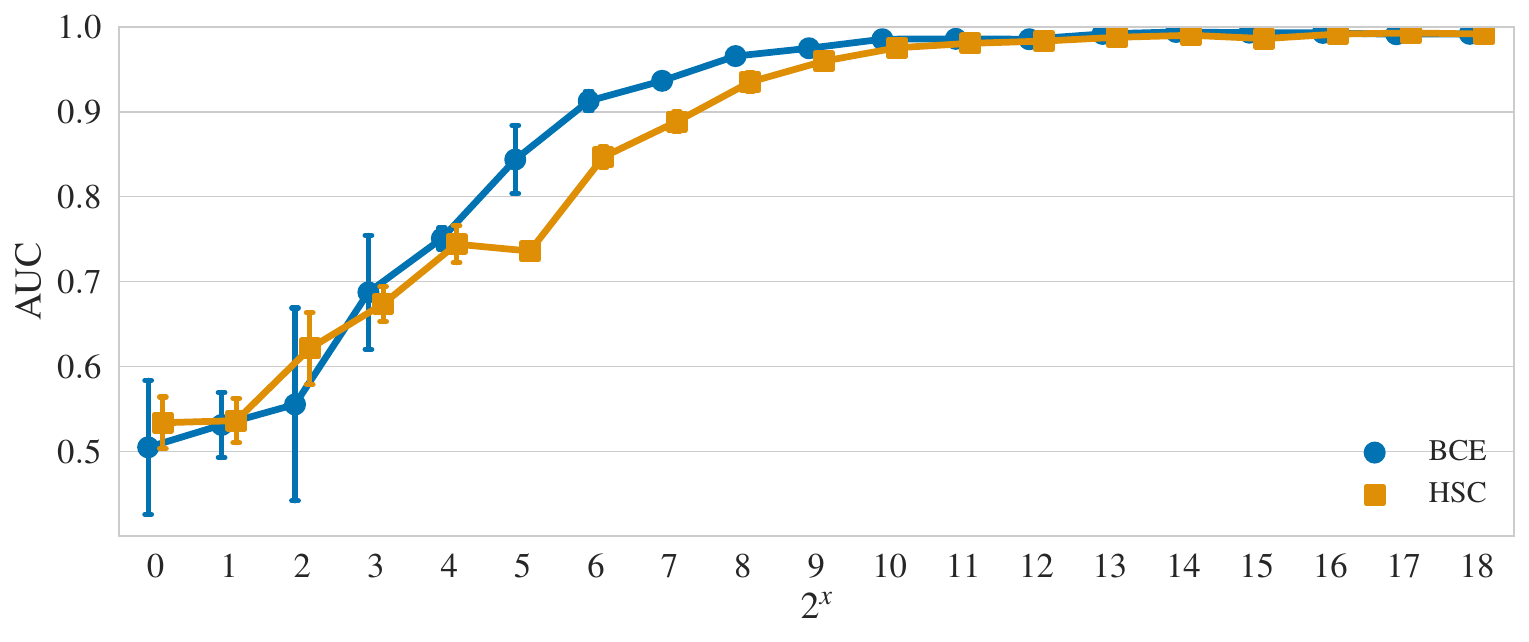}}
\subfigure[Class: goblet]{\includegraphics[width=0.329\linewidth]{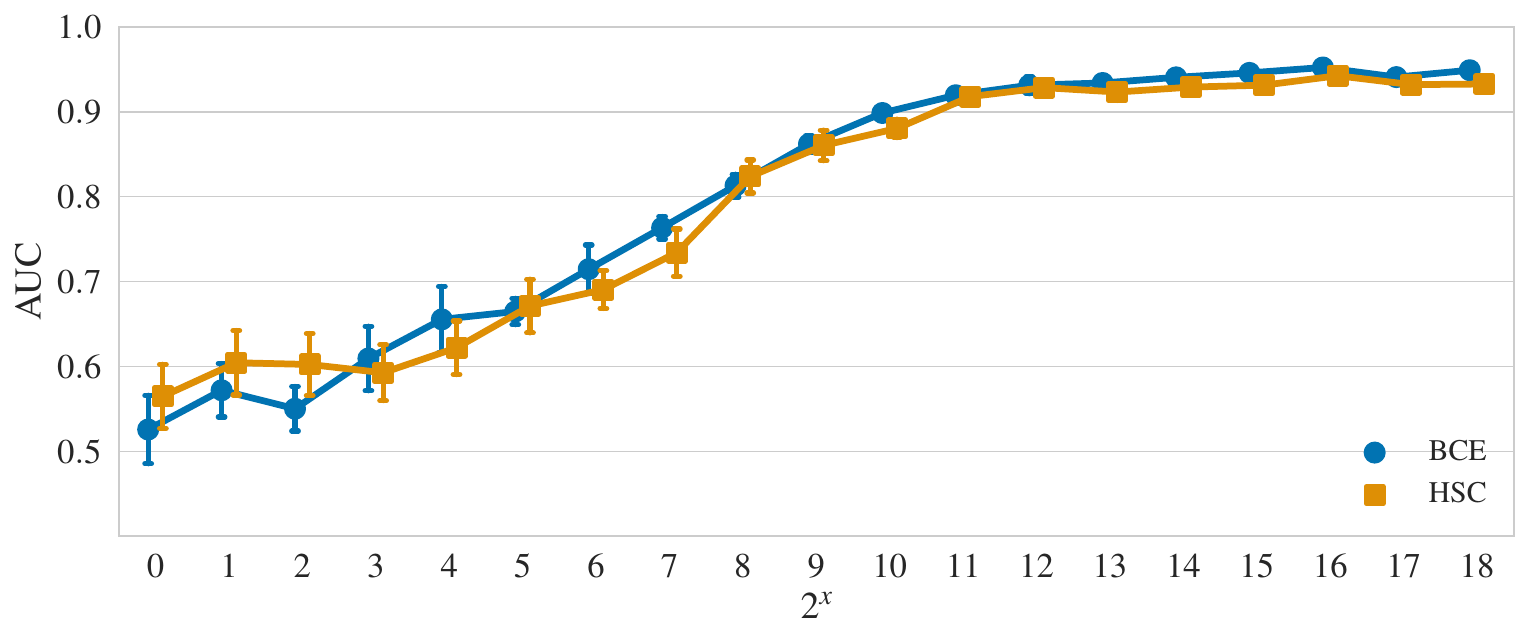}}
\subfigure[Class: grand piano]{\includegraphics[width=0.329\linewidth]{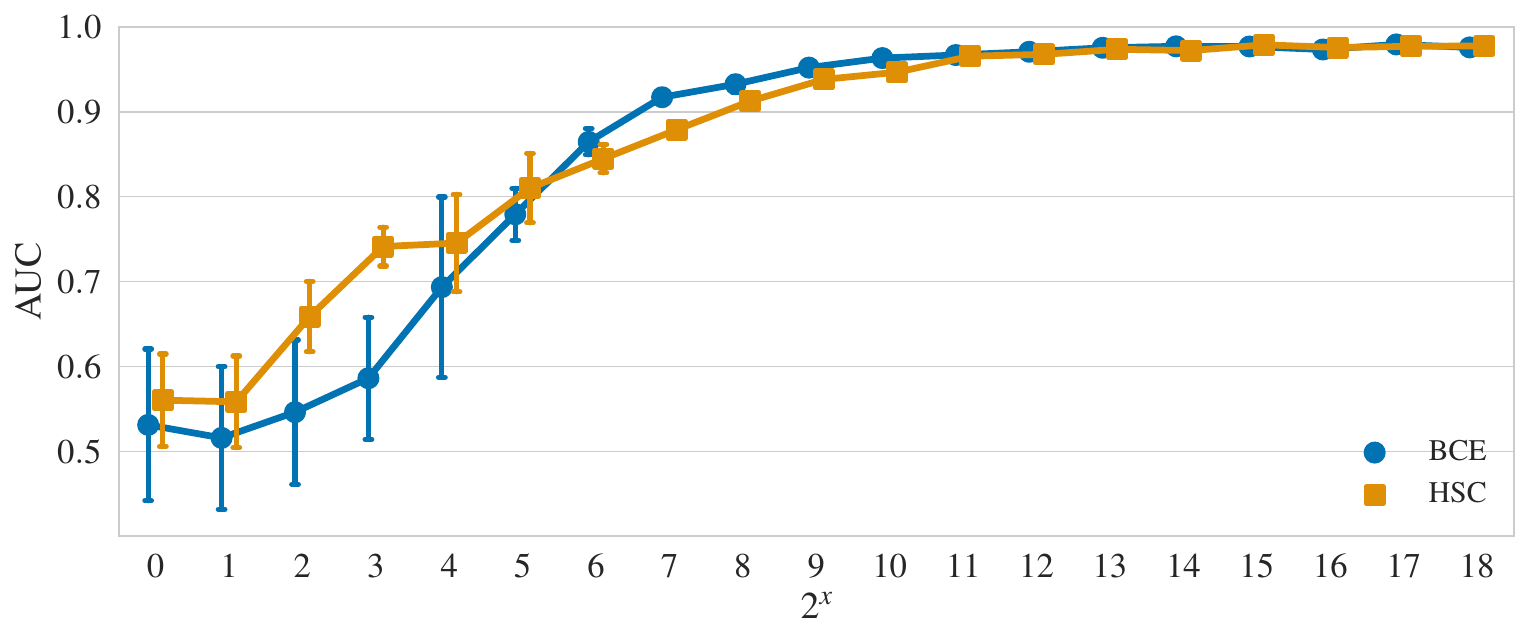}}
\subfigure[Class: hotdog]{\includegraphics[width=0.329\linewidth]{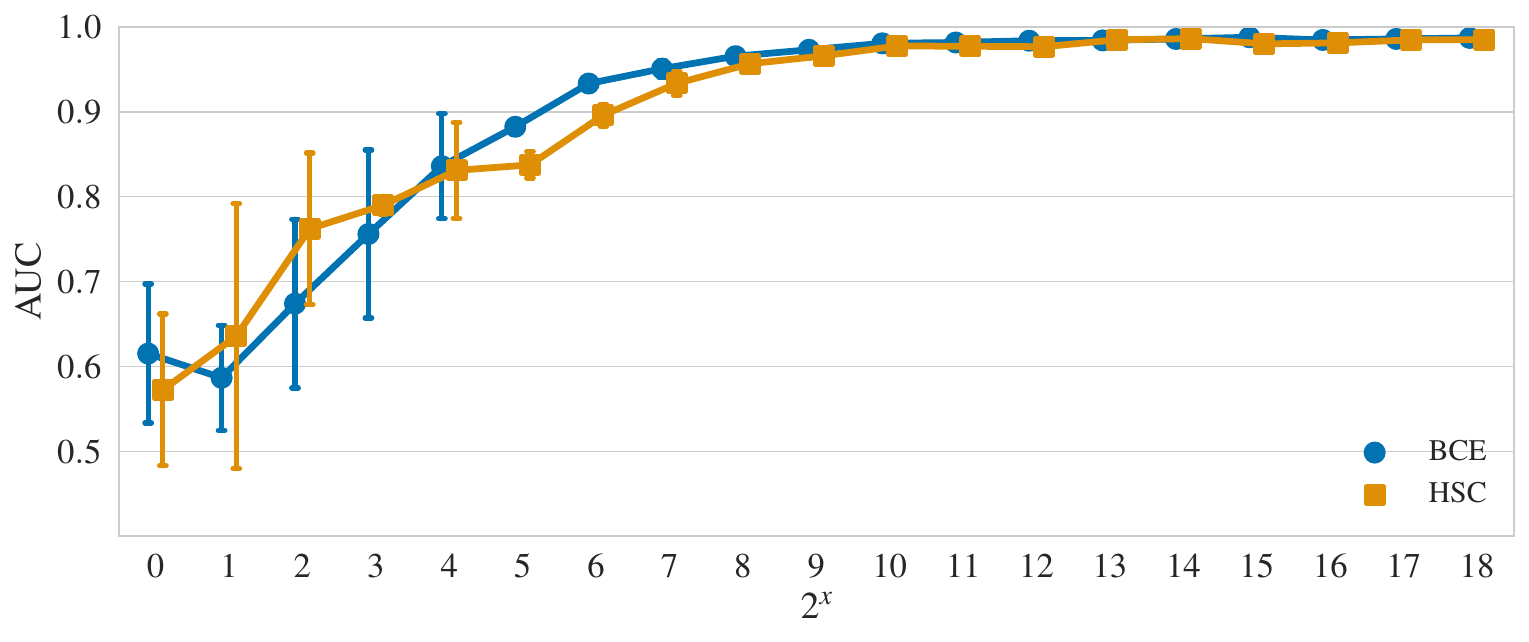}}
\subfigure[Class: hourglass]{\includegraphics[width=0.329\linewidth]{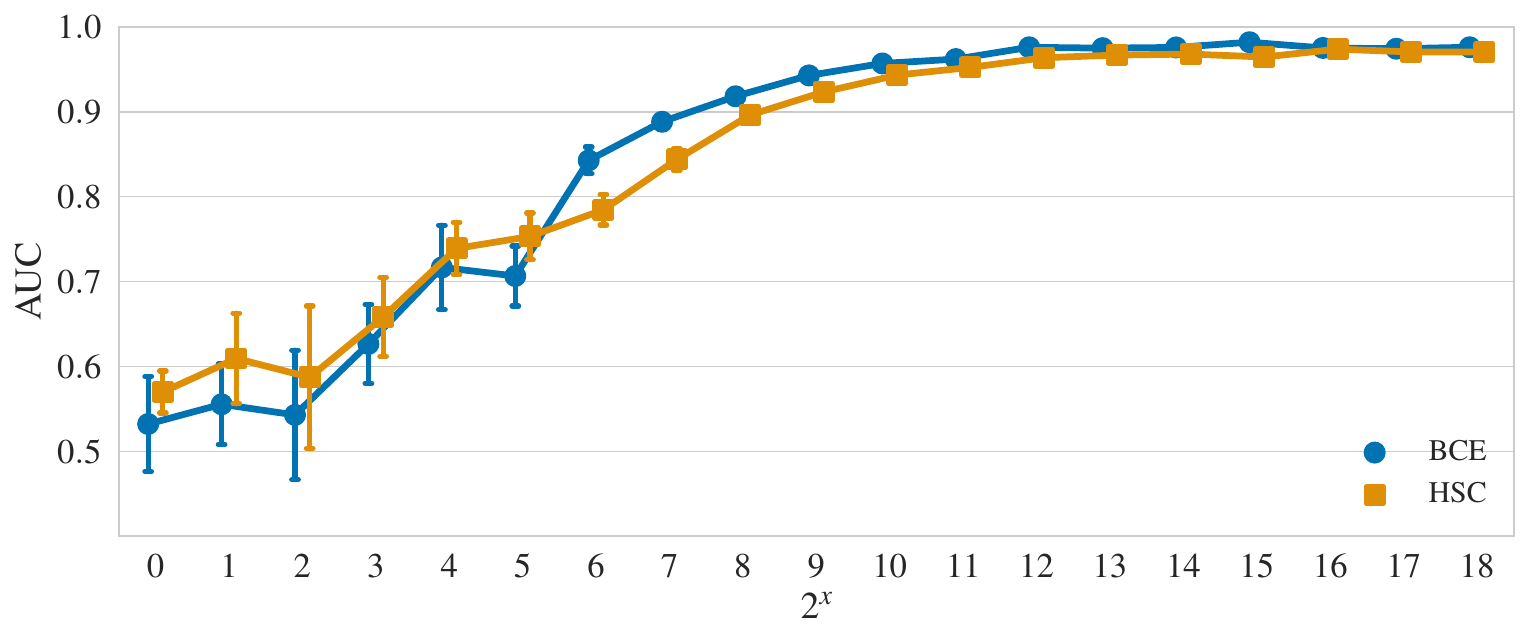}}
\label{fig:imagenet1kvs22k_classes1}
\end{figure*}

\begin{figure*}[th]
\centering
\caption{Mean AUC detection performance in \% (over 5 seeds) for all classes of the ImageNet-1K one vs.~rest benchmark from Section \ref{subsec:exp_imagenet} when varying the number of ImageNet-22K OE samples. These plots correspond to Figure \ref{fig:imagenet1kvs22k}, but here we report the results for all individual classes (from class 16 (manhole cover) to class 30 (volcano)).}
\subfigure[Class: manhole cover]{\includegraphics[width=0.329\linewidth]{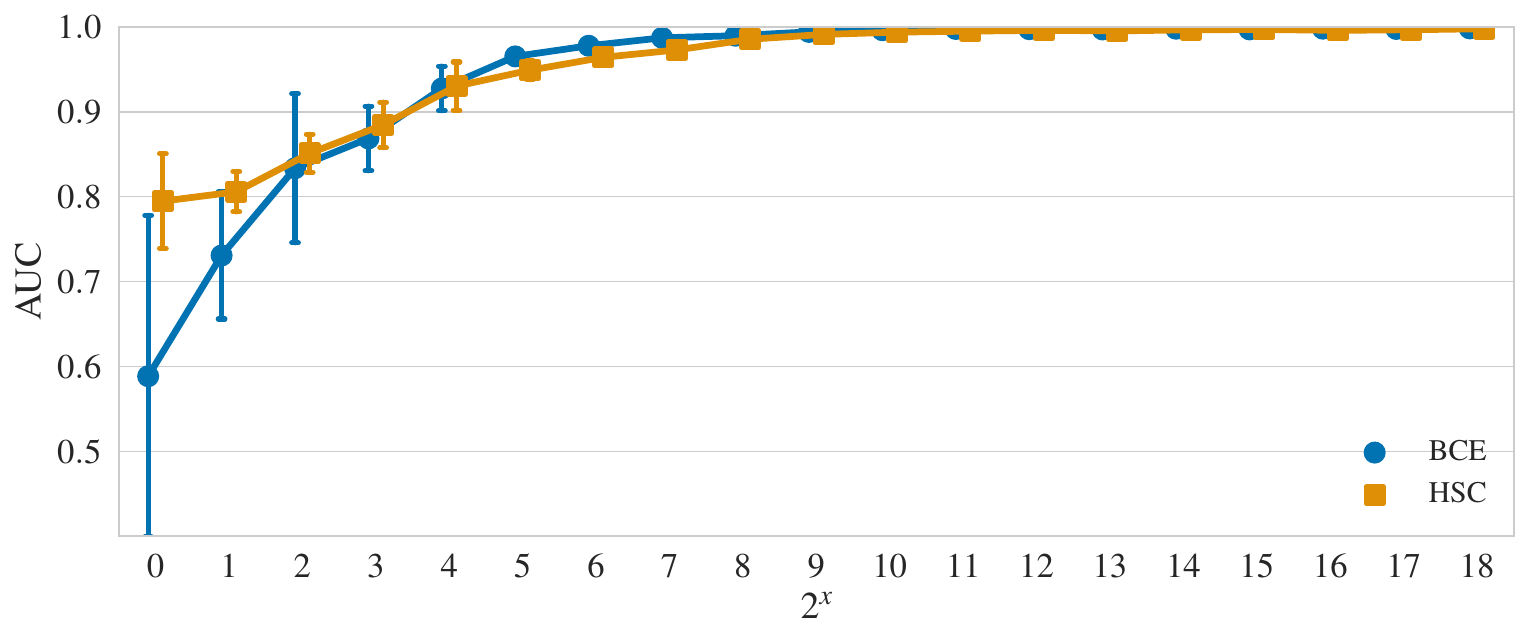}}
\subfigure[Class: mosque]{\includegraphics[width=0.329\linewidth]{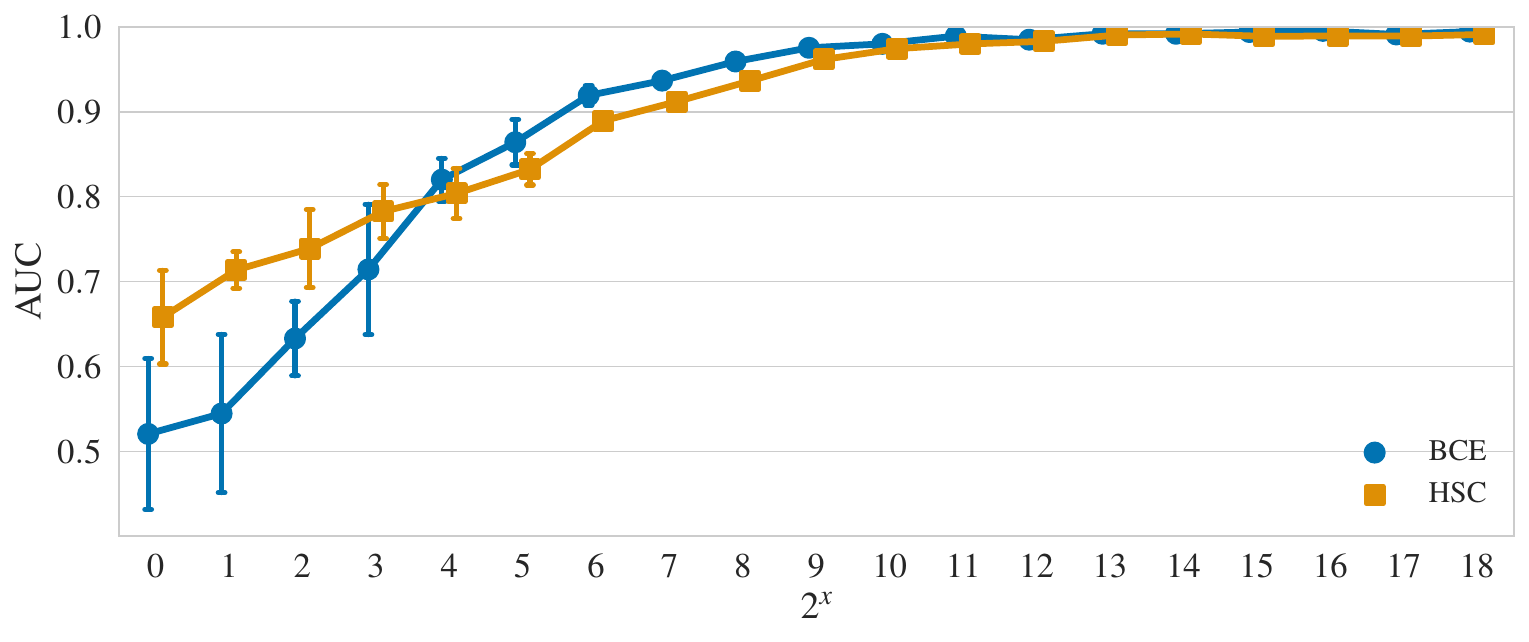}}
\subfigure[Class: nail]{\includegraphics[width=0.329\linewidth]{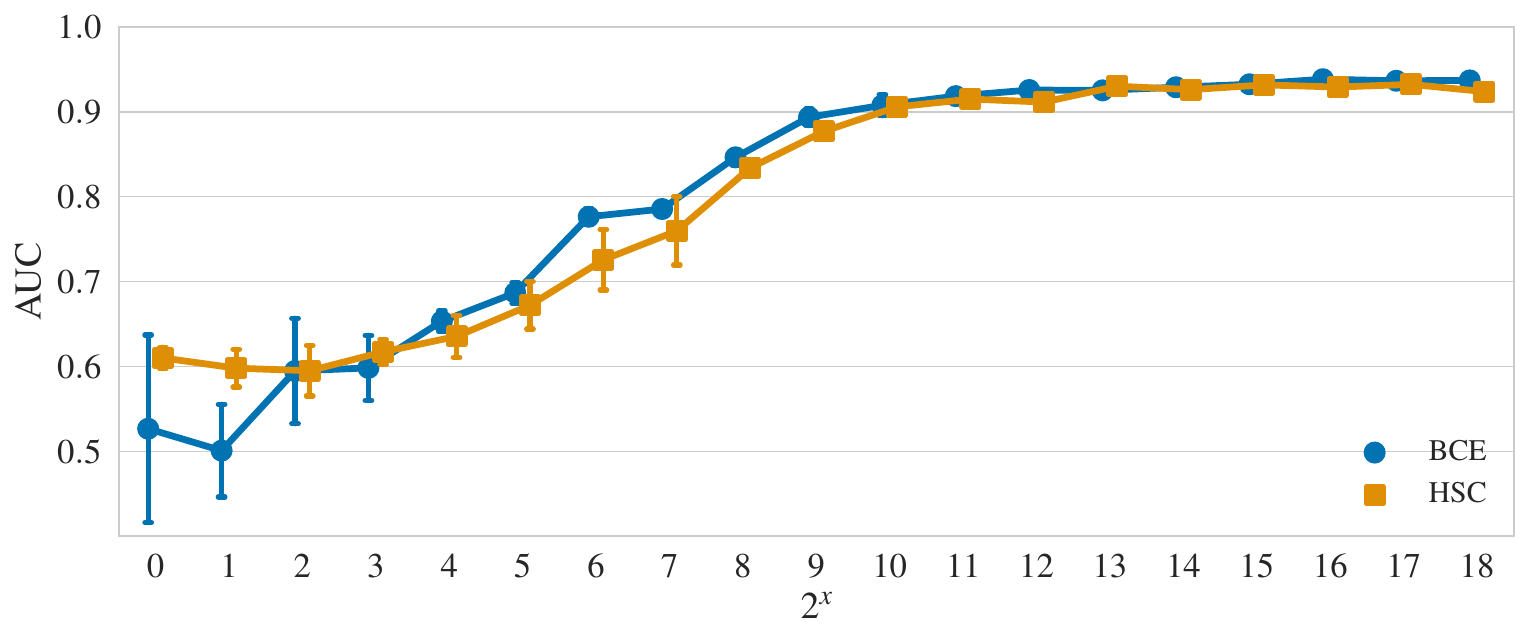}}
\subfigure[Class: parking meter]{\includegraphics[width=0.329\linewidth]{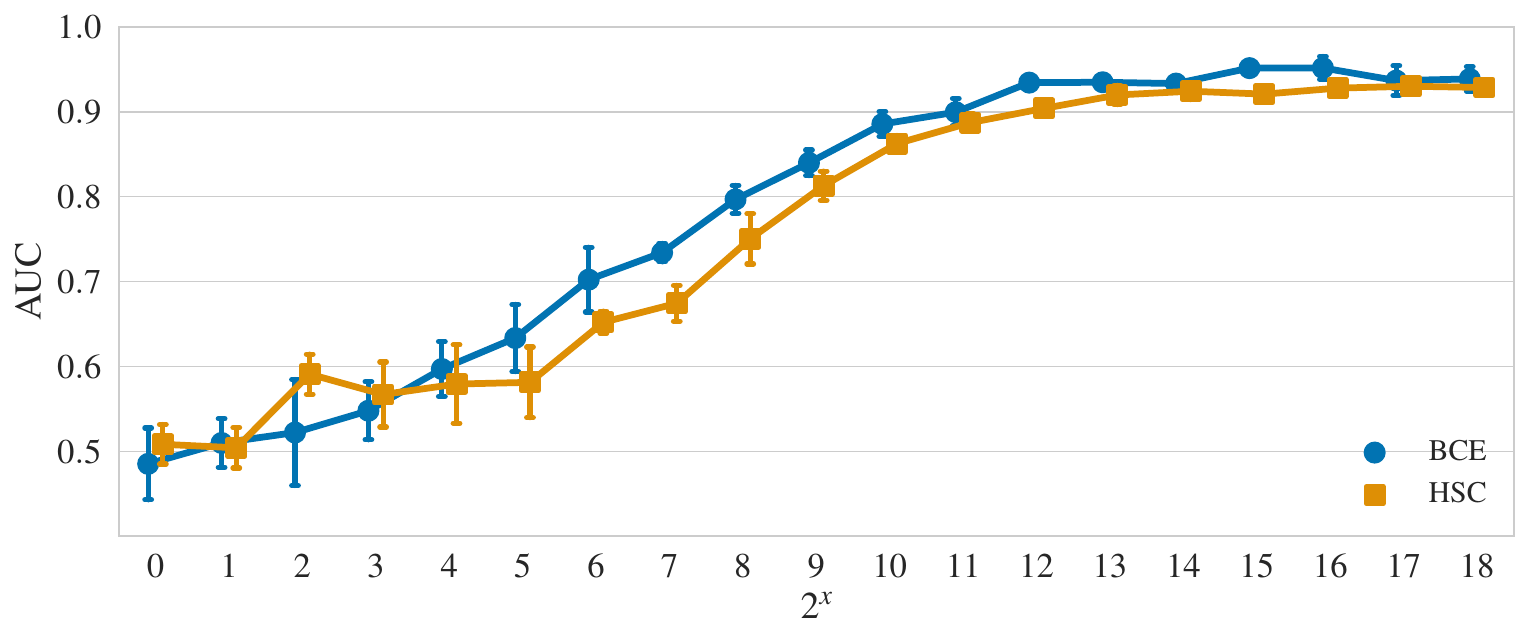}}
\subfigure[Class: pillow]{\includegraphics[width=0.329\linewidth]{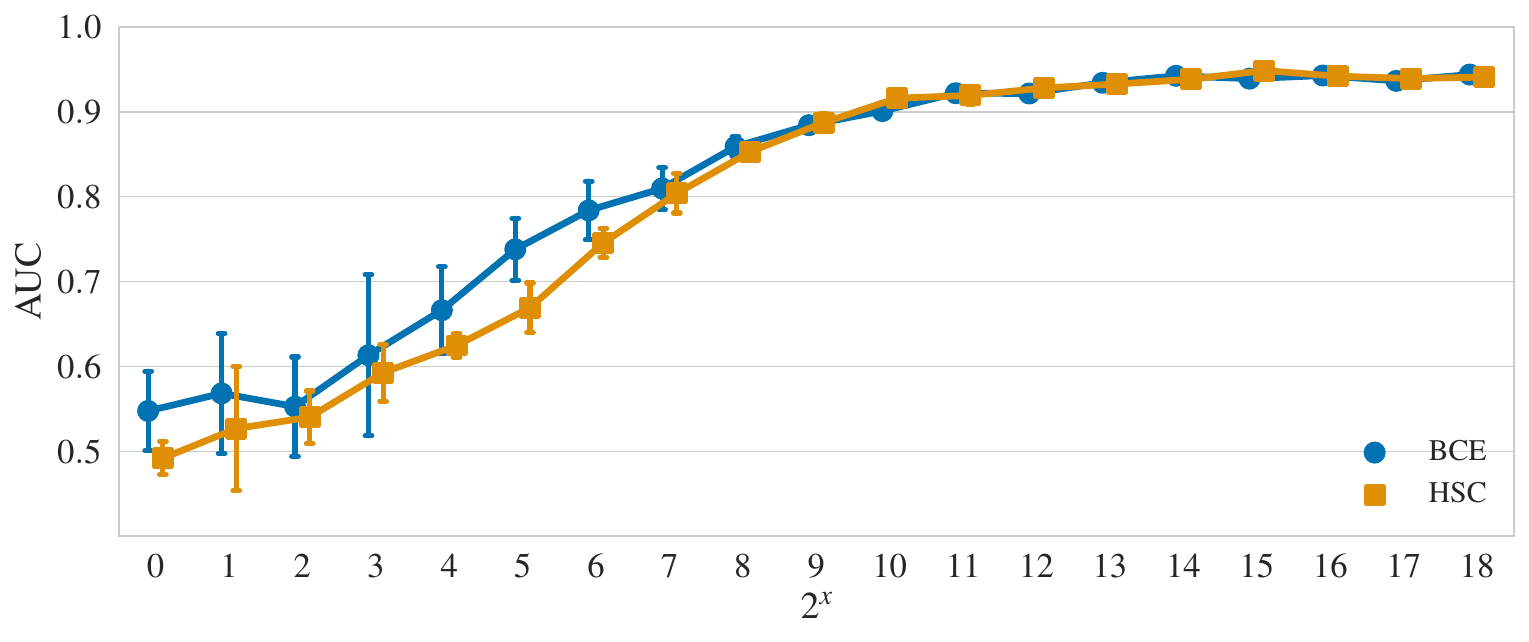}}
\subfigure[Class: revolver]{\includegraphics[width=0.329\linewidth]{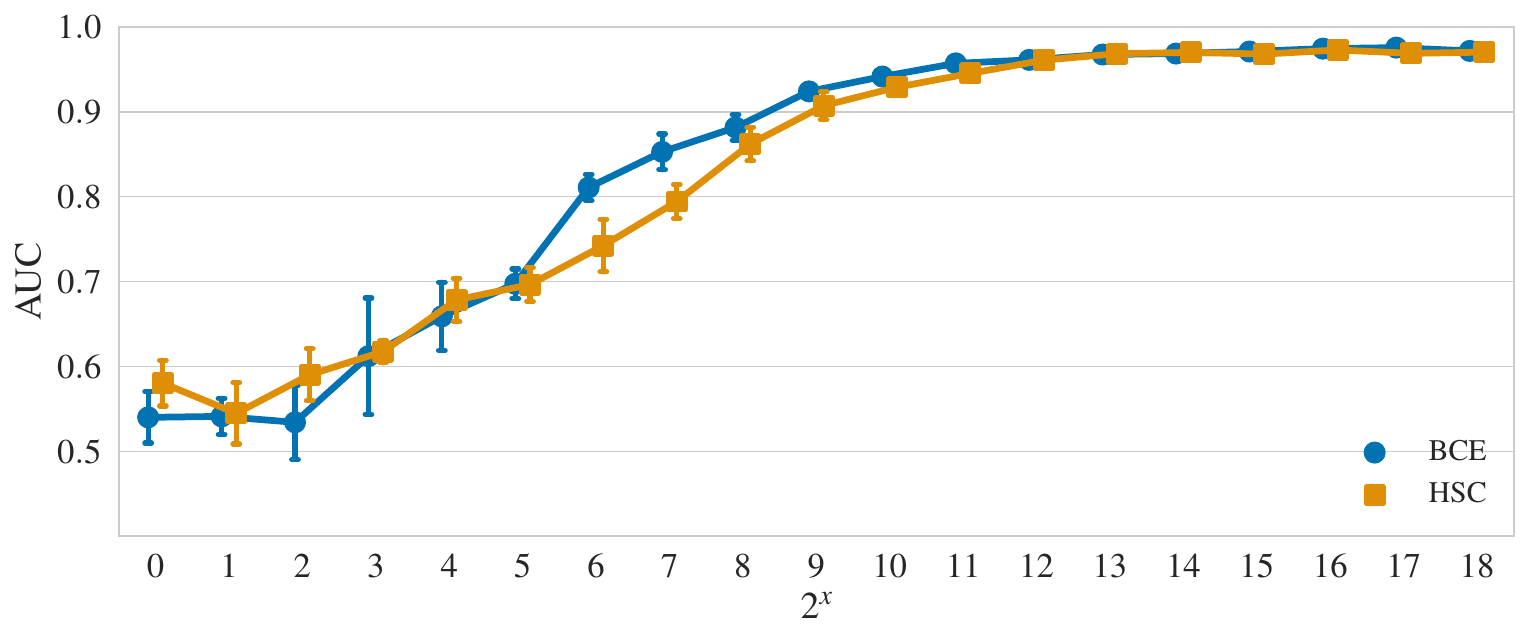}}
\subfigure[Class: rotary dial telephone]{\includegraphics[width=0.329\linewidth]{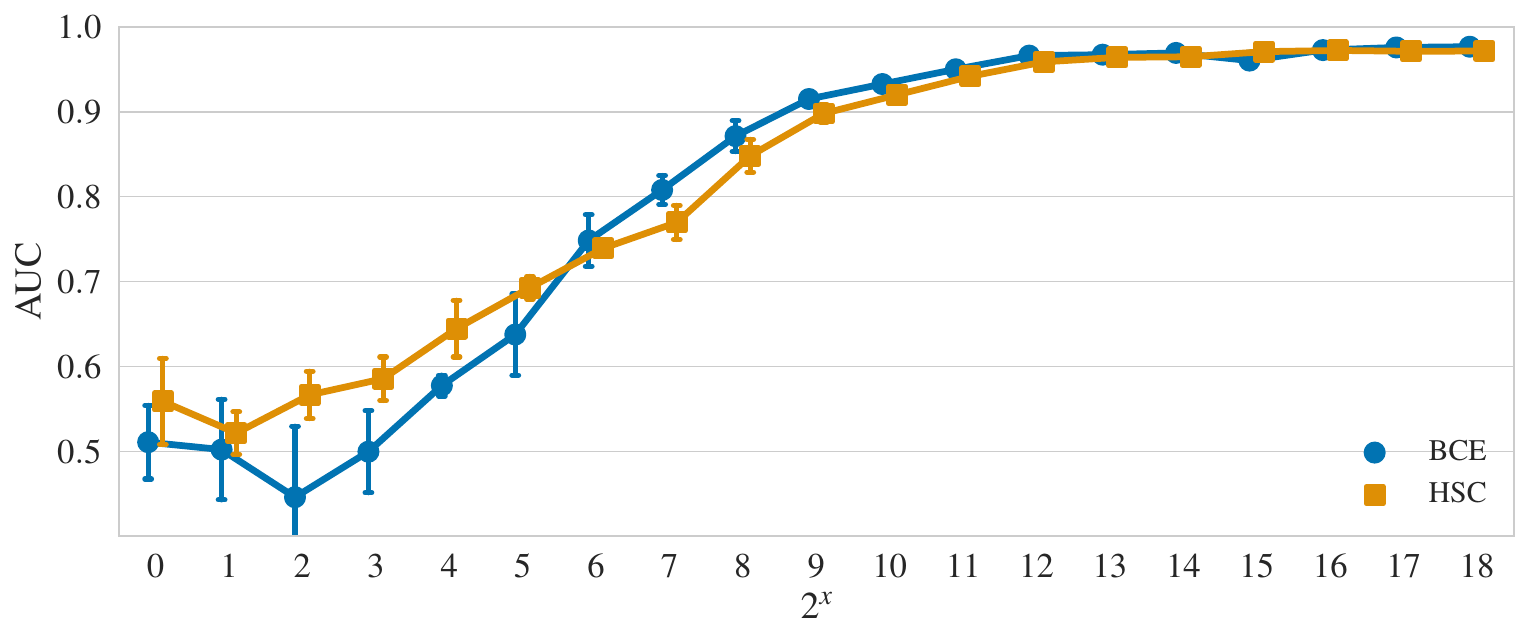}}
\subfigure[Class: schooner]{\includegraphics[width=0.329\linewidth]{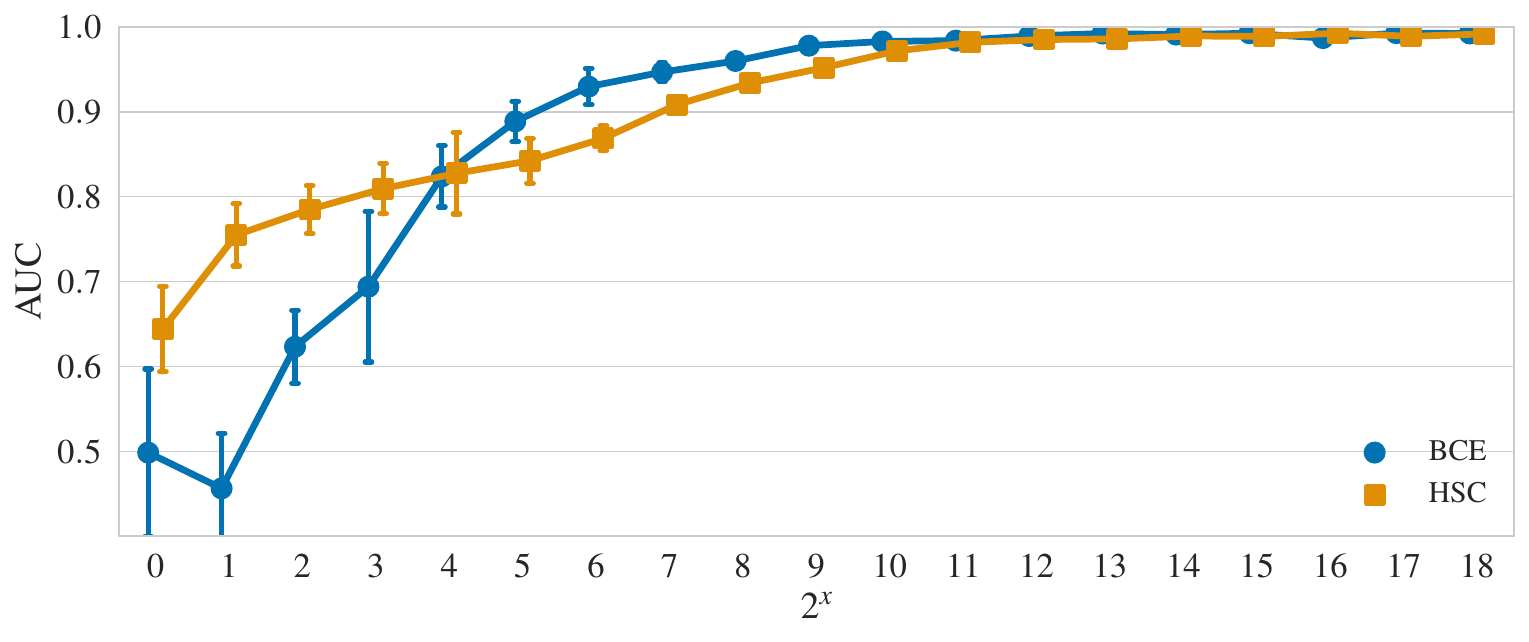}}
\subfigure[Class: snowmobile]{\includegraphics[width=0.329\linewidth]{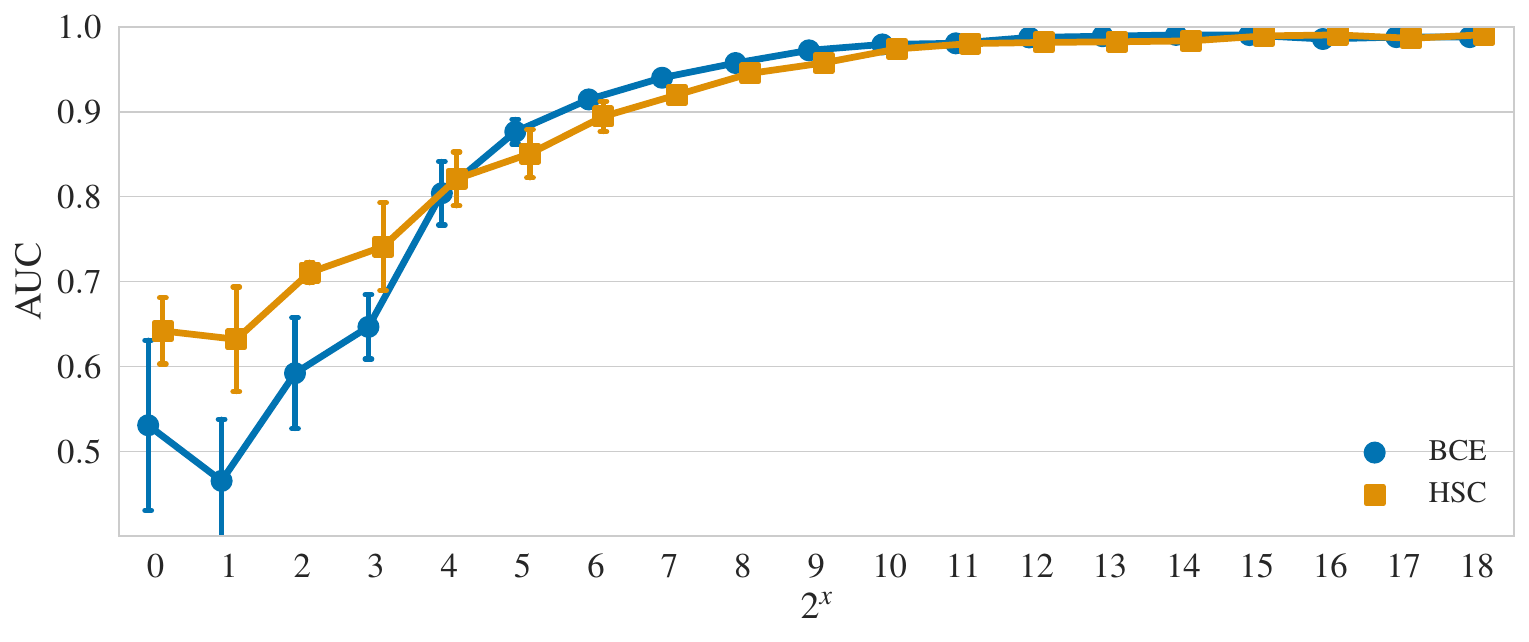}}
\subfigure[Class: soccer ball]{\includegraphics[width=0.329\linewidth]{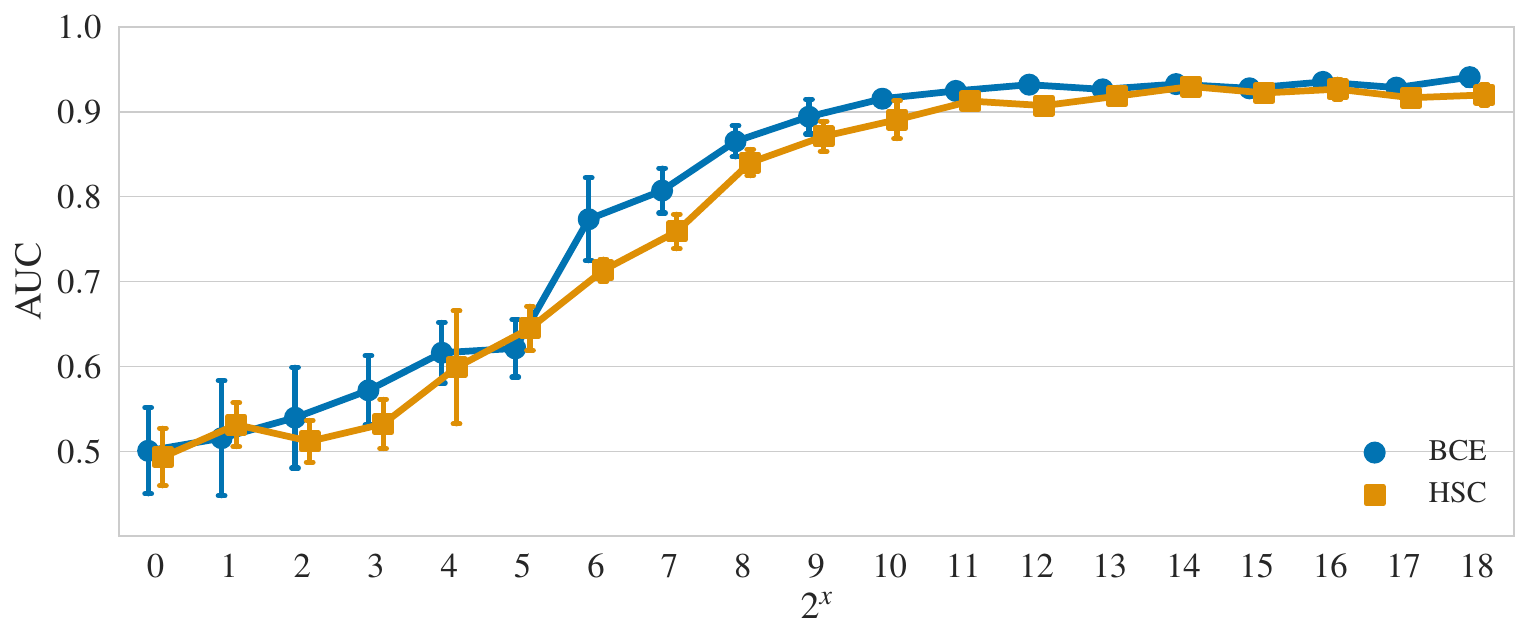}}
\subfigure[Class: stingray]{\includegraphics[width=0.329\linewidth]{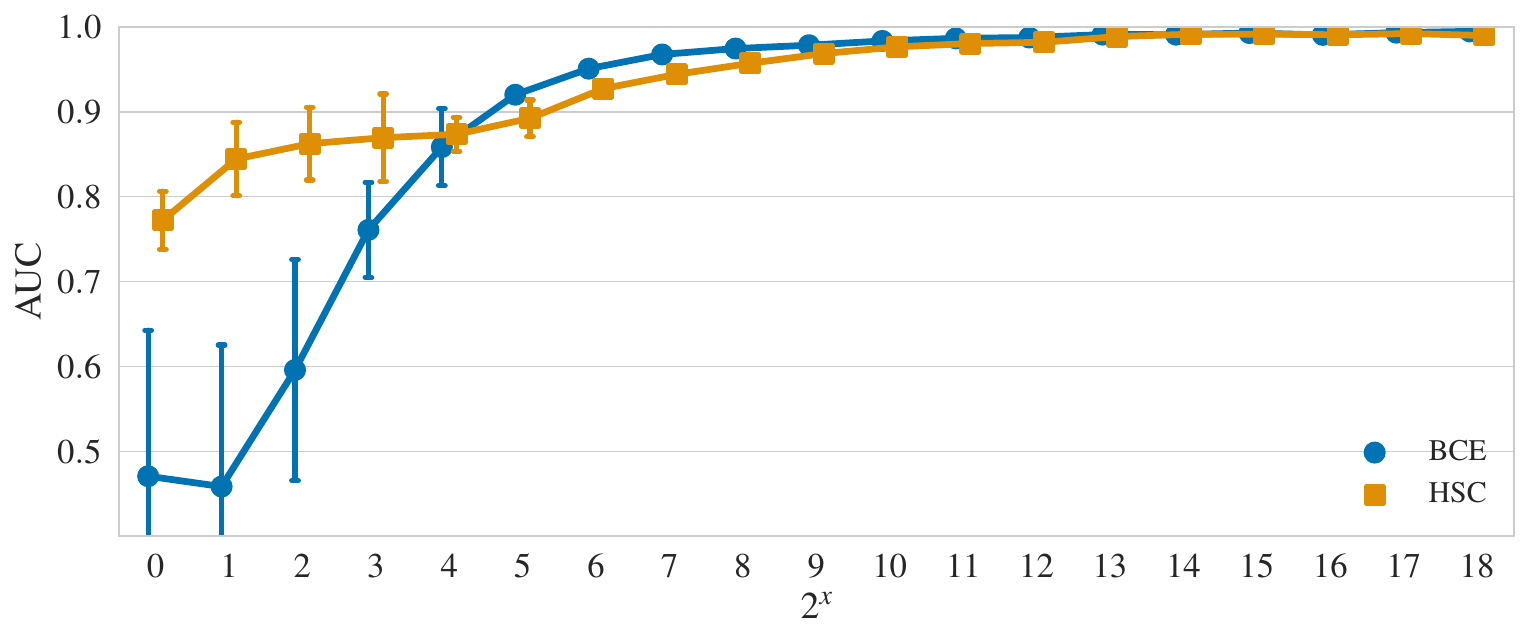}}
\subfigure[Class: strawberry]{\includegraphics[width=0.329\linewidth]{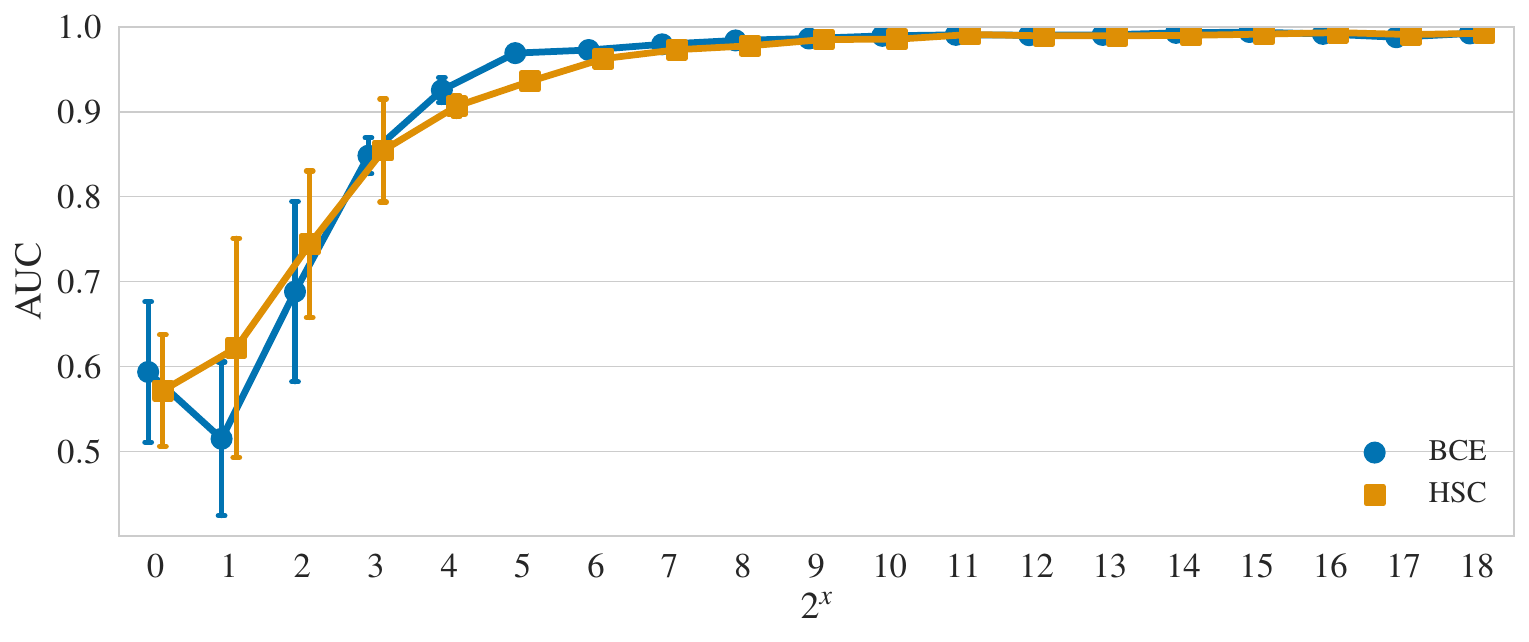}}
\subfigure[Class: tank]{\includegraphics[width=0.329\linewidth]{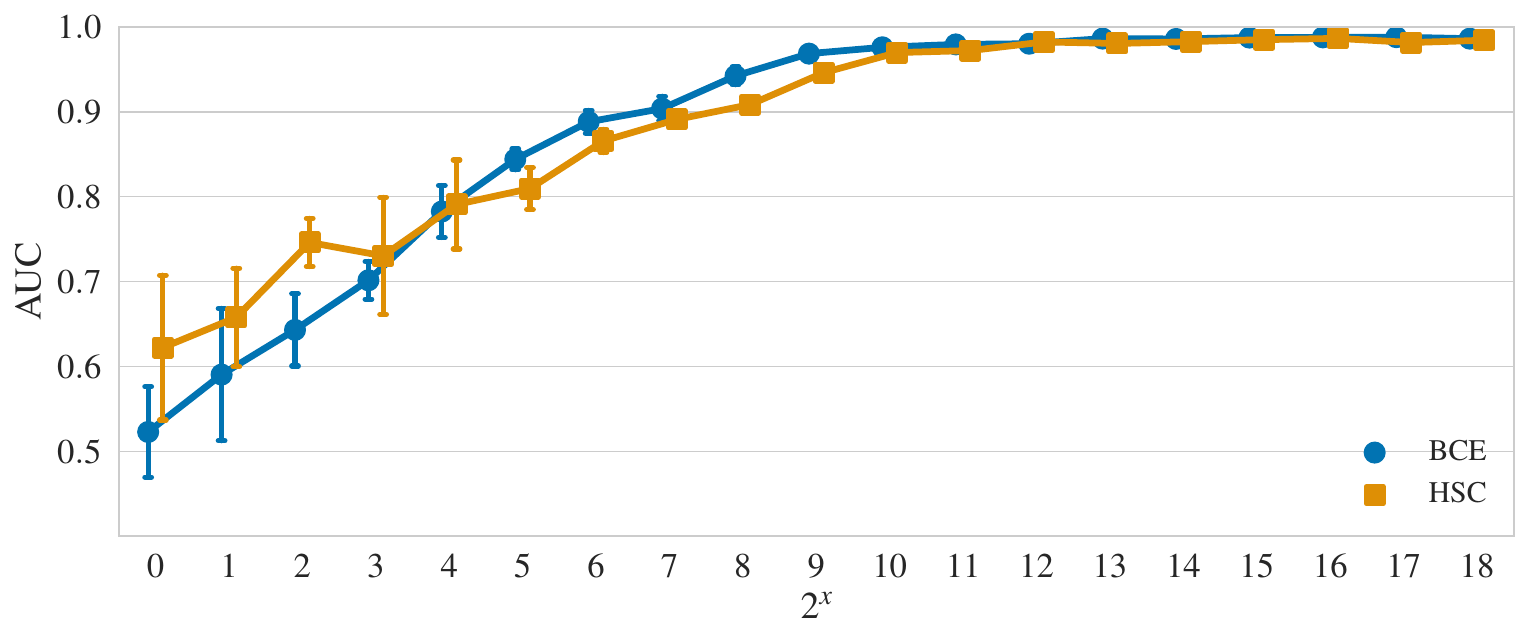}}
\subfigure[Class: toaster]{\includegraphics[width=0.329\linewidth]{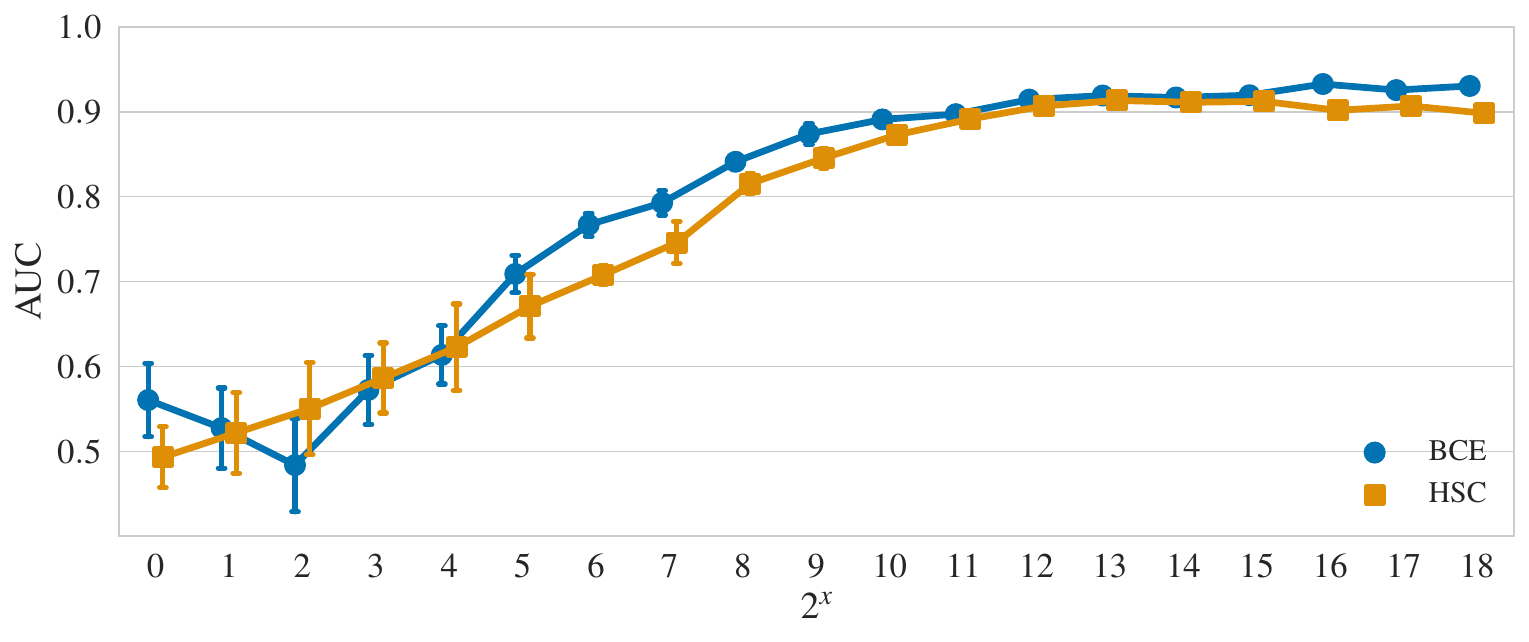}}
\subfigure[Class: volcano]{\includegraphics[width=0.329\linewidth]{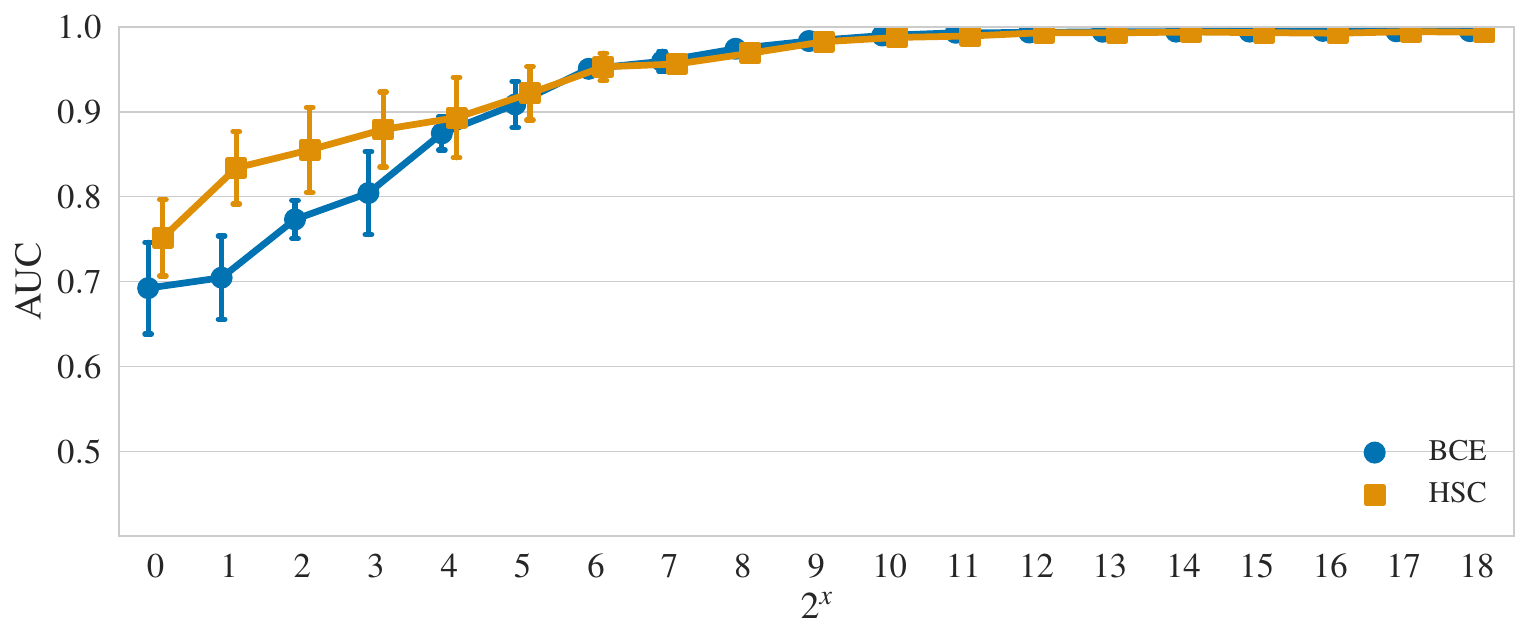}}
\label{fig:imagenet1kvs22k_classes2}
\end{figure*}

\begin{figure*}[th]
\centering
\caption{Mean AUC detection performance in \% (over 10 seeds) for all MNIST classes from the experiment in Appendix \ref{appx:oe_diversity} on varying the number of classes $k$ of the EMNIST-Letters OE data. These plots correspond to Figure \ref{fig:diversity_mnist}, but here we report the results for all individual classes.}
\subfigure[Class: 0]{\label{fig:mnistvsemnist_0}\includegraphics[width=0.329\linewidth]{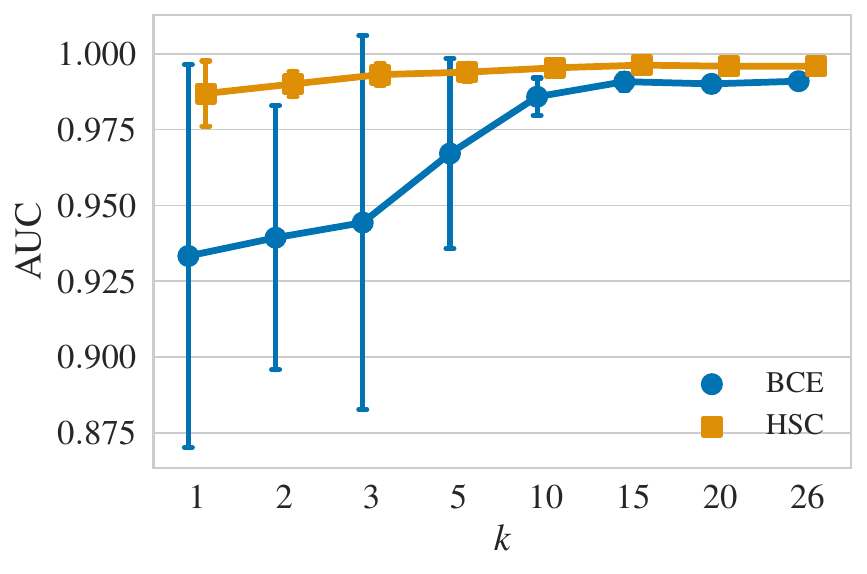}}
\subfigure[Class: 1]{\label{fig:mnistvsemnist_1}\includegraphics[width=0.329\linewidth]{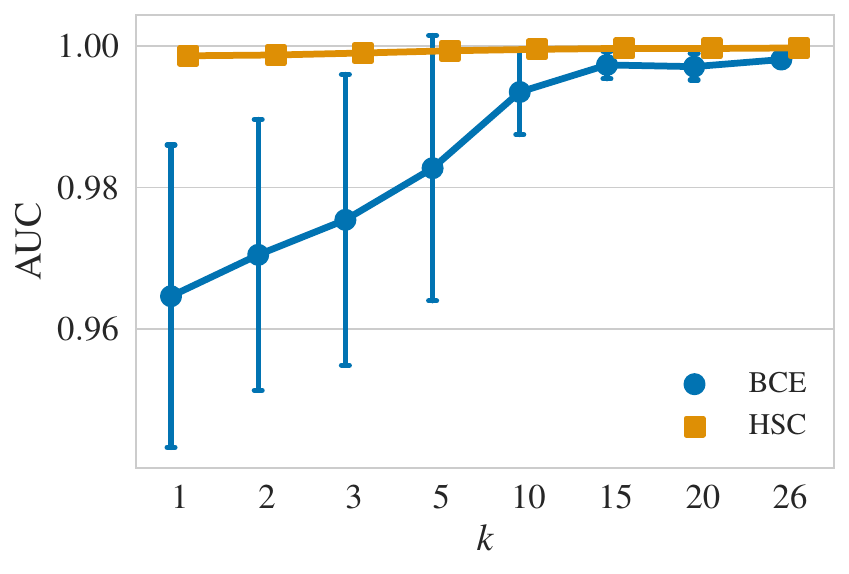}}
\subfigure[Class: 2]{\label{fig:mnistvsemnist_2}\includegraphics[width=0.329\linewidth]{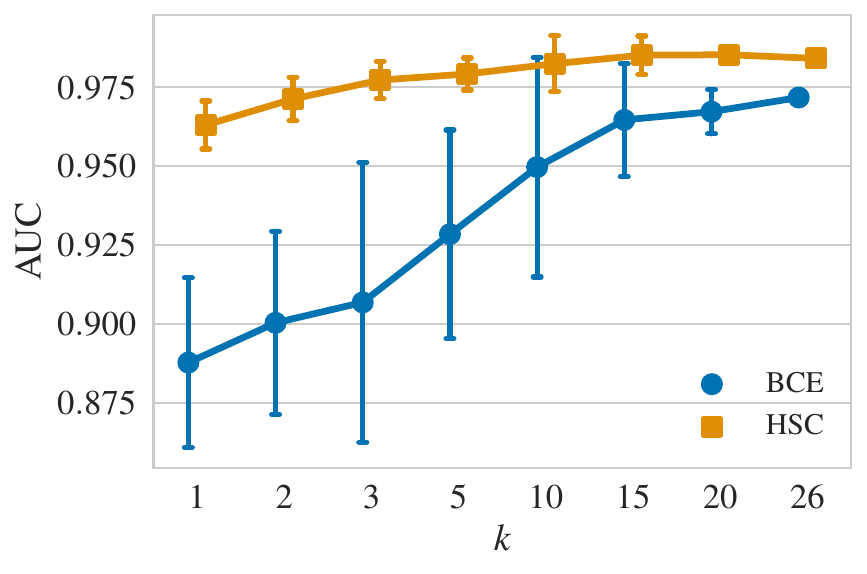}}
\subfigure[Class: 3]{\label{fig:mnistvsemnist_3}\includegraphics[width=0.329\linewidth]{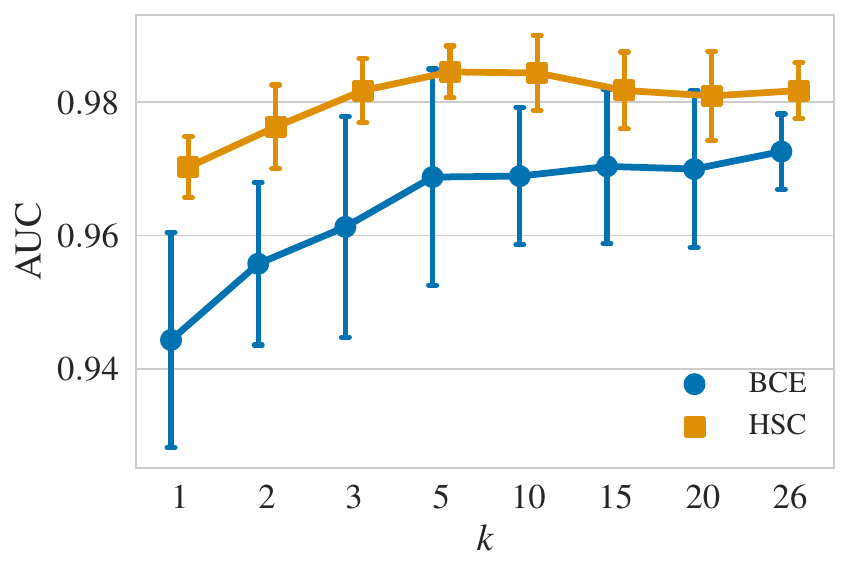}}
\subfigure[Class: 4]{\label{fig:mnistvsemnist_4}\includegraphics[width=0.329\linewidth]{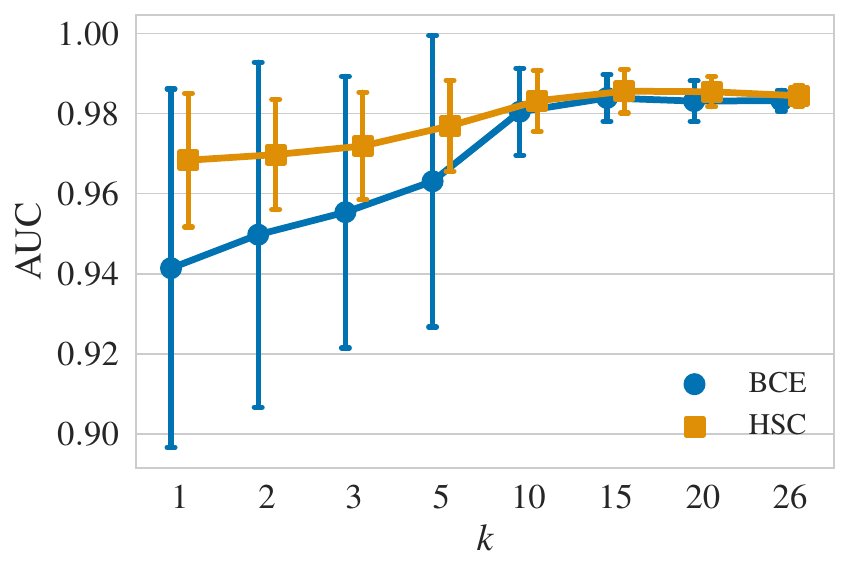}}
\subfigure[Class: 5]{\label{fig:mnistvsemnist_5}\includegraphics[width=0.329\linewidth]{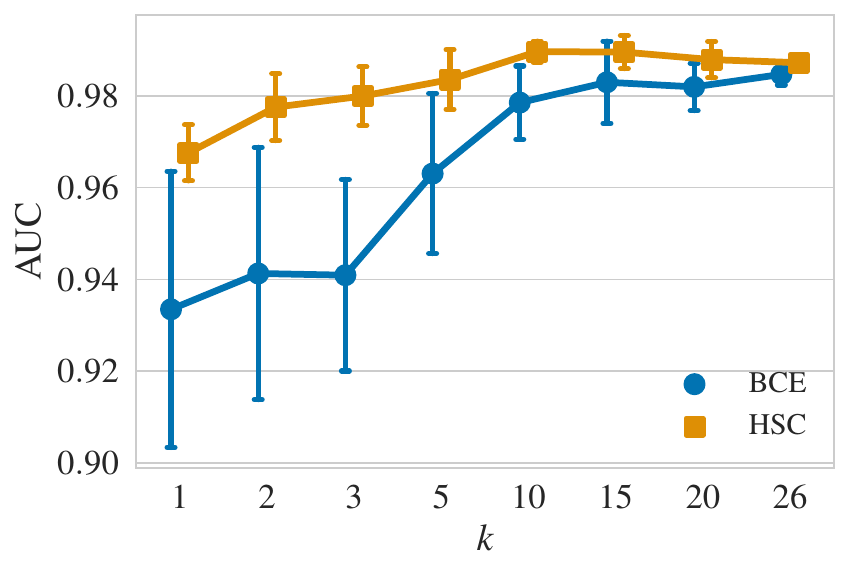}}
\subfigure[Class: 6]{\label{fig:mnistvsemnist_6}\includegraphics[width=0.329\linewidth]{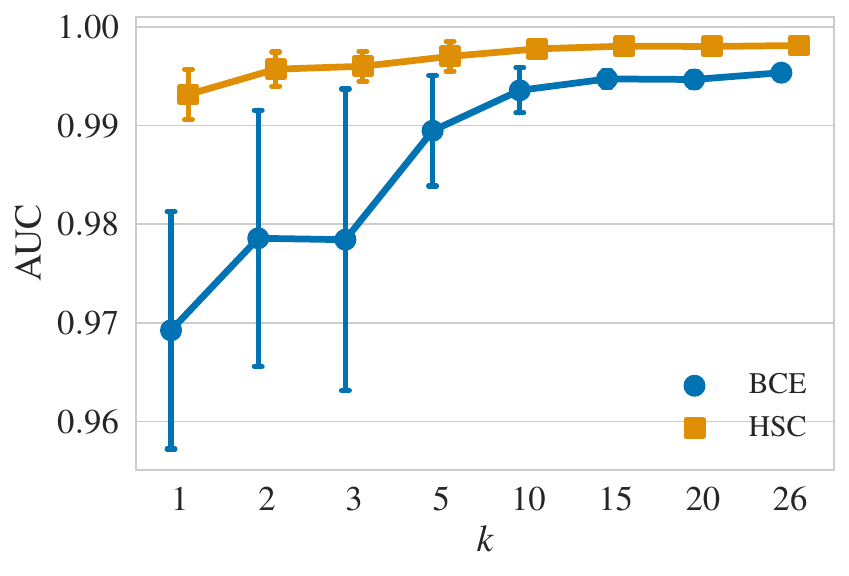}}
\subfigure[Class: 7]{\label{fig:mnistvsemnist_7}\includegraphics[width=0.329\linewidth]{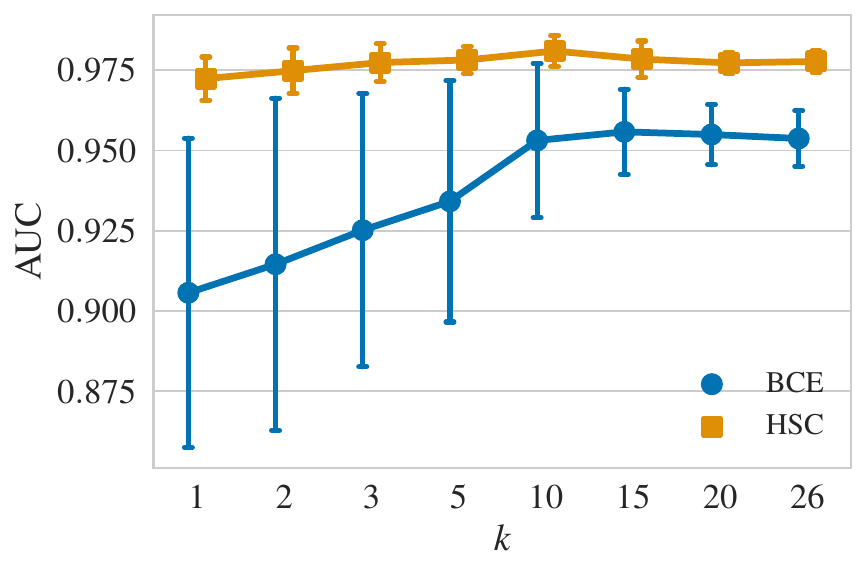}}
\subfigure[Class: 8]{\label{fig:mnistvsemnist_8}\includegraphics[width=0.329\linewidth]{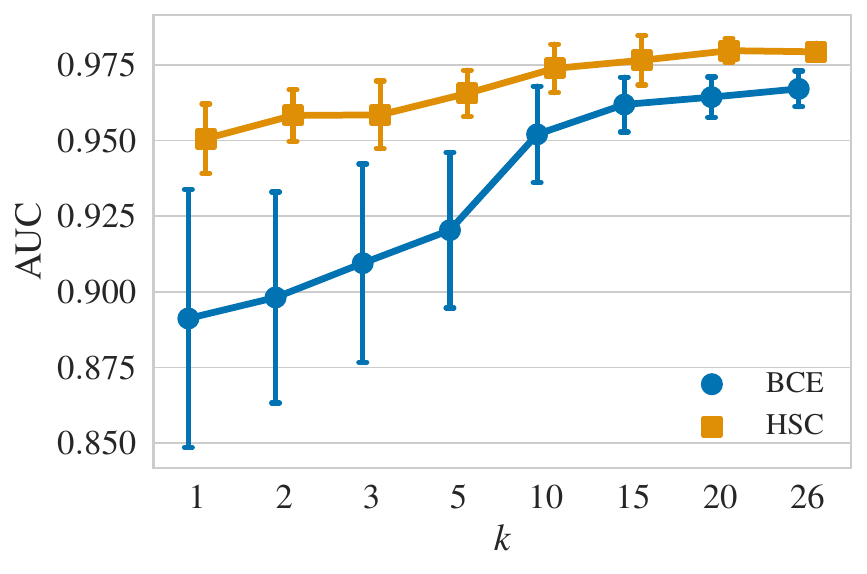}}
\subfigure[Class: 9]{\label{fig:mnistvsemnist_9}\includegraphics[width=0.329\linewidth]{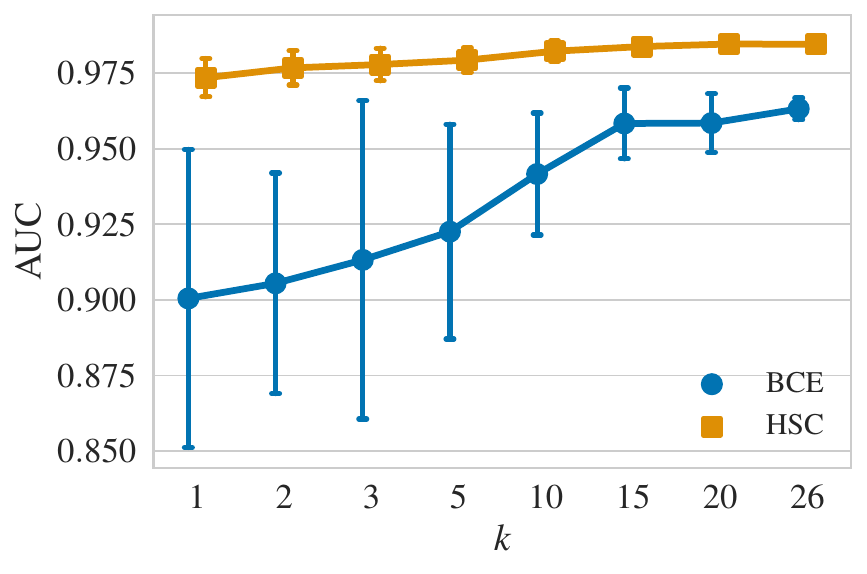}}
\label{fig:mnistvsemnist}
\end{figure*}

\begin{figure*}[th]
\centering
\caption{Mean AUC detection performance in \% (over 10 seeds) for all CIFAR-10 classes from the experiment in Appendix \ref{appx:oe_diversity} on varying the number of classes $k$ of the CIFAR-100 OE data. These plots correspond to Figure \ref{fig:diversity_cifar10}, but here we report the results for all individual classes.}
\subfigure[Class: airplane]{\label{fig:cifar10vscifar100_0}\includegraphics[width=0.329\linewidth]{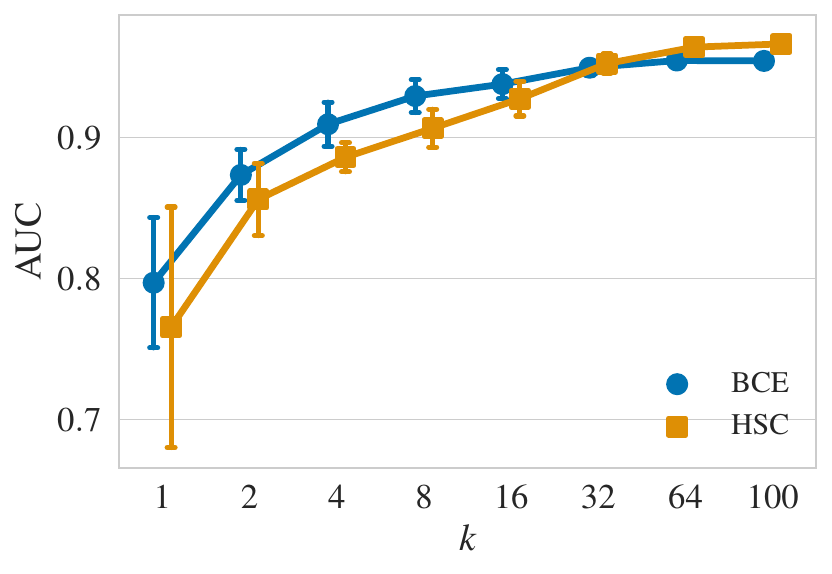}}
\subfigure[Class: automobile]{\label{fig:cifar10vscifar100_1}\includegraphics[width=0.329\linewidth]{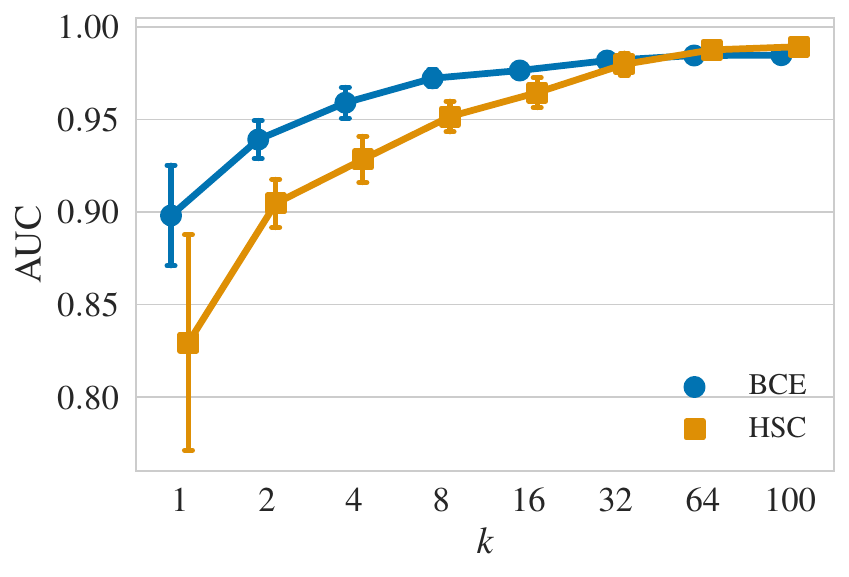}}
\subfigure[Class: bird]{\label{fig:cifar10vscifar100_2}\includegraphics[width=0.329\linewidth]{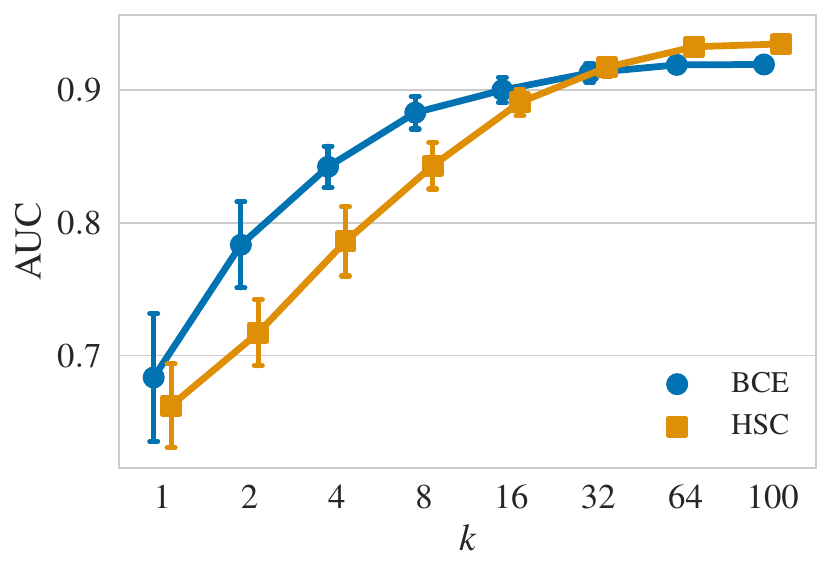}}
\subfigure[Class: cat]{\label{fig:cifar10vscifar100_3}\includegraphics[width=0.329\linewidth]{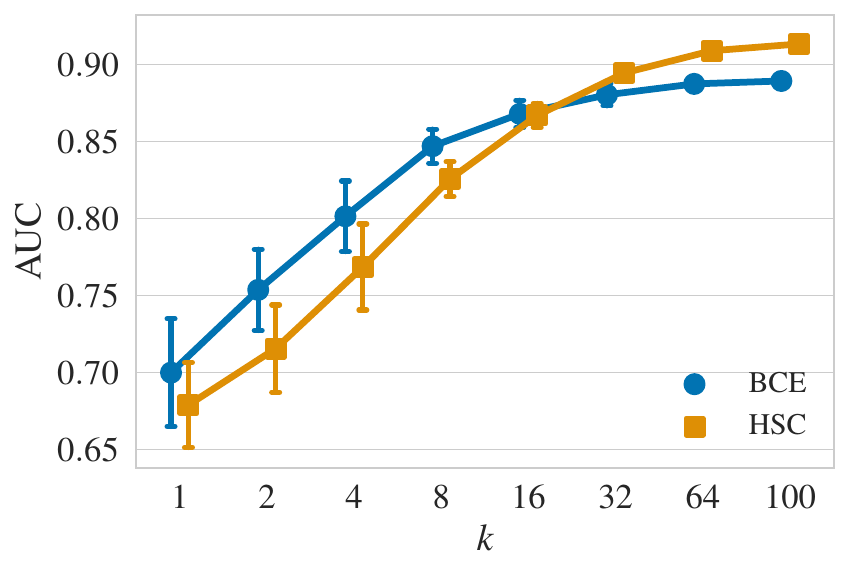}}
\subfigure[Class: deer]{\label{fig:cifar10vscifar100_4}\includegraphics[width=0.329\linewidth]{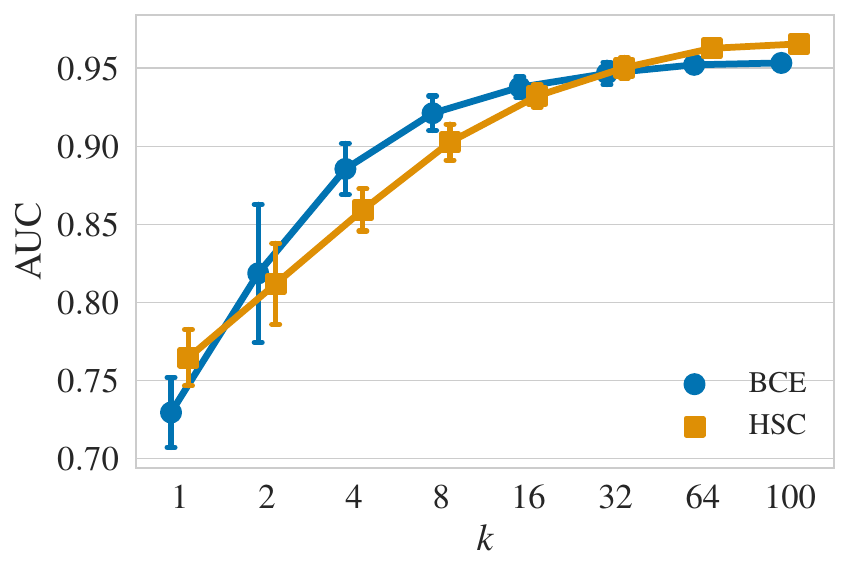}}
\subfigure[Class: dog]{\label{fig:cifar10vscifar100_5}\includegraphics[width=0.329\linewidth]{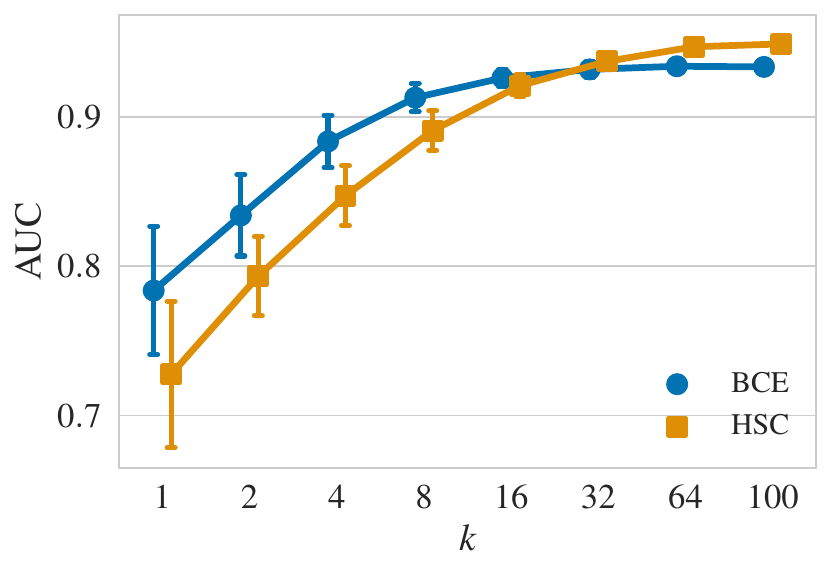}}
\subfigure[Class: frog]{\label{fig:cifar10vscifar100_6}\includegraphics[width=0.329\linewidth]{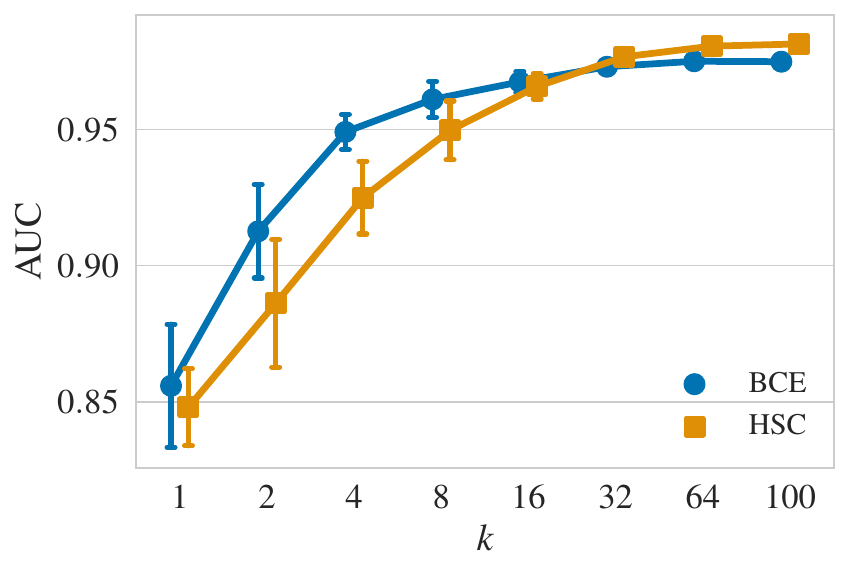}}
\subfigure[Class: horse]{\label{fig:cifar10vscifar100_7}\includegraphics[width=0.329\linewidth]{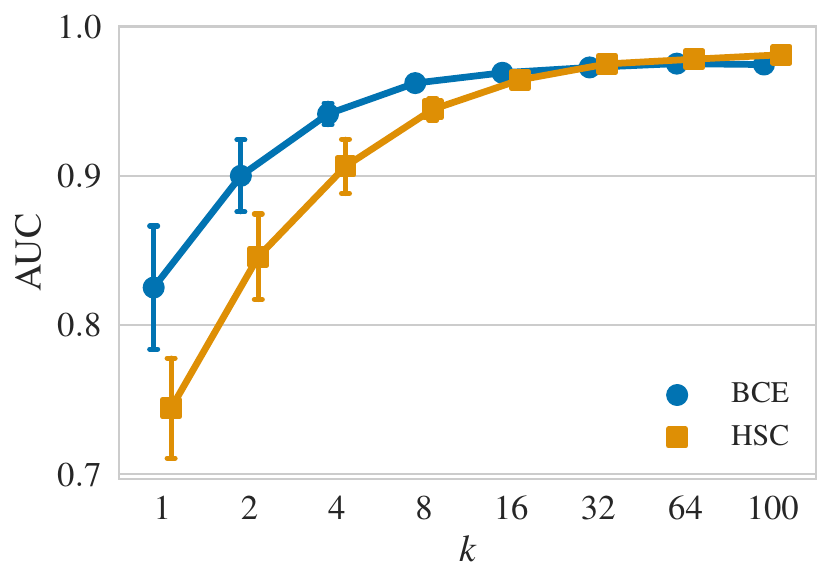}}
\subfigure[Class: ship]{\label{fig:cifar10vscifar100_8}\includegraphics[width=0.329\linewidth]{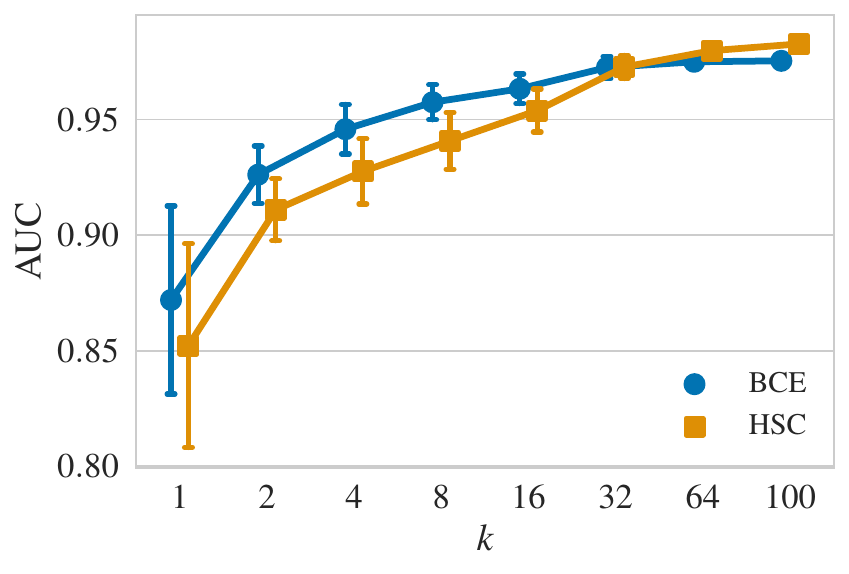}}
\subfigure[Class: truck]{\label{fig:cifar10vscifar100_9}\includegraphics[width=0.329\linewidth]{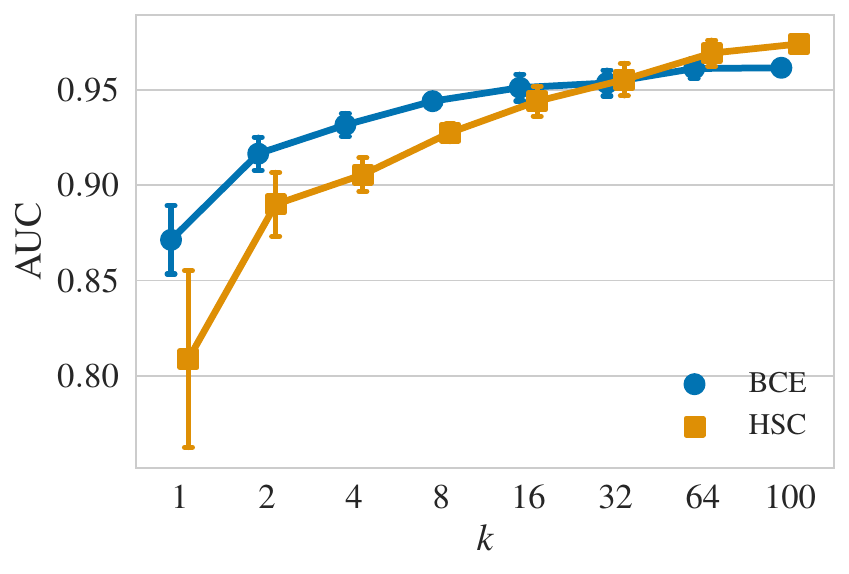}}
\label{fig:cifar10vscifar100}
\end{figure*}

For the CIFAR-10 one vs.-rest benchmark experiments from Section \ref{subsec:exp_cifar10}, we report the results for all individual classes and methods in Table \ref{tab:cifar10_classes}. 
We additionally report the results with standard deviations for our implementations in Table \ref{tab:cifar10_classes}. 
For the ImageNet-1K one vs.-rest benchmark experiments from Section \ref{subsec:exp_imagenet}, we present the performance over all individual classes with standard deviations in Table \ref{tab:imagenet_classes}. 
For the experiments on varying the number of OE samples, we include plots for all individual classes in Figure \ref{fig:cifarvstiny_classes} for CIFAR-10 and in Figures \ref{fig:imagenet1kvs22k_classes1} and \ref{fig:imagenet1kvs22k_classes2} for ImageNet-1K respectively.
Lastly, for the experiments on varying the diversity of OE data on MNIST with EMNIST-Letters OE, and CIFAR-10 with CIFAR-100 OE, we added the plots for all individual classes as well in Figures \ref{fig:mnistvsemnist} and \ref{fig:cifar10vscifar100}.

\begin{table*}[th]
    \caption{Mean AUC detection performance in \% (over 10 seeds) for all individual classes and methods on the CIFAR-10 one vs.~rest benchmark with 80MTI OE from Section \ref{subsec:exp_cifar10}.}
    \label{tab:cifar10_classes}
    \vspace{0.5em}
    \centering\small
    \begin{tabular}{lccccccccccc}
    \toprule
                & \multicolumn{5}{c|}{Unsupervised} & \multicolumn{3}{c|}{Unsupervised OE} & \multicolumn{3}{c}{Supervised OE}\\
    Class       & SVDD* & DSVDD*    & GT*  & IT*   & \multicolumn{1}{c|}{GT+*}    & GT+*     & DSAD  & \multicolumn{1}{c|}{HSC}   & Focal*    & Focal & BCE\\
    \midrule
    Airplane    & 65.6  & 61.7      & 74.7  & 78.5  & 77.5                          & 90.4      & 94.2      & 96.3                      & 87.6      & 95.9  & \bf 96.4\\
    Automobile  & 40.9  & 65.9      & 95.7  & 89.8  & 96.9                          & \bf 99.3  & 98.1      & 98.7                      & 93.9      & 98.7  & 98.8\\
    Bird        & 65.3  & 50.8      & 78.1  & 86.1  & 87.3                          & \bf 93.7  & 89.8      & 92.7                      & 78.6      & 92.3  & 93.0\\
    Cat         & 50.1  & 59.1      & 72.4  & 77.4  & 80.9                          & 88.1      & 87.4      & 89.8                      & 79.9      & 88.8  & \bf 90.0\\
    Deer        & 75.2  & 60.9      & 87.8  & 90.5  & 92.7                          & \bf 97.4  & 95.0      & 96.6                      & 81.7      & 96.6  & 97.1\\
    Dog         & 51.2  & 65.7      & 87.8  & 84.5  & 90.2                          & \bf 94.3  & 93.0      & 94.2                      & 85.6      & 94.1  & 94.2\\
    Frog        & 71.8  & 67.7      & 83.4  & 89.2  & 90.9                          & 97.1      & 96.9      & 97.9                      & 93.3      & 97.8  & \bf 98.0\\
    Horse       & 51.2  & 67.3      & 95.5  & 92.9  & 96.5                          & \bf 98.8  & 96.8      & 97.6                      & 87.9      & 97.6  & 97.6\\
    Ship        & 67.9  & 75.9      & 93.3  & 92.0  & 95.2                          & \bf 98.7  & 97.1      & 98.2                      & 92.6      & 98.0  & 98.1\\
    Truck       & 48.5  & 73.1      & 91.3  & 85.5  & 93.3                          & \bf 98.5  & 96.2      & 97.4                      & 92.1      & 97.5  & 97.7\\
    \midrule
    Mean AUC    & 58.8  & 64.8      & 86.0  & 86.6  & 90.1                          & 95.6      & 94.5      & 95.9  & 87.3      & 95.8  & \bf 96.1\\
    \bottomrule
\end{tabular}
\end{table*}

\begin{table*}[th]
    \caption{Mean AUC detection performance in \% (over 10 seeds) with standard deviations for all individual classes for our implementations on the CIFAR-10 one vs.~rest benchmark with 80MTI OE from Section \ref{subsec:exp_cifar10}.}
    \label{tab:cifar10_classes_}
    \vspace{0.5em}
    \centering\small
    \begin{tabular}{lcccc}
    \toprule
                & \multicolumn{2}{c|}{Unsupervised OE} & \multicolumn{2}{c}{Supervised OE}\\
    Class       & DSAD & \multicolumn{1}{c|}{HSC} & Focal & BCE \\
    \midrule
    Airplane    & 94.2 $\pm$ 0.34 & 96.3 $\pm$ 0.13 & 95.9 $\pm$ 0.11 & 96.4 $\pm$ 0.17\\
    Automobile  & 98.1 $\pm$ 0.19 & 98.7 $\pm$ 0.07 & 98.7 $\pm$ 0.09 & 98.8 $\pm$ 0.06\\
    Bird        & 89.8 $\pm$ 0.54 & 92.7 $\pm$ 0.27 & 92.3 $\pm$ 0.32 & 93.0 $\pm$ 0.14\\
    Cat         & 87.4 $\pm$ 0.38 & 89.8 $\pm$ 0.27 & 88.8 $\pm$ 0.33 & 90.0 $\pm$ 0.27\\
    Deer        & 95.0 $\pm$ 0.22 & 96.6 $\pm$ 0.17 & 96.6 $\pm$ 0.10 & 97.1 $\pm$ 0.10\\
    Dog         & 93.0 $\pm$ 0.30 & 94.2 $\pm$ 0.13 & 94.1 $\pm$ 0.21 & 94.2 $\pm$ 0.12\\
    Frog        & 96.9 $\pm$ 0.22 & 97.9 $\pm$ 0.08 & 97.8 $\pm$ 0.07 & 98.0 $\pm$ 0.09\\
    Horse       & 96.8 $\pm$ 0.15 & 97.6 $\pm$ 0.10 & 97.6 $\pm$ 0.16 & 97.6 $\pm$ 0.09\\
    Ship        & 97.1 $\pm$ 0.21 & 98.2 $\pm$ 0.09 & 98.0 $\pm$ 0.11 & 98.1 $\pm$ 0.08\\
    Truck       & 96.2 $\pm$ 0.22 & 97.4 $\pm$ 0.13 & 97.5 $\pm$ 0.12 & 97.7 $\pm$ 0.16\\
    \midrule
    Mean AUC    & 94.5 $\pm$ 3.30 & 95.9 $\pm$ 2.68 & 95.8 $\pm$ 2.97 & 96.1 $\pm$ 2.71\\
    \bottomrule
\end{tabular}
\end{table*}

\begin{table*}[th]
    \caption{Mean AUC detection performance in \% (over 10 seeds) for all individual classes for our implementations of the ImageNet-1K one vs.~rest benchmark with ImageNet-22K OE from Section \ref{subsec:exp_imagenet}. Note that for GT+* and Focal*, as reported in Table \ref{tab:imagenet} in the main paper, \citet{hendrycks2019using} do not provide results on a per class basis.}
    \label{tab:imagenet_classes}
    \vspace{0.5em}
    \centering\small
    \begin{tabular}{lcccc}
    \toprule
                & \multicolumn{2}{c|}{Unsupervised OE} & \multicolumn{2}{c}{Supervised OE}\\
    Class       & DSAD & \multicolumn{1}{c|}{HSC} & Focal & BCE \\
    \midrule
    acorn                   & 98.5 $\pm$ 0.28 & 98.8 $\pm$ 0.42 & 99.0 $\pm$ 0.15 & 99.0 $\pm$ 0.19 \\
    airliner                & 99.6 $\pm$ 0.16 & 99.8 $\pm$ 0.10 & 99.9 $\pm$ 0.02 & 99.8 $\pm$ 0.04 \\
    ambulance               & 99.0 $\pm$ 0.13 & 99.8 $\pm$ 0.13 & 99.2 $\pm$ 0.14 & 99.9 $\pm$ 0.07 \\
    american alligator      & 92.9 $\pm$ 1.06 & 98.0 $\pm$ 0.32 & 94.7 $\pm$ 0.67 & 98.2 $\pm$ 0.27 \\
    banjo                   & 97.0 $\pm$ 0.51 & 98.2 $\pm$ 0.41 & 97.0 $\pm$ 0.33 & 98.7 $\pm$ 0.22 \\
    barn                    & 98.5 $\pm$ 0.29 & 99.8 $\pm$ 0.05 & 98.7 $\pm$ 0.24 & 99.8 $\pm$ 0.08 \\
    bikini                  & 96.5 $\pm$ 0.84 & 98.6 $\pm$ 0.57 & 97.2 $\pm$ 0.89 & 99.1 $\pm$ 0.30 \\
    digital clock           & 99.4 $\pm$ 0.33 & 96.8 $\pm$ 0.79 & 99.8 $\pm$ 0.03 & 97.2 $\pm$ 0.29 \\
    dragonfly               & 98.8 $\pm$ 0.28 & 98.4 $\pm$ 0.16 & 99.1 $\pm$ 0.21 & 98.3 $\pm$ 0.04 \\
    dumbbell                & 93.0 $\pm$ 0.53 & 91.6 $\pm$ 0.88 & 94.0 $\pm$ 0.04 & 92.6 $\pm$ 0.97 \\
    forklift                & 90.6 $\pm$ 1.43 & 99.1 $\pm$ 0.33 & 94.2 $\pm$ 0.90 & 99.5 $\pm$ 0.09 \\
    goblet                  & 92.4 $\pm$ 1.05 & 93.8 $\pm$ 0.38 & 93.8 $\pm$ 0.27 & 94.7 $\pm$ 1.43 \\
    grand piano             & 99.7 $\pm$ 0.06 & 97.4 $\pm$ 0.37 & 99.9 $\pm$ 0.04 & 97.6 $\pm$ 0.34 \\
    hotdog                  & 95.9 $\pm$ 2.01 & 98.5 $\pm$ 0.34 & 97.2 $\pm$ 0.05 & 98.8 $\pm$ 0.34 \\
    hourglass               & 96.3 $\pm$ 0.37 & 96.9 $\pm$ 0.26 & 97.5 $\pm$ 0.17 & 97.6 $\pm$ 0.48 \\
    manhole cover           & 98.5 $\pm$ 0.29 & 99.6 $\pm$ 0.34 & 99.2 $\pm$ 0.09 & 99.8 $\pm$ 0.01 \\
    mosque                  & 98.6 $\pm$ 0.29 & 99.1 $\pm$ 0.26 & 98.9 $\pm$ 0.30 & 99.3 $\pm$ 0.15 \\
    nail                    & 92.8 $\pm$ 0.80 & 94.0 $\pm$ 0.76 & 93.5 $\pm$ 0.32 & 94.5 $\pm$ 1.37 \\
    parking meter           & 98.5 $\pm$ 0.29 & 93.3 $\pm$ 1.64 & 99.3 $\pm$ 0.04 & 94.7 $\pm$ 0.76 \\
    pillow                  & 99.3 $\pm$ 0.14 & 94.0 $\pm$ 0.47 & 99.2 $\pm$ 0.14 & 94.2 $\pm$ 0.42 \\
    revolver                & 98.2 $\pm$ 0.30 & 97.6 $\pm$ 0.25 & 98.6 $\pm$ 0.11 & 97.7 $\pm$ 0.68 \\
    rotary dial telephone   & 90.4 $\pm$ 1.99 & 97.7 $\pm$ 0.50 & 92.2 $\pm$ 0.33 & 98.3 $\pm$ 0.75 \\
    schooner                & 99.1 $\pm$ 0.18 & 99.2 $\pm$ 0.20 & 99.6 $\pm$ 0.02 & 99.1 $\pm$ 0.26 \\
    snowmobile              & 97.7 $\pm$ 0.86 & 99.0 $\pm$ 0.22 & 98.1 $\pm$ 0.15 & 99.1 $\pm$ 0.25 \\
    soccer ball             & 97.3 $\pm$ 1.70 & 92.9 $\pm$ 1.18 & 98.6 $\pm$ 0.13 & 93.6 $\pm$ 0.61 \\
    stingray                & 99.3 $\pm$ 0.20 & 99.1 $\pm$ 0.33 & 99.7 $\pm$ 0.04 & 99.2 $\pm$ 0.10 \\
    strawberry              & 97.7 $\pm$ 0.64 & 99.1 $\pm$ 0.20 & 99.1 $\pm$ 0.03 & 99.2 $\pm$ 0.22 \\
    tank                    & 97.3 $\pm$ 0.51 & 98.6 $\pm$ 0.18 & 97.3 $\pm$ 0.47 & 98.9 $\pm$ 0.13 \\
    toaster                 & 97.7 $\pm$ 0.56 & 92.2 $\pm$ 0.78 & 98.3 $\pm$ 0.05 & 92.2 $\pm$ 0.65 \\
    volcano                 & 89.6 $\pm$ 0.44 & 99.5 $\pm$ 0.09 & 91.6 $\pm$ 0.90 & 99.4 $\pm$ 0.19 \\
    \midrule
    Mean AUC                & 96.7 $\pm$ 2.98 & 97.3 $\pm$ 2.53 & 97.5 $\pm$ 2.43 & 97.7 $\pm$ 2.34 \\
    \bottomrule
\end{tabular}
\end{table*}

\end{document}